\documentclass{article}

\usepackage{microtype}
\usepackage{graphicx}
\usepackage{subcaption}
\usepackage{booktabs} 

\usepackage{xfrac}
\usepackage{colortbl}
\usepackage{bm}
\usepackage{amsfonts}
\usepackage{nicefrac}
\usepackage{latexsym}
\usepackage{multirow}
\usepackage{tabularx}
\usepackage{makecell}
\usepackage{xcolor}

\definecolor{mygreen}{RGB}{231,255,231}
\definecolor{myyellow}{RGB}{255,252,233}
\definecolor{mypurple}{RGB}{230,230,254}
\definecolor{bgcolor}{RGB}{255, 253, 248}
\definecolor{bgyellow}{RGB}{255, 251, 240}
\definecolor{bgred}{RGB}{255, 245, 246}

\definecolor{keywordcolor}{RGB}{178,34,34}

\usepackage{wrapfig}
\usepackage{enumitem}
\usepackage{titletoc}
\usepackage{arydshln}
\usepackage{caption}
\usepackage{float}
\usepackage{nicematrix}
\usepackage{dsfont}

\usepackage[colorlinks=true]{hyperref}
\usepackage{xurl}  

\definecolor{redlinkcolor}{rgb}{0.79607843, 0.25098039, 0.25882353}
\definecolor{bluecitecolor}{rgb}{0,0.36,0.69}
\hypersetup{linkcolor=redlinkcolor,citecolor=bluecitecolor,urlcolor=bluecitecolor}

\let\icmlorigaddcontentsline\addcontentsline
\usepackage[accepted]{icml2026}
\let\addcontentsline\icmlorigaddcontentsline

\usepackage{amsmath}
\usepackage{amssymb}
\usepackage{mathtools}
\usepackage{amsthm}

\usepackage[capitalize,noabbrev]{cleveref}

\theoremstyle{plain}

\theoremstyle{definition}

\theoremstyle{remark}

\usepackage[textsize=tiny]{todonotes}

\icmltitlerunning{SWE-IF: Aligning Code Evaluation with Human Preference}

\begin{document}

\twocolumn[
  \icmltitle{SWE-IF: Aligning Code Evaluation with Human Preference}

  \begin{icmlauthorlist}
    \icmlauthor{Ming Zhong}{gdm,uiuc}
    \icmlauthor{Xiang Zhou}{gdm}
    \icmlauthor{Ting-Yun Chang}{gdm,usc}
    \icmlauthor{Qingze Wang}{gdm}
    \icmlauthor{Nan Xu}{gdm}
    \icmlauthor{Xiance Si}{gdm} \\
    \icmlauthor{Dan Garrette}{gdm}
    \icmlauthor{Shyam Upadhyay}{gdm}
    \icmlauthor{Jeremiah Liu}{gdm}
    \icmlauthor{Jiawei Han}{uiuc}
    \icmlauthor{Benoit Schillings}{gdm}
    \icmlauthor{Jiao Sun}{gdm}
  \end{icmlauthorlist}
  \icmlaffiliation{gdm}{Google DeepMind}
  \icmlaffiliation{uiuc}{UIUC}
  \icmlaffiliation{usc}{USC}
  \icmlcorrespondingauthor{Ming Zhong}{mingz5@illinois.edu}
  \icmlcorrespondingauthor{Jiao Sun}{jiaosun@google.com}

  \icmlkeywords{Large Language Models, Code Generation, Evaluation, Instruction Following}

  \vskip 0.3in
]



\printAffiliationsAndNotice{}

\begin{abstract}
Large Language Models (LLMs) have catalyzed vibe coding, where users leverage LLMs to generate and iteratively refine code through natural language interactions until it passes their \textit{vibe check}. \textit{Vibe check} reflects human preference and goes beyond functionality: the solution should feel right, read cleanly, preserve intent, and remain correct. However, current code evaluation remains anchored to pass@k and captures only functional correctness, overlooking non-functional instructions that users routinely apply. In this paper, we hypothesize that instruction following is the missing piece underlying \textit{vibe check} besides functional correctness. To quantify models' code instruction-following capabilities with measurable signals, we present \textsc{VeriCode}, a taxonomy of 30 verifiable code instructions together with deterministic verifiers. We use the taxonomy to augment established evaluation suites, resulting in \textsc{SWE-IF}, a testbed to assess both instruction following and functional correctness. Evaluating 31 LLMs, we show that even the strongest models struggle to comply with multiple instructions and exhibit functional regression. Most importantly, \emph{a composite score of functional correctness and instruction following correlates best with human preference}, with instruction following emerging as the primary differentiator among LLMs. Our code, data, and taxonomy are available at \href{https://github.com/maszhongming/SWE-IF}{this link}.
\end{abstract}

\section{Introduction}

Large Language Models (LLMs) have reshaped how humans write code, fostering a workflow termed ``\textit{vibe coding}''~\citep{vibe_coding, vibe_coding_blog}. In this paradigm, AI's role shifts from a one-shot code completion tool for developers to an interactive collaborator for a broader audience, including users with limited coding experience. Through multi-turn natural language interactions, users can create and refine solutions from scratch, requiring the model to maintain context, adapt to evolving requirements, and iteratively improve the code until it meets their needs~\citep{programmers_assistant, intercode}. The user's final accept/reject choice serves as a real-time evaluation: what we call the ``\textit{vibe check},'' a subjective preference typically based on whether the solution feels right, reads cleanly, avoids obvious issues or anti-patterns, and preserves intent and correct functionality. This collaborative workflow, popularized by tools such as Copilot\footnote{\url{https://github.com/features/copilot}} and Cursor\footnote{\url{https://cursor.com}}, is rapidly becoming standard practice in modern software development~\citep{impact_of_ai, so_survey_2025}.

Despite the shift toward vibe coding, existing code evaluation remains anchored to functional correctness, typically measured as pass@k~\citep{humaneval,mbpp,swebench}. These metrics indicate whether code passes unit tests but abstract away non-functional expectations that users apply when selecting a response, including adherence to project conventions, documentation clarity, minimal and targeted edits, and preservation of prior intent across interactions. This disconnection is evident in platforms such as Copilot Arena~\citep{copilot_arena}, a large-scale vibe-checking scenario where human programmers choose preferred candidate snippets. Strikingly, rankings of code LLMs from Copilot Arena exhibit weak or negative correlations with functional scores on popular benchmarks. Moreover, pass@k remains a dominant verifiable reward signal in RLVR~\citep{deepseek_r1, agent_rlvr}, steering optimization toward an incomplete notion of code quality. Consequently, models can achieve high leaderboard scores yet fail the vibe check in practice, producing code that is technically correct but misaligned with user preferences.

\begin{figure*}[t]
    \centering
    \includegraphics[width=1.0\linewidth]{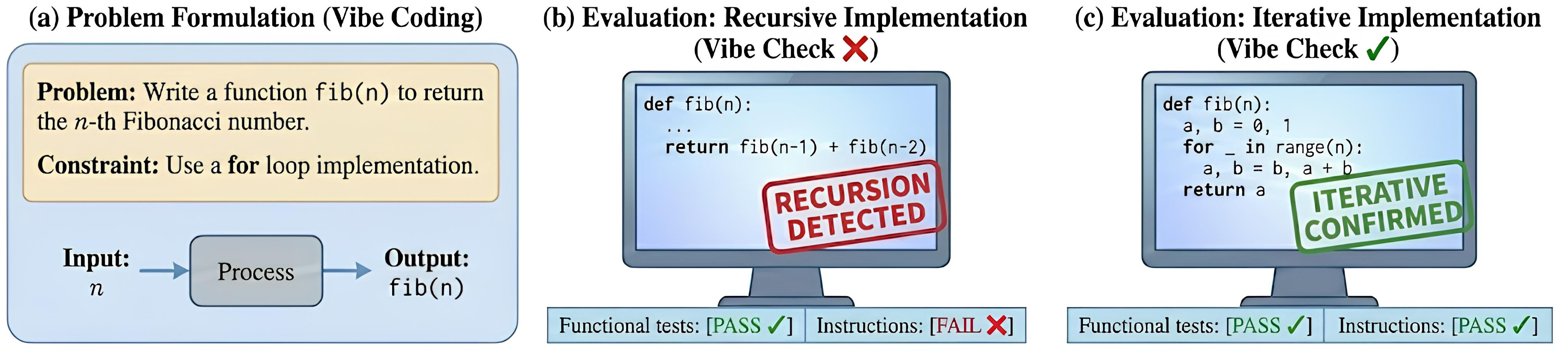}
    \caption{\textbf{\textit{Vibe check} goes beyond functionality}, requiring code to satisfy instructions such as coding style and logic patterns, which are also key factors of human preference. The example shown is a simplified case study and is not part of our taxonomy.}
    \label{fig:intro_fig}
\end{figure*}

To bridge this gap, we hypothesize that the non-functional signals emerging from interactions are an important, yet under-measured, component of the vibe check. We first introduce \textsc{VeriCode}, a taxonomy of verifiable code instructions designed to capture what users routinely screen for during code selection. Grounded in hundreds of rules from industrial linters and style guides, we perform manual curation and automated filtering to distill a core set of 30 instructions across five categories. Each instruction is paired with a verifier implemented using standard linters and static analysis. These verifiers yield a binary pass or fail score, enabling reliable automatic evaluation while also providing a verifiable and scalable reward source for model training.

Building on \textsc{VeriCode}, we augment established benchmarks, BigCodeBench~\citep{bigcodebench} and LiveCodeBench~\citep{livecodebench}, with these verifiable instructions to better simulate real-world interactions. We refer to the augmented variants as Big-SWE-IF and Live-SWE-IF. For each user query, an LLM-driven selector chooses a relevant and non-conflicting subset of instructions from our taxonomy to add as explicit constraints. Functional unit tests together with our instruction verifiers constitute a unified testbed, \textsc{SWE-IF}, which measures both functional correctness and instruction following (IF). Using this testbed, we evaluate 31 LLMs from 10 model families in two realistic settings: single-turn generation, in which the model must satisfy all constraints in one pass, and multi-turn editing, in which constraints are introduced sequentially while preserving prior intent. This setup allows us to study both dimensions across contexts.

Our analysis on \textsc{SWE-IF} testbed yields several key insights:

\begin{itemize}[leftmargin=*]
    \item \textbf{Non-functional instructions cause notable functional regression.} Although the added instructions do not target functionality, pass@1 decreases across all models. Under five instructions, average pass@1 drops by 5.85\% and 6.61\% on the two augmented benchmarks (Section~\ref{sec:result_func}).
    \item \textbf{Following multiple instructions remains challenging for LLMs.} Even the best performing model reaches only 46.75\% and 40.95\% success rate under five instructions on Big-SWE-IF and Live-SWE-IF (Section~\ref{sec:result_if}). Models also exhibit a position bias for instruction following, with mid-position instructions followed less reliably than those at the beginning or end (Section~\ref{sec:position_analysis}).
    \item \textbf{Single-turn vs. multi-turn interactions alter LLM behavior.} Under the same tasks, single-turn generation better preserves functionality but follows fewer instructions, whereas multi-turn editing achieves higher IF at the cost of more functional regressions (Sections~\ref{sec:result_func} and~\ref{sec:result_if}).
    \item \textbf{Human preference reflects a mixture of functional correctness and instruction following.} On the coding subset of LMArena~\citep{lmarena}, a composite of functional correctness and our IF score correlates better with model ratings than either metric alone, with IF emerging as the key differentiator among advanced models on the real-world programming tasks (Section~\ref{sec:correlation_analysis}).
\end{itemize}

In summary, this work establishes IF as an essential, yet overlooked, component of code evaluation. Our \textsc{VeriCode} taxonomy and \textsc{SWE-IF} testbed offer a concrete path to benchmark and develop models against a more human-aligned notion of code quality beyond functionality.

\section{\textsc{VeriCode}: A Taxonomy of Verifiable Code Instructions}
\label{sec:vericode}

To quantify IF, we first construct \textsc{VeriCode}, a taxonomy of verifiable code instructions. This section presents its design principles, construction process, and resulting structure.

\subsection{Design Principles}

We design \textsc{VeriCode} around four core principles to ensure it is rigorous, relevant, and useful:

\begin{itemize}[leftmargin=*]
    \item \textbf{Verifiability.} Each instruction is paired with an automated, deterministic verifier that returns a binary pass/fail signal, enabling objective and scalable evaluation.
    \item \textbf{Practice Grounding.} Instructions reflect developer expectations and conventions, drawing on widely used standards rather than synthetic or adversarial constraints.
    \item \textbf{Comprehensive Coverage.} The set spans key non-functional aspects, including coding style, logic patterns, documentation, error handling, and library constraints.
    \item \textbf{Difficulty.} Instructions are curated to be meaningfully challenging and diagnostic, ensuring that recent advanced LLMs exhibit imperfect adherence.
\end{itemize}

\subsection{Taxonomy Construction Process}

We carefully curate \textsc{VeriCode} in three stages: sourcing a candidate pool, performing multi-stage filtering, and finalizing the set with expert review and verifier implementation.

\textbf{Candidate Pool Sourcing.} We source our initial candidate pool from Ruff, an industry-standard Python linter that aggregates more than 800 rules drawn from popular tools\footnote{\url{https://docs.astral.sh/ruff/rules}}. This provides a high-coverage inventory of practices that users routinely follow and check. Static linting, however, inspects only code and cannot evaluate instructions that target the entire response (e.g., append a JSON explanation after the code block).  To close this gap, we add instructions focusing on documentation outside the code blocks, extending coverage beyond what static analysis can capture.

\textbf{Scope and Relevance Filtering.} The initial pool is first filtered for scope and relevance. We apply a top-down consolidation to address rule overlap, prioritizing broader instructions over their more specific subsets. This stage ensures that each instruction is broadly applicable across common coding tasks and not confined to niche scenarios.

\textbf{Difficulty Filtering.} We then screen the remaining candidates for difficulty. Using Gemini 2.5 Flash~\citep{gemini_2.5} on a challenging test set, BigCodeBench-Hard~\citep{bigcodebench}, we measure instruction following rate alongside functional correctness at pass@1. Any instruction with a success rate above 90\% and no degradation in pass@1 is removed. Borderline cases are flagged for manual review. This step focuses on non-trivial constraints that challenge advanced LLMs.

\textbf{Review and Verifier Implementation.} The final instruction set is manually reviewed by domain experts on the author team with coding-research experience to ensure clarity and real-world relevance. For verification, we prioritize linter-backed checks when available and implement deterministic tests using Abstract Syntax Tree (AST) analysis and regular expressions when no direct rule exists. All verifiers share a common interface: a testing function that returns a binary pass or fail, enabling scalable evaluation and reproducibility.

\subsection{Resulting \textsc{VeriCode} Taxonomy}

The multi-stage construction process yields our final verifiable taxonomy \textsc{VeriCode}.

\textbf{Taxonomy Structure.} The final set contains 30 instructions organized into five categories: \textit{Coding Style \& Conventions} \textit{(9)}, \textit{Logic \& Code Patterns} \textit{(9)}, \textit{Documentation \& Commenting} \textit{(6)}, \textit{Error Handling \& Exception Management} \textit{(4)}, and \textit{Library \& API Constraints} \textit{(2)}. The taxonomy is organized hierarchically: the root represents the overall concept of verifiable code instructions, the five categories form the top-level nodes, and the 30 individual instructions are the leaf nodes. Our current instantiation focuses on Python, the dominant language in code evaluation, but the framework is language-agnostic and can be applied to other languages using standard linters.

\textbf{Instruction Schema.} Each instruction specifies five necessary elements: 1) category, 2) description, 3) distinct prompts for both single-turn generation and multi-turn editing, 4) configurable \textit{parameters} with recommended or supported values, and 5) the verification code that returns a binary score. A full version of the instructions is available in Appendix~\ref{sec:full_case_study}.

\definecolor{mygreen}{HTML}{D5E8D4}
\definecolor{redlinkcolor}{HTML}{B30000}

\newcommand{\param}[1]{\texttt{\{#1\}}}

\renewcommand\arraystretch{1.4}

\begin{table*}[t]
    \centering \footnotesize
    \begin{NiceTabular}{m{2.5cm} m{5.0cm} m{2.5cm} m{4.0cm}}
    \toprule
    \textbf{Category} & \textbf{Prompt} & \textbf{Verifier} & \textbf{Parameter} \\
    \midrule

    \textbf{Coding Style \&\newline Conventions} 
        & Write code ensuring all lines are no longer than \param{line\_length} characters. 
        & \href{https://docs.astral.sh/ruff/rules/line-too-long}{\textcolor{redlinkcolor}{E501 Rule}}  
        & \texttt{line\_length} (int)\newline Recommended: 79 (classic), \newline 88 (modern) \\

    \hdashline[1pt/2pt]

    \textbf{Logic \& Code\newline Patterns} 
        & Ensure each function has at most \param{max\_branches} branches. 
        & \href{https://docs.astral.sh/ruff/rules/too-many-branches}{\textcolor{redlinkcolor}{PLR0912 Rule}} 
        & \texttt{max\_branches} (int)\newline Recommended: 2–4 \\

    \hdashline[1pt/2pt]

    \textbf{Documentation\newline \& Commenting} 
        & Document your code using the \param{convention} docstring format. 
        & \href{https://docs.astral.sh/ruff/settings/\#lintpydocstyle}{\textcolor{redlinkcolor}{D Rule}} 
        & \texttt{convention} (str)\newline Supported: Google, NumPy, \newline PEP 257 \\

    \hdashline[1pt/2pt]

    \textbf{Error Handling\newline \& Management} 
        & Replace all aliases with the canonical \emph{OSError} exception. 
        & \href{https://docs.astral.sh/ruff/rules/os-error-alias}{\textcolor{redlinkcolor}{UP024 Rule}} 
        & None \\

    \hdashline[1pt/2pt]

    \textbf{Library \& API\newline Constraints} 
        & Replace all \emph{os}, \emph{os.path}, \emph{glob}, and \emph{open} with their \emph{pathlib} equivalents. 
        & \href{https://docs.astral.sh/ruff/rules/\#flake8-use-pathlib-pth}{\textcolor{redlinkcolor}{PTH Rule}} 
        & None \\

    \bottomrule
    \end{NiceTabular}
    \vspace{3mm}
    \caption{\textbf{Refined examples from \textsc{VeriCode} taxonomy} (one per category shown). Each instruction maps to a verifiable linter rule and includes tunable parameters where applicable. Full versions are provided in Appendix~\ref{sec:full_case_study}.}
    \label{tab:taxonomy_case}
\end{table*}

A key feature of our taxonomy is its extensibility, which is achieved through the \textit{Parameters} field. As illustrated in Table~\ref{tab:taxonomy_case}, parameters such as \texttt{line\_length}, \texttt{max\_branches}, or documentation conventions allow a single instruction to generate multiple variants with different difficulty levels. This flexibility enables our set of 30 core instructions to be programmatically expanded into hundreds of distinct and checkable constraints, providing a scalable framework for future research.

\section{\textsc{SWE-IF}: A New Testbed for Code Evaluation}

Building on \textsc{VeriCode}, we introduce \textsc{SWE-IF} (Software Engineering Instruction Following) -- a testbed that extends standard code benchmarks with explicit, verifiable instructions grounded in software-engineering standards. It evaluates models under both single- and multi-turn protocols, measuring functional correctness as well as IF.

\subsection{Benchmark Augmentation}
\label{sec:benchmark_augmentation}

We ground our evaluation in established benchmarks, which allows us to leverage their unit tests to consistently measure functional correctness and situate our analysis within widely used evaluation suites. Concretely, we construct two augmented variants:

\vspace{-2mm}
\begin{itemize}[leftmargin=*]
    \item \textbf{Big-SWE-IF}, adapted from BigCodeBench to cover real-world programming tasks.
    \item \textbf{Live-SWE-IF}, adapted from LiveCodeBench to cover algorithmic/contest problems.
\end{itemize}
\vspace{-2mm}

This combination ensures that our evaluation covers a diverse range of coding challenges. Our augmentation process involves the following stages:

\textbf{Instruction Selection.} For each user query, we randomly permute the full set of 30 taxonomy instructions to form an ordered list. An LLM-based selector then scans this permuted list once, deciding whether to keep or discard each instruction based on two criteria: 1) \textit{Relevance}: the instruction must pertain to the query and plausibly influence the implementation, and 2) \textit{Non-conflict}: the instruction must not contradict any instruction already selected earlier in the pass. The accepted instructions, in this permuted order, constitute the constraint set used to evaluate all models.

\textbf{Parameter Selection and Validation.} Once the instructions are selected, we prompt an LLM to assign specific parameter values to each one. To guide this generation, the prompt includes the supported keys, types, ranges, and recommended values in our taxonomy, as well as the context of the user query, aiming for parameters that are both achievable and challenging. Finally, the generated parameters undergo a rule-based validation step: any parameter keys not explicitly defined for that instruction are removed, and any invalid values are reverted to predefined defaults.

Both Gemini 2.5 Pro~\citep{gemini_2.5} and Claude 4 Opus~\citep{claude_4} are tested as selectors in our augmentation pipeline, yielding similar instruction-category distributions. The final benchmark is augmented by Claude 4 Opus, chosen for its lower invalid-parameter rate (0.96\% vs. 2.47\% for Gemini 2.5 Pro). The resulting distributions show that instructions for \textit{Coding Logic}, \textit{Coding Style}, and \textit{Documentation} are most prevalent, with \textit{Coding Logic} being particularly frequent in the algorithm-focused LiveCodeBench (see Figure~\ref{fig:category_distribution} in the Appendix for a full breakdown).

\subsection{Evaluation Protocol}
\label{sec:evaluation}

\begin{figure*}[t]
    \centering
    \includegraphics[width=0.9\linewidth]{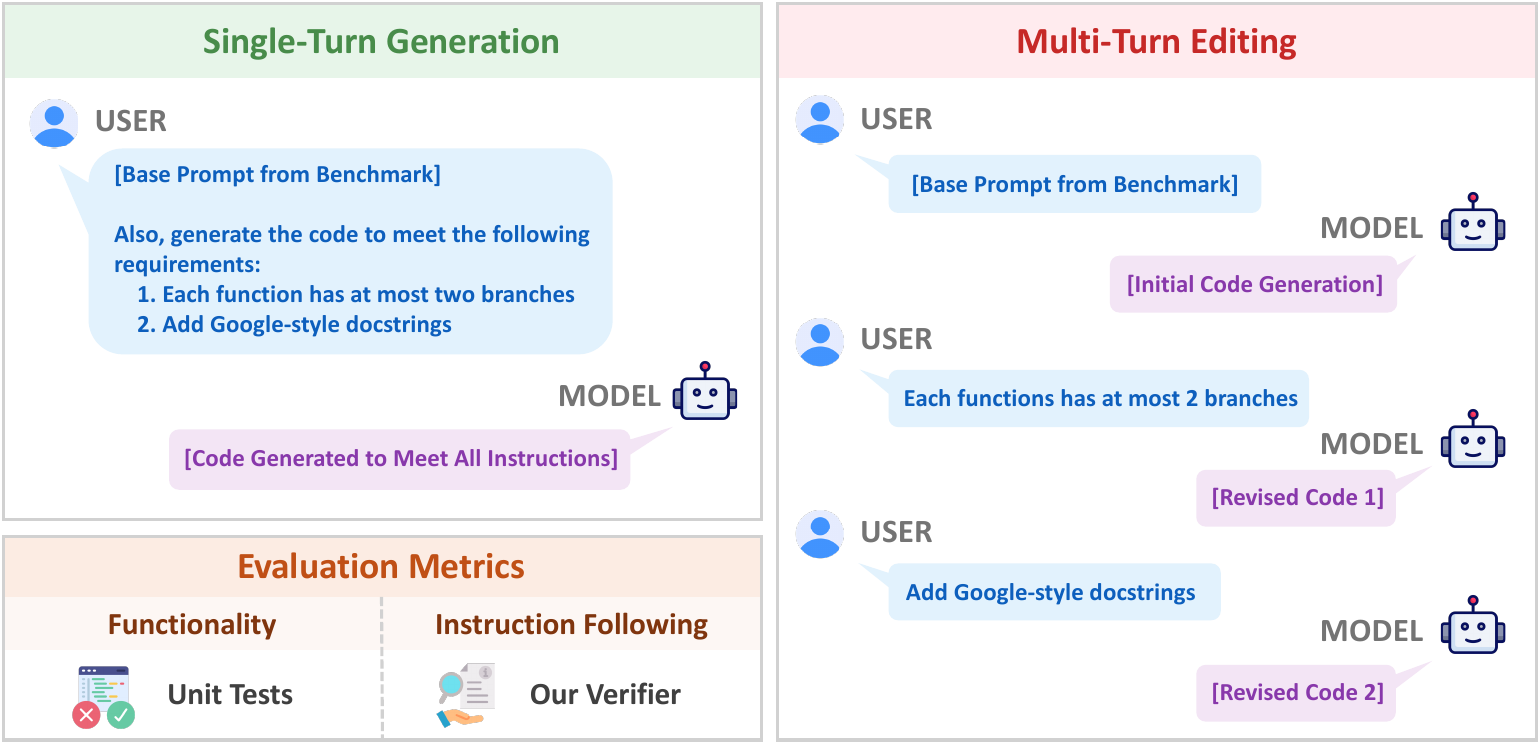}
    \caption{\textbf{Evaluation protocol simulates two interaction patterns}: \textit{single-turn generation}, where all instructions are given in one prompt, and \textit{multi-turn editing}, where instructions are introduced sequentially to refine a solution, measuring both functionality and IF.}
    \label{fig:eval_protocol}
\end{figure*}

Our evaluation protocol, illustrated in Figure~\ref{fig:eval_protocol}, mirrors real-world usage by providing single- and multi-turn interactive settings with two evaluation metrics.

\textbf{Interactive Settings.} We use two settings that differ in how instructions are presented:

\vspace{-2mm}
\begin{itemize}[leftmargin=*]
    \item \textit{Single-Turn Generation} presents all selected instructions after the original query within one prompt. The model returns a single implementation.
    \item \textit{Multi-Turn Editing} first elicits an initial implementation in response to the original query, then reveals the selected instructions one at a time. At each round, the model sees the full interaction history and updates the solution. The code from the last round is used for evaluation.
\end{itemize}
\vspace{-2mm}

\textbf{Evaluation Metrics.} For both settings, we evaluate the code on two axes:

\vspace{-2mm}
\begin{itemize}[leftmargin=*]
    \item \textit{Functionality:} We measure functional correctness with unit tests and report functional regression \(\mathrm{FR}_k\) from adding \(k\) instructions. Let \(S_k\) denote the functional score (typically pass@1) after injecting \(k\) instructions, with \(S_0\) the score on the original prompt. The rate is calculated as:
    \[
    \mathrm{FR}_k=\frac{S_0 - S_k}{S_0}.
    \]
    \item \textit{Instruction Following:} We report IF at two granularities. For a task with \(k\) instructions, let \(I_j\in\{0,1\}\) indicate whether instruction \(j\) passes its verifier. The \emph{instruction-level} score averages per-instruction passes, and the \emph{task-level} score requires all passes:
    \[
    \mathrm{IF}_{\text{instruction}}=\frac{1}{k}\sum_{j=1}^{k} I_j, \qquad \mathrm{IF}_{\text{task}}=\mathds{1}\!\left[\sum_{j=1}^{k} I_j = k\right].
    \]
    Here, a task refers to a benchmark problem together with its selected instruction set.
\end{itemize}
\vspace{-2mm}

\section{Experiments}

Based on \textsc{SWE-IF}, this section investigates the trade-off between functionality and instruction following, analyzes LLM behaviors, and correlates metrics with user preference.

\subsection{Experimental Setup}
\label{sec:exp_setup}

\textbf{Models.} To ensure a comprehensive analysis, we select a cohort of 31 powerful LLMs spanning 10 distinct model families, including Gemini~\citep{gemini_2.5}, Claude~\citep{claude_3.5, claude_4}, OpenAI~\citep{gpt_4o, o3}, DeepSeek~\citep{deepseek_v3, deepseek_r1}, Qwen~\citep{qwen2.5_coder, qwen3}, Grok~\citep{grok3, grok4}, Gemma~\citep{gemma3}, Mistral~\citep{mistral_medium_3}, MiniMax~\citep{minimax_m1}, and Kimi~\citep{kimi_k2}.

\textbf{Benchmarks.} We construct Big-SWE-IF and Live-SWE-IF by augmenting the full sets of BigCodeBench (1,140 instances) and LiveCodeBench v1--v6 (1,055 problems, May 2023 to May 2025). Each instance across both benchmarks is augmented with 5 instructions from \textsc{VeriCode} taxonomy, resulting in over 10K instruction-level evaluations.

\textbf{Implementation Details.} All models are queried via the Vertex AI\footnote{\url{https://cloud.google.com/vertex-ai/docs/reference/rest}} and OpenRouter\footnote{\url{https://openrouter.ai}} APIs. During benchmark augmentation, we use a deterministic temperature of 0.0. During evaluation, we follow the defaults of the underlying benchmarks: 0.0 for Big-SWE-IF and 0.2 for Live-SWE-IF. We enable thinking mode on all models that support it. For Claude models with thinking mode enabled, the API requires temperature 1.0, so we set it accordingly; all other models use the benchmark defaults. The context length is capped at 32,768 tokens for all models.

\renewcommand\arraystretch{1.3}
\definecolor{RegLight}{HTML}{D55C5A}
\definecolor{RegDeep}{HTML}{B00020} 
\newcommand{\RegDeepText}[1]{\textbf{\textcolor{RegDeep}{#1}}}

\begin{table*}[t]
    \centering \scriptsize
    \tabcolsep0.12 in
    \begin{NiceTabular}{lccccccccccc}
    \CodeBefore
        \rectanglecolor{bgyellow}{4-3}{10-7}
        \rectanglecolor{bgyellow}{12-3}{18-7}
        \rectanglecolor{bgred}{4-8}{10-12}
        \rectanglecolor{bgred}{12-8}{18-12}
    \Body
    \toprule
    \multirow[c]{2}{*}[-1.0ex]{\textbf{Model}} & &
    \multicolumn{5}{c}{\textbf{Single-Turn Generation} $\downarrow$} &
    \multicolumn{5}{c}{\textbf{Multi-Turn Editing} $\downarrow$} \\
    \cmidrule(lr){3-7} \cmidrule(lr){8-12}
    & Base & 1 Inst & 2 Inst & 3 Inst & 4 Inst & 5 Inst & 1 Inst & 2 Inst & 3 Inst & 4 Inst & 5 Inst \\
    \midrule
    \multicolumn{12}{l}{\textbf{Big-SWE-IF: Real-World Programming Tasks}} \\
    \midrule
    Gemini 2.5 Pro & 50.35 & 0.34 & 2.60 & 0.87 & -0.36 & 1.39 & 1.75 & 2.44 & 4.01 & 4.89 & \textcolor{RegLight}{5.04} \\
    Gemini 2.5 Flash & 47.37 & 0.74 & 1.12 & 2.60 & 1.31 & 2.41 & 0.93 & 1.12 & 1.48 & 2.98 & 3.72 \\
    Claude 4 Opus & 51.05 & -0.86 & -2.23 & -4.31 & -1.72 & -2.08 & 0.51 & 1.02 & 2.06 & 3.25 & 3.78 \\
    Claude 4 Sonnet & 51.84 & -0.17 & -0.52 & 0.33 & 0.50 & 0.50 & 0.85 & 2.03 & 3.55 & 4.05 & \textcolor{RegLight}{5.40} \\
    GPT 5 & 46.49 & 0.56 & \textcolor{RegLight}{5.66} & 2.26 & 3.20 & 1.89 & 1.70 & 2.82 & 4.35 & \textcolor{RegLight}{5.27} & \textcolor{RegLight}{5.46} \\
    o4 mini & 52.28 & 4.02 & \textcolor{RegLight}{9.39} & \textcolor{RegLight}{5.87} & \textcolor{RegLight}{7.38} & \textcolor{RegLight}{9.56} & 2.18 & 4.71 & \textcolor{RegLight}{7.04} & \textcolor{RegLight}{7.04} & \textcolor{RegLight}{8.05} \\
    Kimi K2 & 47.19 & -1.12 & -0.19 & -0.93 & 0.17 & 2.03 & 2.23 & 4.09 & 2.78 & 4.45 & \textcolor{RegLight}{6.12} \\
    \midrule
    \multicolumn{12}{l}{\textbf{Live-SWE-IF: Algorithmic Programming Contest Problems}} \\
    \midrule
    Gemini 2.5 Pro & 85.31 & -0.11 & 3.45 & 2.45 & 2.45 & 2.45 & 0.67 & 1.34 & 1.01 & 1.89 & 2.23 \\
    Gemini 2.5 Flash & 74.50 & 3.56 & \textcolor{RegLight}{5.34} & \textcolor{RegLight}{8.01} & \textcolor{RegLight}{5.60} & \textcolor{RegLight}{6.74} & 0.12 & 1.14 & 1.65 & 3.44 & 3.69 \\
    Claude 4 Opus & 68.72 & 4.55 & \textcolor{RegLight}{8.56} & \textcolor{RegLight}{8.41} & \textcolor{RegLight}{8.13} & \textcolor{RegLight}{8.96} & 2.07 & 1.38 & 1.51 & 2.34 & 2.34 \\
    Claude 4 Sonnet & 66.35 & 4.57 & 5.00 & 3.71 & \textcolor{RegLight}{6.99} & \textcolor{RegLight}{9.00} & 0.42 & 0.86 & 1.15 & 1.72 & 2.14 \\
    GPT 5 & 71.47 & 1.72 & 2.13 & 3.32 & \textcolor{RegLight}{7.16} & \textcolor{RegLight}{6.76} & 2.25 & 4.24 & \textcolor{RegLight}{5.57} & \textcolor{RegLight}{7.43} & \textcolor{RegLight}{9.02} \\
    o4 mini & 80.95 & \textcolor{RegLight}{5.74} & \textcolor{RegLight}{9.02} & \textcolor{RegLight}{9.02} & \RegDeepText{11.37} & \RegDeepText{12.29} & 3.63 & \textcolor{RegLight}{8.91} & \RegDeepText{10.19} & \RegDeepText{11.71} & \RegDeepText{15.92} \\
    Kimi K2 & 63.58 & \textcolor{RegLight}{8.92} & \RegDeepText{15.48} & \RegDeepText{16.07} & \RegDeepText{15.48} & \RegDeepText{16.36} & 2.64 & \textcolor{RegLight}{5.63} & \textcolor{RegLight}{9.50} & \RegDeepText{12.49} & \RegDeepText{12.79} \\
    \bottomrule
    \end{NiceTabular}
    \vspace{3mm}
    \caption{\textbf{Top-performing models still suffer from functional regression when non-functional instructions are added.} \emph{Base} is pass@1 on the original query. All other columns report the regression rate (\%) relative to \emph{Base}. \(k\) Inst is the number of added instructions. \textcolor{RegLight}{Light red} marks \(>5\%\) regression and \RegDeepText{deep red} denotes \(>10\%\). Full results for all 31 LLMs are listed in the Appendix~\ref{sec:func_results}.}
    \label{tab:main_func}
\end{table*}

\begin{figure*}[t!]
    \centering
    \begin{subfigure}[b]{0.495\textwidth}
        \centering
        \includegraphics[width=\textwidth]{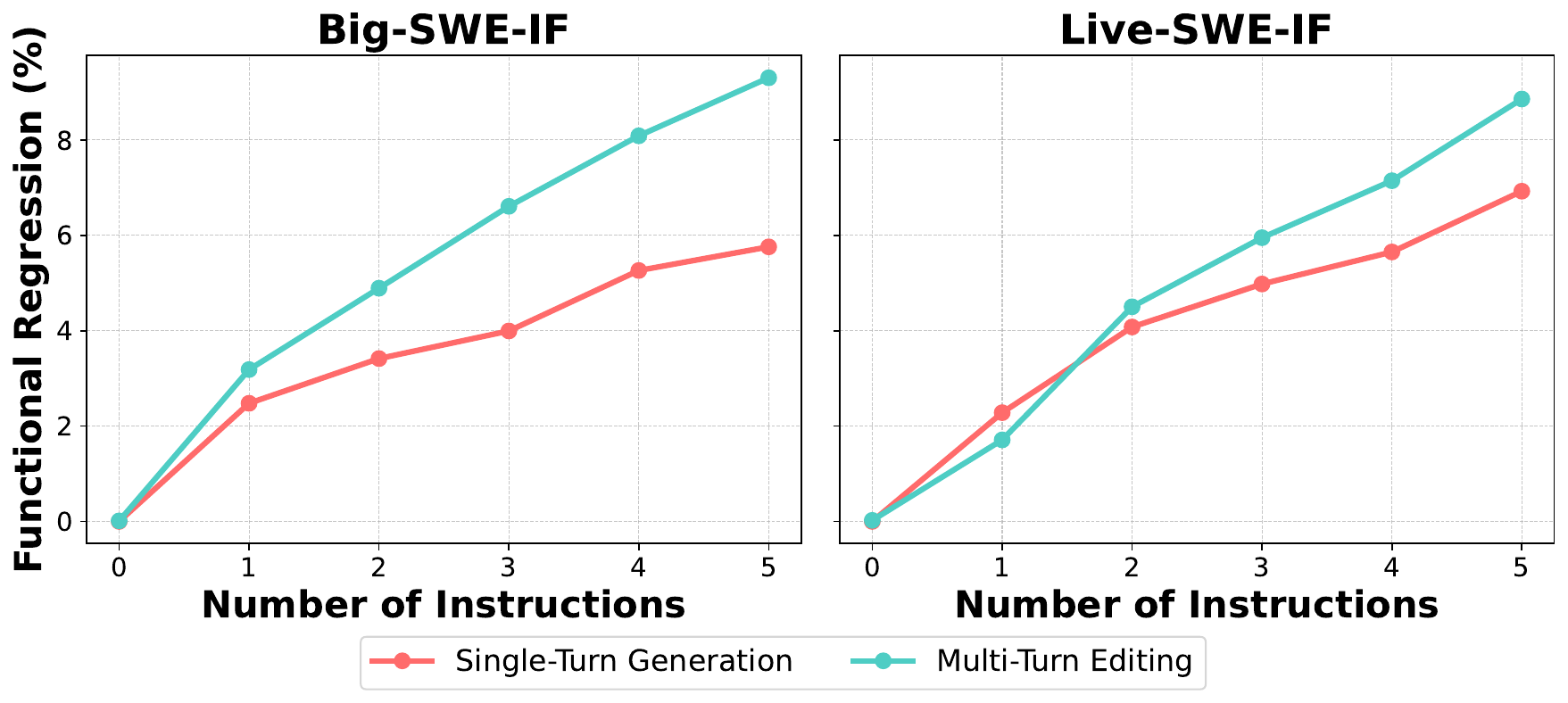}
        \caption{Functional Regression Rate.}
        \label{fig:main_func}
    \end{subfigure}
    \hfill
    \begin{subfigure}[b]{0.495\textwidth}
        \centering
        \includegraphics[width=\textwidth]{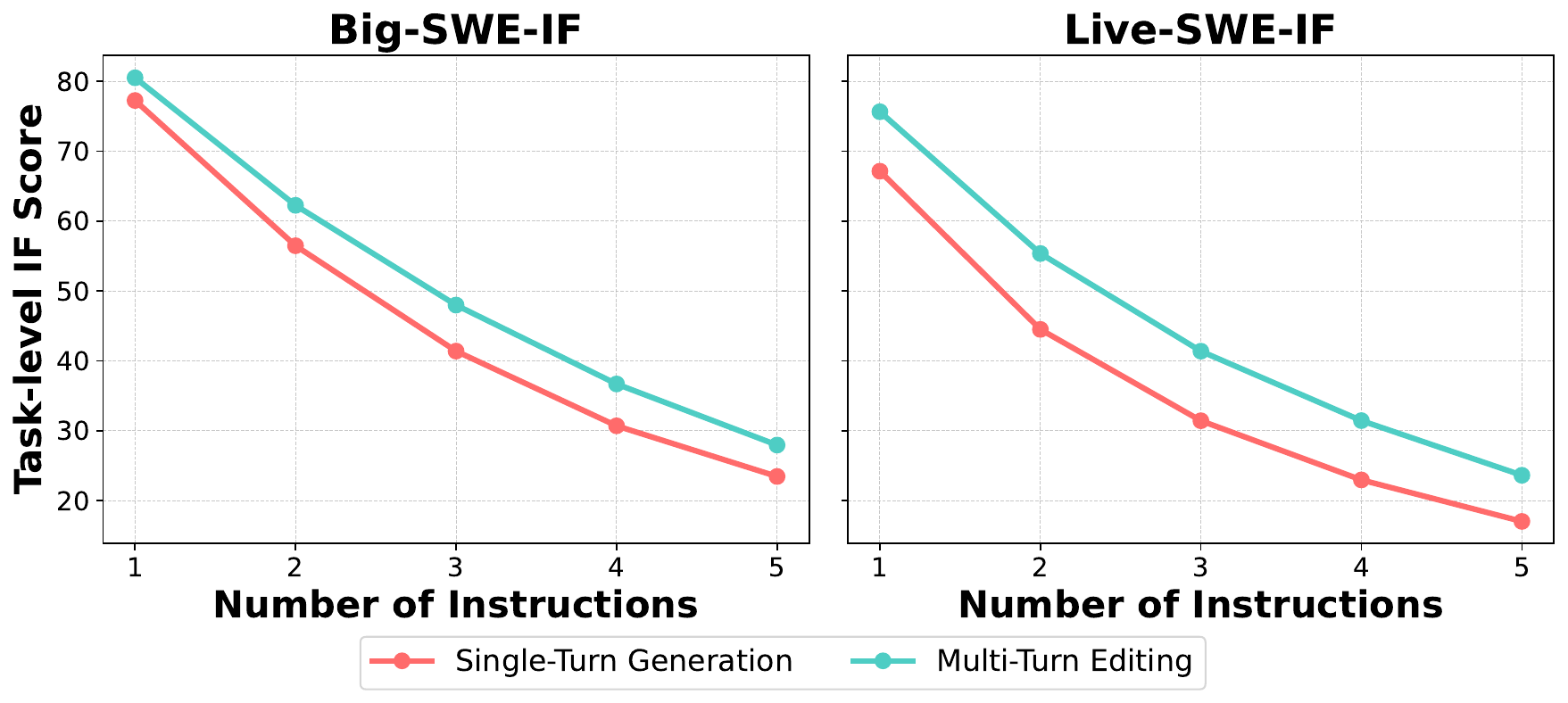}
        \caption{Task-Level Instruction Following.}
        \label{fig:main_if}
    \end{subfigure}
    \caption{\textbf{Trends averaged over all evaluated models.} As the number of instructions increases, functional regression grows steadily, while the task-level IF score drops markedly. Single-turn generation better preserves functionality, whereas multi-turn editing achieves higher instruction following. This pattern is consistent across all model families (Appendix~\ref{app:trending_analysis}). Robust instruction following thus emerges not as a quirk of any single model but as a frontier challenge for even the strongest LLMs.}
    \label{fig:main_trend}
    \vspace{-1mm}
\end{figure*}

\subsection{Results for Functionality}
\label{sec:result_func}

\renewcommand\arraystretch{1.3}
\definecolor{RegLight}{HTML}{D55C5A}
\definecolor{RegDeep}{HTML}{B00020} 

\begin{table*}[t]
    \centering \scriptsize
    \tabcolsep0.13 in
    \begin{NiceTabular}{lcccccccccc}
    \CodeBefore
        \rectanglecolor{bgyellow}{4-2}{10-6}
        \rectanglecolor{bgyellow}{12-2}{18-6}
        \rectanglecolor{bgred}{4-7}{10-11}
        \rectanglecolor{bgred}{12-7}{18-11}
    \Body
    \toprule
    \multirow[c]{2}{*}[-1.0ex]{\textbf{Model}} &
    \multicolumn{5}{c}{\textbf{Single-Turn Generation} $\uparrow$} &
    \multicolumn{5}{c}{\textbf{Multi-Turn Editing} $\uparrow$} \\
    \cmidrule(lr){2-6} \cmidrule(lr){7-11}
    & 1 Inst & 2 Inst & 3 Inst & 4 Inst & 5 Inst & 1 Inst & 2 Inst & 3 Inst & 4 Inst & 5 Inst \\
    \midrule
    \multicolumn{11}{l}{\textbf{Big-SWE-IF: Real-World Programming Tasks}} \\
    \midrule
    Gemini 2.5 Pro        & 82.19 & 60.70 & \textcolor{RegLight}{48.16} & \textcolor{RegLight}{37.46} & \textcolor{RegLight}{30.70} & 84.56 & 68.33 & 55.61 & \textcolor{RegLight}{44.21} & \textcolor{RegLight}{33.68} \\
    Gemini 2.5 Flash      & 81.67 & 61.05 & \textcolor{RegLight}{43.68} & \textcolor{RegLight}{30.53} & \RegDeepText{25.70} & 78.68 & 56.75 & \textcolor{RegLight}{40.96} & \RegDeepText{29.12} & \RegDeepText{21.75} \\
    Claude 4 Opus         & 88.77 & 76.32 & 64.21 & 52.98 & \textcolor{RegLight}{46.75} & 87.02 & 73.16 & 61.05 & 51.32 & \textcolor{RegLight}{42.11} \\
    Claude 4 Sonnet       & 84.91 & 67.19 & 52.28 & \textcolor{RegLight}{42.98} & \textcolor{RegLight}{35.26} & 86.40 & 72.54 & 61.23 & 51.05 & \textcolor{RegLight}{42.89} \\
    GPT 5                 & 82.89 & 67.63 & 54.04 & \textcolor{RegLight}{42.98} & \textcolor{RegLight}{34.39} & 84.91 & 72.37 & 62.98 & 55.26 & \textcolor{RegLight}{48.51} \\
    o4 mini               & 84.82 & 70.79 & 57.11 & \textcolor{RegLight}{47.98} & \textcolor{RegLight}{41.32} & 88.51 & 74.74 & 61.23 & 50.09 & \textcolor{RegLight}{41.84} \\
    Kimi K2               & 85.00 & 68.86 & 53.68 & \textcolor{RegLight}{41.23} & \textcolor{RegLight}{30.18} & 89.12 & 77.11 & 66.40 & 53.95 & \textcolor{RegLight}{44.04} \\
    \midrule
    \multicolumn{11}{l}{\textbf{Live-SWE-IF: Algorithmic Programming Contest Problems}} \\
    \midrule
    Gemini 2.5 Pro        & 75.83 & 56.78 & \textcolor{RegLight}{45.50} & \textcolor{RegLight}{37.63} & \RegDeepText{29.57} & 78.96 & 61.61 & 51.18 & \textcolor{RegLight}{41.04} & \textcolor{RegLight}{32.80} \\
    Gemini 2.5 Flash      & 66.54 & \textcolor{RegLight}{45.97} & \textcolor{RegLight}{32.89} & \RegDeepText{23.03} & \RegDeepText{17.06} & 72.80 & \textcolor{RegLight}{51.09} & \textcolor{RegLight}{34.98} & \RegDeepText{25.31} & \RegDeepText{17.82} \\
    Claude 4 Opus         & 78.86 & 57.91 & \textcolor{RegLight}{47.96} & \textcolor{RegLight}{38.96} & \textcolor{RegLight}{35.17} & 85.59 & 72.89 & 61.71 & 52.04 & \textcolor{RegLight}{43.70} \\
    Claude 4 Sonnet       & 75.73 & 56.40 & \textcolor{RegLight}{44.17} & \textcolor{RegLight}{35.36} & \RegDeepText{28.53} & 84.45 & 73.46 & 62.37 & 52.70 & \textcolor{RegLight}{44.64} \\
    GPT 5                 & 82.18 & 68.53 & 55.17 & \textcolor{RegLight}{47.01} & \textcolor{RegLight}{40.95} & 85.59 & 74.50 & 66.64 & 57.35 & 50.14 \\
    o4 mini               & 73.18 & 53.93 & \textcolor{RegLight}{43.22} & \textcolor{RegLight}{33.36} & \RegDeepText{27.20} & 81.52 & 66.64 & 54.60 & \textcolor{RegLight}{42.84} & \textcolor{RegLight}{32.61} \\
    Kimi K2               & 62.75 & \textcolor{RegLight}{41.61} & \RegDeepText{27.77} & \RegDeepText{19.05} & \RegDeepText{11.94} & 76.97 & 57.35 & \textcolor{RegLight}{44.17} & \textcolor{RegLight}{35.73} & \RegDeepText{27.87} \\
    \bottomrule
    \end{NiceTabular}
    \vspace{3mm}
    \caption{\textbf{Following multiple instructions remains challenging for top-performing models.} We report the task-level IF scores on both benchmarks. \textcolor{RegLight}{Light red} marks IF score $<50$ and \RegDeepText{deep red} indicates IF $<30$. Full results are provided in the Appendix~\ref{sec:if_results}.}
    \label{tab:main_if}
    \vspace{-2mm}
\end{table*}

\textbf{Adding non-functional instructions leads to functional regression.} Table~\ref{tab:main_func} reports regression rates on Big-SWE-IF for real-world programming and Live-SWE-IF for algorithmic problems. Handling multiple non-functional instructions is routine in practice, yet it still causes notable functional loss even for state-of-the-art models. On Big-SWE-IF, under multi-turn editing with five instructions, every model shows a regression above 5\% except Gemini 2.5 Flash and Claude 4 Opus. The effect is amplified on Live-SWE-IF: regressions above 5\% occur frequently for all models except Gemini 2.5 Pro, with the impact particularly pronounced for o4 mini and Kimi K2, which exceed 10\% in more than half of the test configurations.

\textbf{Single-turn generation better preserves functionality than multi-turn editing.} As illustrated in Figure~\ref{fig:main_func}, regression increases monotonically with the number of instructions. On Big-SWE-IF, average regression for single-turn climbs from 2.48\% with one instruction to 5.76\% with five, while multi-turn rises from 3.18\% to 9.31\% over the same range. On Live-SWE-IF, the gap is smaller: with two instructions, the two interaction modes are comparable, but as constraints increase, the single-turn setting gradually opens a clearer lead. Overall, single-turn generation more reliably preserves functionality, and its advantage grows with the number of instructions.

\subsection{Results for Instruction Following}
\label{sec:result_if}

\textbf{Task-level success collapses under multiple instructions.} Table~\ref{tab:main_if} presents the task-level IF score, where success requires satisfying all constraints simultaneously. The performance decay is rapid: with three or more instructions, most advanced models fall below 50 across both benchmarks. The decline is sharper on Live-SWE-IF, where 5 of the 7 leading models do not reach 30 in the single-turn setting. Such a steep drop is not entirely unexpected, as even the best models remain below 90 on a single instruction. With each added instruction, the probability of satisfying all constraints decreases multiplicatively, yielding an exponential decay in task-level success. Such performance degradation indicates that IF remains a challenge for state-of-the-art models and should be prioritized in both evaluation and training to meet the demands of real-world, multi-instruction scenarios.

\textbf{Multi-turn editing is more effective for following instructions.} In contrast to the functionality results, multi-turn editing consistently outperforms single-turn generation in instruction following, as shown in Figure~\ref{fig:main_if}. On Big-SWE-IF, the multi-turn setting maintains a 3\% to 4.5\% advantage in the task-level IF score. This gap widens on Live-SWE-IF, where the advantage reaches around 8\%. Given that the tasks are identical across settings, the consistent gap plausibly reflects the difference between the interactive patterns: single-turn must integrate all constraints in one pass and tends to prioritize preserving overall correctness, whereas the iterative nature of multi-turn supports targeted revisions that better satisfy newly introduced instructions.

\subsection{Instruction Position Analysis}
\label{sec:position_analysis}

\begin{figure}[t]
    \centering
    \includegraphics[width=1.0\columnwidth]{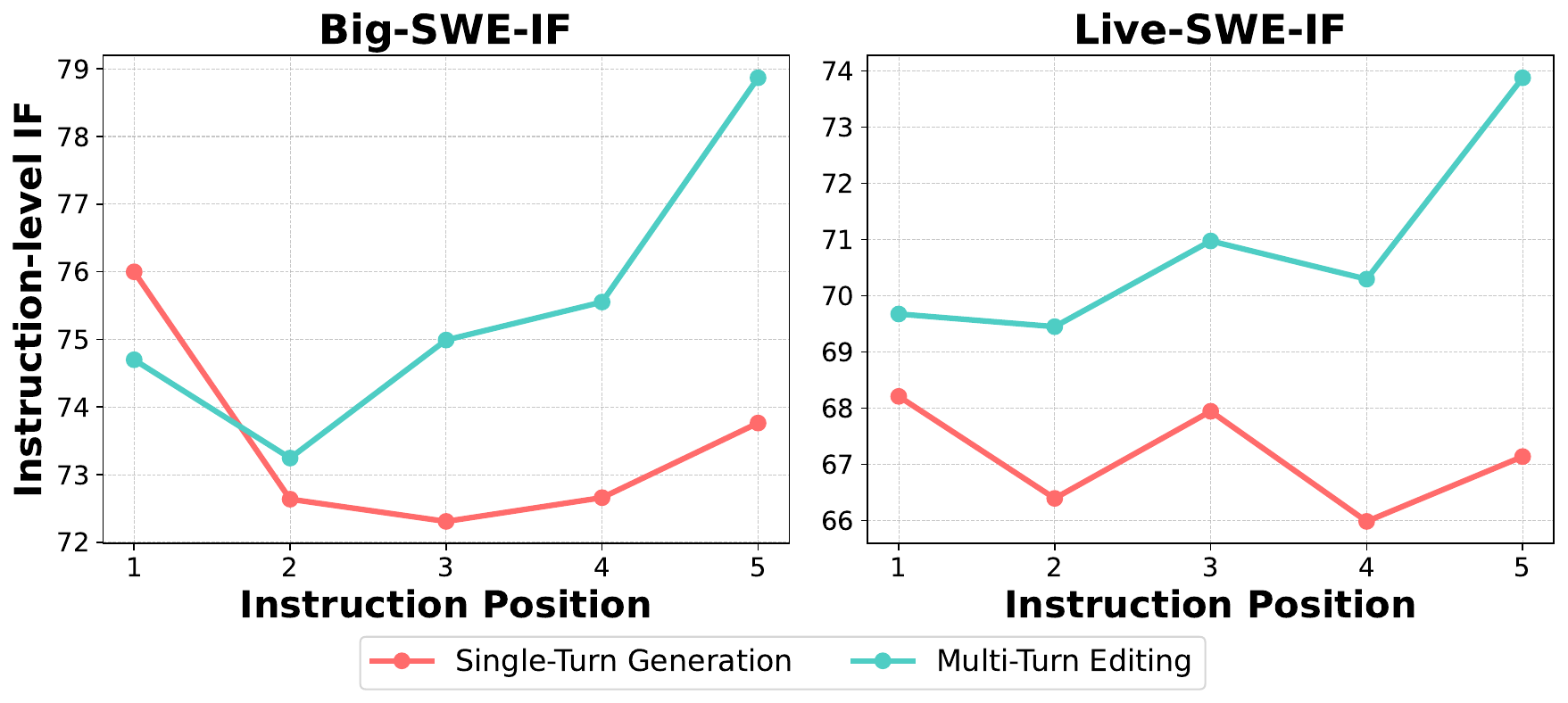}
    \caption{\textbf{Average instruction-level IF trends by position.} On Big-SWE-IF, scores trace a U-shaped ``lost-in-the-middle'' pattern; across both benchmarks, single-turn generation shows a primacy bias (peaking on the first instruction) while multi-turn editing shows a recency bias (peaking on the last).}
    \label{fig:position_analysis}
    \vspace{-2mm}
\end{figure}

\textbf{Models exhibit position bias in instruction following.} We define \textit{instruction position} as the index of each constraint: for single-turn generation, the number in the list appended to the base prompt; for multi-turn editing, the round in which the constraint is introduced, starting at 1. On Big-SWE-IF, Figure~\ref{fig:position_analysis} shows a clear U-shape, the classic ``lost-in-the-middle'' pattern typically reported for long-context generation~\citep{lost_in_the_middle}, despite our prompts being only a few hundred tokens long. Furthermore, single-turn generation shows a primacy bias, performing best on the first instruction, while multi-turn editing displays a clear recency bias, peaking on the final position. While the distinct U-shape does not generalize to Live-SWE-IF, the underlying positional preferences remain consistent: single-turn generation favors the first instruction, while multi-turn editing consistently performs best on the last.

\subsection{Correlating with Human Preference}
\label{sec:correlation_analysis}

\begin{figure*}[t]
    \centering
    \begin{subfigure}[b]{0.495\textwidth}
        \centering
        \includegraphics[width=\textwidth]{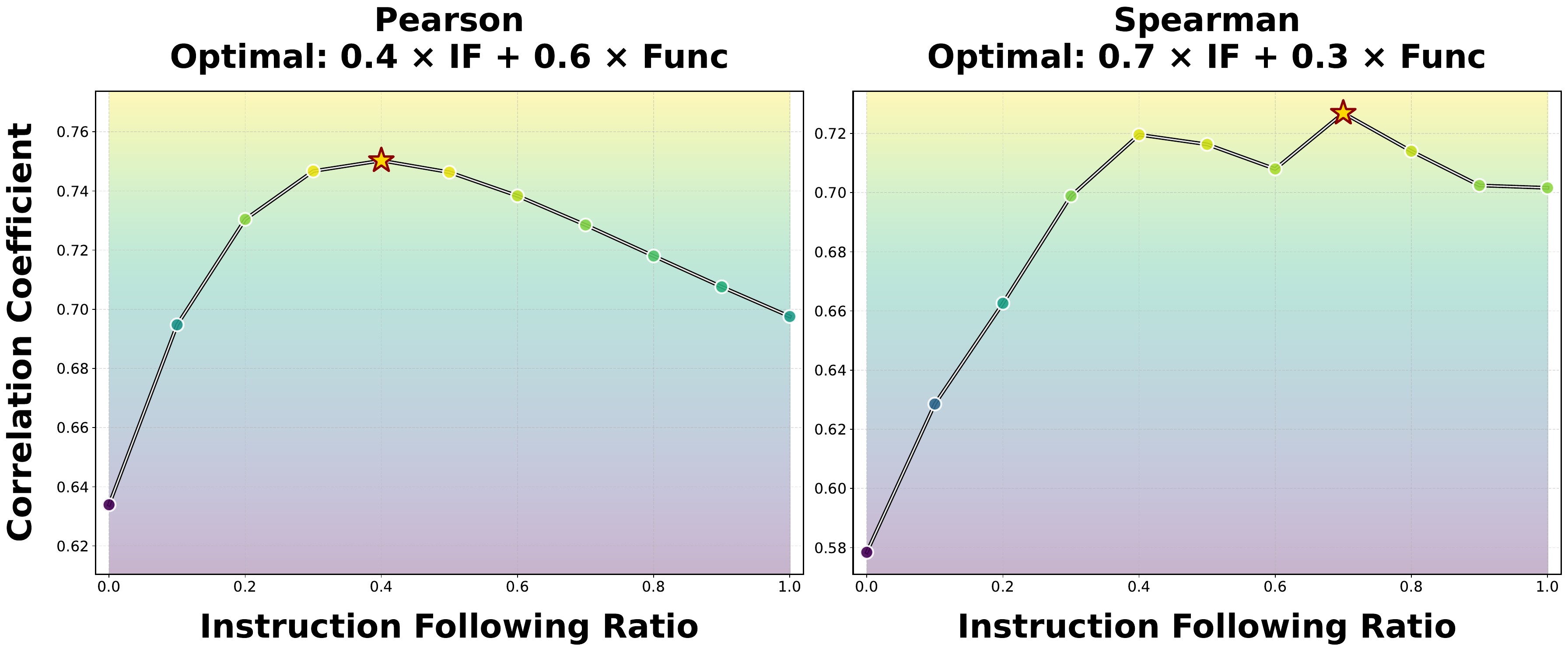}
        \caption{Real-World Programming}
        \label{fig:bigvibebench_correlation}
    \end{subfigure}
    \hfill
    \begin{subfigure}[b]{0.495\textwidth}
        \centering
        \includegraphics[width=\textwidth]{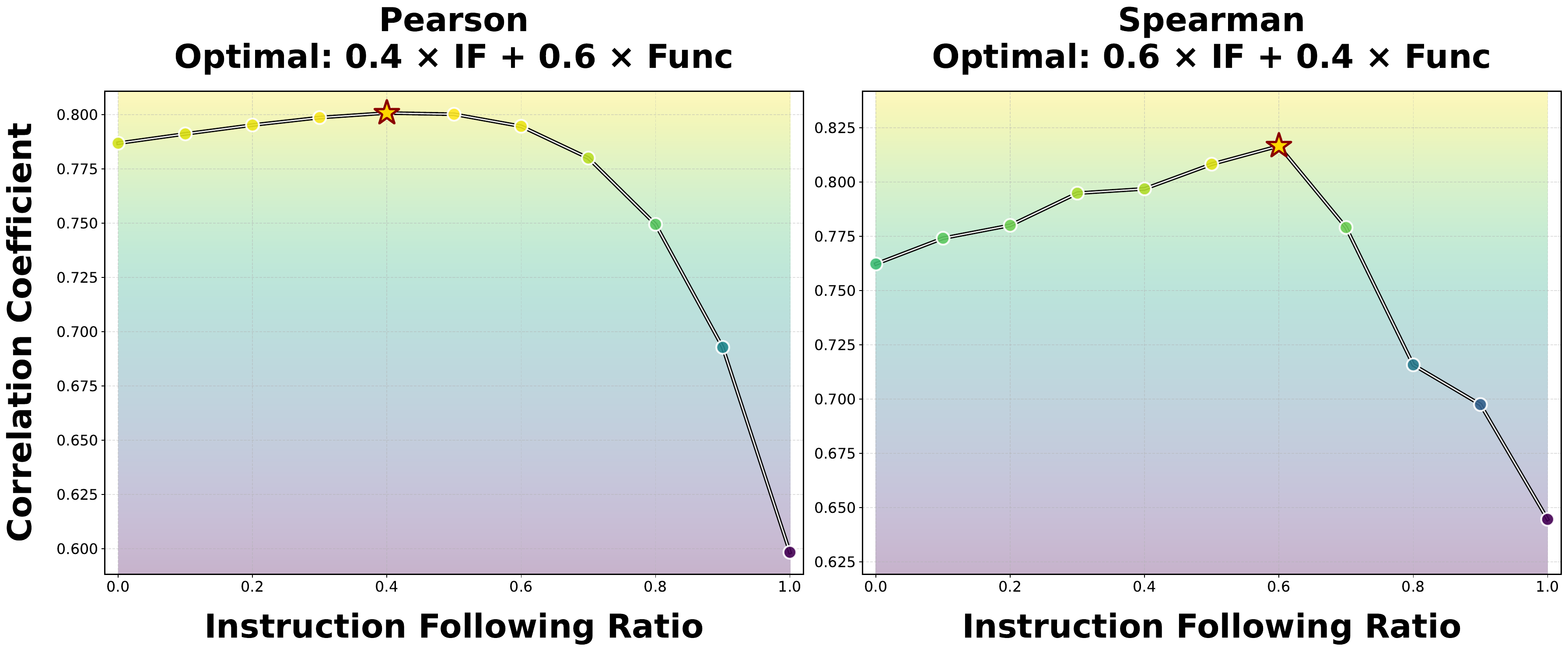}
        \caption{Algorithmic Programming}
        \label{fig:livevibebench_correlation}
    \end{subfigure}
    \caption{\textbf{Human preference aligns best with a mix of IF and functionality.} We correlate LMArena coding Elo with a composite score \(\alpha\,\text{IF} + (1-\alpha)\,\text{Func}\), where \(\alpha\in[0,1]\) is the weight on IF (x-axis). The peak correlation (starred) for both benchmarks is achieved with a mixture of the two metrics.}
    \label{fig:correlation_analysis}
    \vspace{-1mm}
\end{figure*}

Having established metrics for both functionality and instruction following, we now investigate how these signals relate to overall human preference in coding tasks.

To explore this, we use LMArena~\citep{lmarena}, currently the largest source of human preference data for LLMs. Its coding subset alone contains over 800K human votes, aggregated into Elo ratings for each model\footnote{\url{https://lmarena.ai/leaderboard/text/coding}}. We take the latest default Elo ratings from this subset (see Appendix Table~\ref{table:models}) and compute correlations against two metrics derived from \textsc{SWE-IF}: \textbf{Func}, defined as pass@1 on the original problems, and \textbf{IF}, taken from the single-turn setting under one instruction. We then evaluate a composite score \(\alpha\,\text{IF} + (1-\alpha)\,\text{Func}\) with \(\alpha\in[0,1]\), and report correlations.

\textbf{Human preference correlates best with a mixture of instruction following and functionality.} Across both benchmarks, the peak correlation occurs at intermediate \(\alpha\) (starred in Figure~\ref{fig:correlation_analysis}), indicating that neither IF nor Func alone explains preference as well as their combination. Concretely, on Big-SWE-IF, the optimum for Pearson correlation places a 40\% weight on IF (\(\alpha=0.4\)), while for Spearman correlation, the weight on IF rises to 70\% (\(\alpha=0.7\)). The optimal blend for Live-SWE-IF is remarkably similar. In all cases, the mixture outperforms either isolated metric by a clear margin. More correlation types and results with LMArena style control~\citep{style_control} disabled are reported in Appendix~\ref{sec:full_correlation}, with conclusions remaining consistent.

\textbf{Which single factor users value depends on the coding scenario.} While a mix is always best, the importance of each metric considered alone differs by the type of programming task. For the real-world programming tasks in Big-SWE-IF, instruction following plays a more critical role. On the Spearman correlation, pure IF (\(\alpha=1\)) correlates over 0.1 points higher with human preference than pure Func (\(\alpha=0\)). For algorithmic programming tasks in Live-SWE-IF, the opposite is true: pure Func holds a clear advantage over pure IF. This suggests that for practical, day-to-day coding, users place a high value on a model’s ability to adhere to non-functional instructions, whereas, in competitive programming scenarios, functional correctness is the paramount factor.

\textbf{Overall implication.} Our results provide evidence that IF is a critical, under-measured component of human preference in coding tasks. Beyond functional correctness, adherence to non-functional constraints offers a strong signal for distinguishing real-world utility. Consequently, integrating IF alongside functionality in both evaluation and training provides a practical path toward models that align more closely with real-world user preferences.

\section{Related Work}

\paragraph{Instruction Following.} Research in general instruction following focuses on stress-testing models with synthetic constraints (e.g., forced word repetition) and evaluates with either deterministic checkers~\citep{ifeval, vff, ifbench} or LLM-as-a-judge~\citep{followbench, infobench}. A prevailing trend leverages large-scale, verifiable instructions to boost capabilities via post-training, such as SFT and RL~\citep{vff, ifbench}.
In contrast, instructions in the coding domain are tied to practical software development, concerning aspects such as logic patterns, coding style, and library usage. Prior work is sparse, and existing benchmarks for such code instructions lack verifiability. They typically compare to ground truth with DiffBLEU~\citep{nofuneval} or use LLM and human judgment~\citep{codeif}, which is unreliable and hard to scale. To bridge this gap, we introduce a taxonomy of verifiable code instructions, each paired with a verifier, enabling scalable evaluation and training.

\paragraph{Code Evaluation.} Functional correctness dominates code evaluation: the generated code is run against unit tests, from snippet-level functions~\citep{humaneval, mbpp, apps, classeval, ds1000, evalplus, livecodebench, bigcodebench, livecodebench_pro} to repository-level tasks~\citep{swebench, swebench_verified, swt_bench, swebench_multimodal, commit0, swebench_multi_language, swebench_live}. Research on non-functional requirements is a relatively small branch of research, covering aspects like adherence to task-oriented instructions~\citep{codeif}, runtime efficiency, maintainability, and security~\citep{nofuneval}.
We move beyond evaluating these aspects in isolation. On top of \textsc{SWE-IF} testbed, we systematically analyze the trade-off between functional correctness and instruction following, and provide evidence that human preference reflects a composite of both dimensions.

\section{Conclusion}

In this paper, we challenged the prevailing focus on functional correctness in code evaluation. We study the \textit{vibe check} as a subjective judgment tied to real-world human preference and approximate it with measurable signals. We present \textsc{VeriCode}, a verifiable taxonomy of non-functional code instructions, and \textsc{SWE-IF}, a testbed that augments established evaluation suites. Across 31 leading LLMs, a composite of functional correctness and instruction following predicts human preference substantially better than either metric alone. Our work calls for moving beyond pass@k and for optimizing both functional and non-functional qualities in future research for coding. Our code, data, and taxonomy are available at \href{https://github.com/maszhongming/SWE-IF}{this link}.

\section*{Acknowledgements}

We sincerely thank the anonymous reviewers for their careful reading and valuable feedback, which helped us substantially improve the clarity, rigor, and presentation of this work. We are also grateful to Heng-Tze Cheng, Le Hou, Quoc Le, Charles Sutton, Kefan Xiao, and Pengcheng Yin for the insightful discussions and constructive suggestions that helped shape and strengthen this project.

\section*{Impact Statement}

This paper presents work whose goal is to advance the field of machine learning. There are many potential societal impacts of our work, none of which we feel must be specifically highlighted here.

\bibliography{custom}
\bibliographystyle{icml2026}

\newpage
\appendix
\onecolumn
\newcommand\DoToC{%
  \startcontents
  \setcounter{tocdepth}{2}
  \printcontents{}{1}{\textbf{Appendix Table of Contents}\vskip3pt\hrule\vskip5pt}
  \vskip3pt
}
\DoToC

\newpage
\clearpage
\section{\textsc{VeriCode} Taxonomy}
\label{app:vericode}

\subsection{Verification Code with Ruff}

\begin{figure*}[h]
    \centering
    \includegraphics[width=0.9\linewidth]{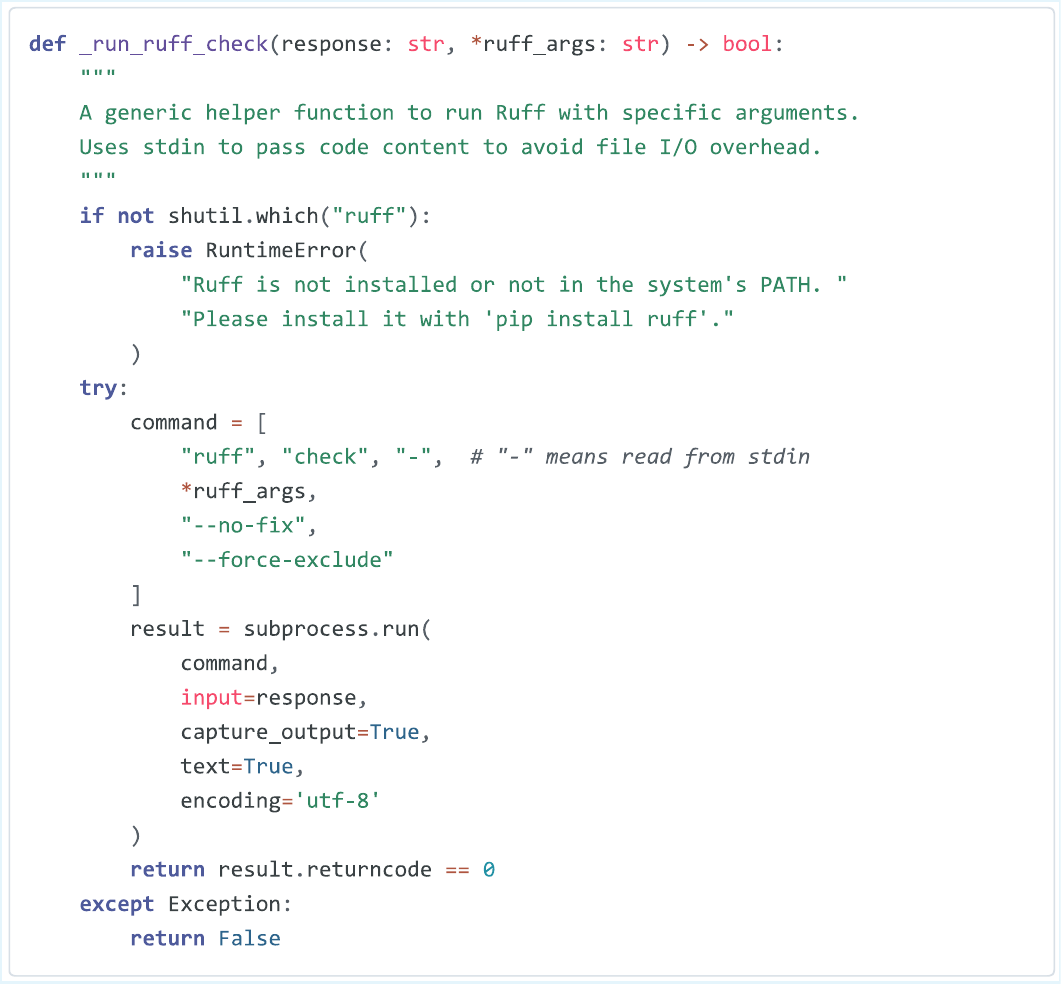}
    \caption{\textbf{Implementation of the core helper function} used to run Ruff checks within \textsc{VeriCode}.}
    \label{fig:ruff_helper_function}
\end{figure*}

Given that 27 of the 30 verifiers in our \textsc{VeriCode} taxonomy are implemented via Python linter Ruff, we present the helper function in Figure~\ref{fig:ruff_helper_function}.

\newpage
\clearpage
\subsection{Case Studies from \textsc{VeriCode}}
\label{sec:full_case_study}

\begin{figure*}
    \centering
    \includegraphics[width=0.9\linewidth]{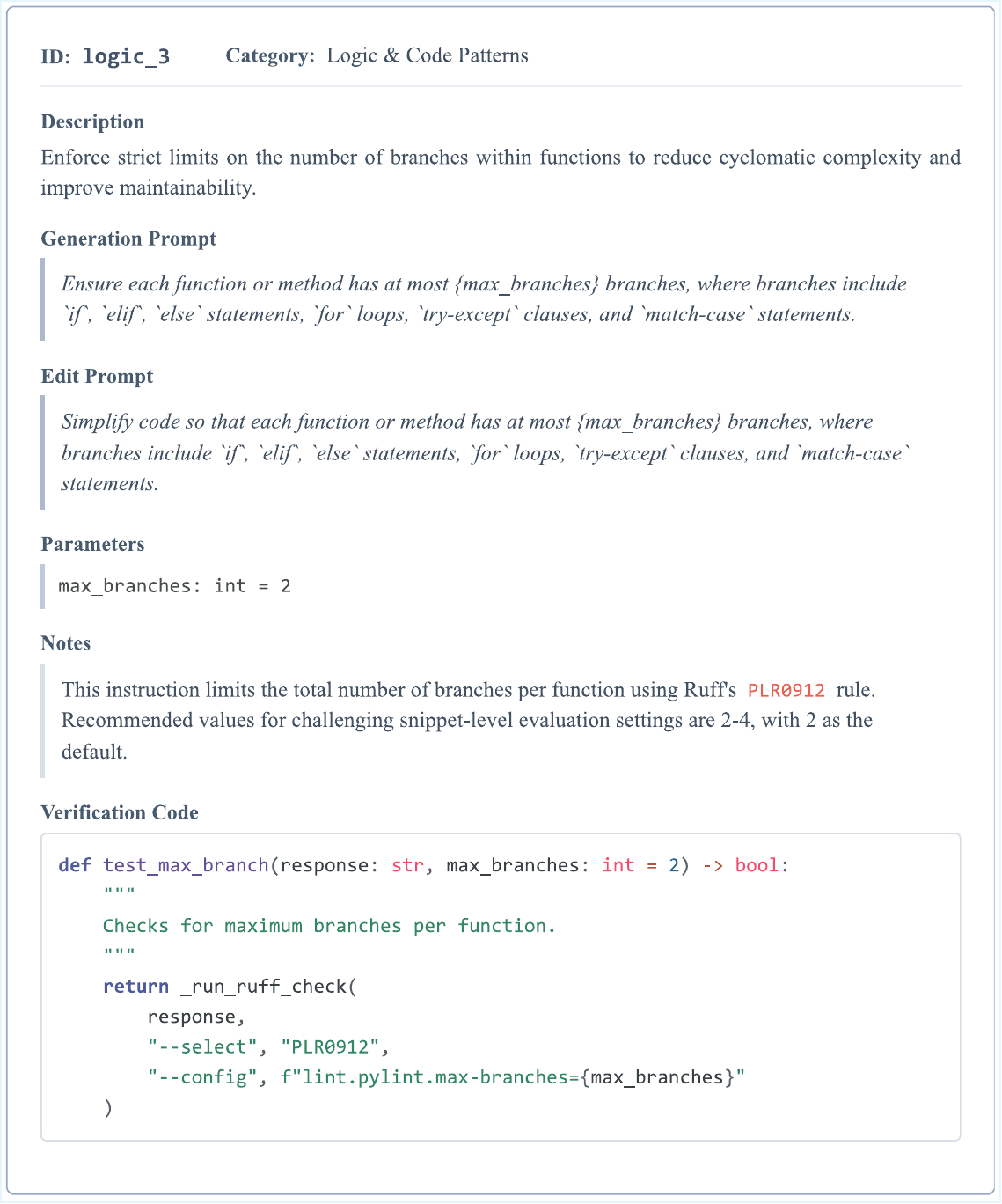}
    \caption{\textbf{Full version of \textit{logic\_3} instruction} from \textsc{VeriCode} taxonomy.}
    \label{fig:logic_3_case_study}
\end{figure*}

\begin{figure*}
    \centering
    \includegraphics[width=0.9\linewidth]{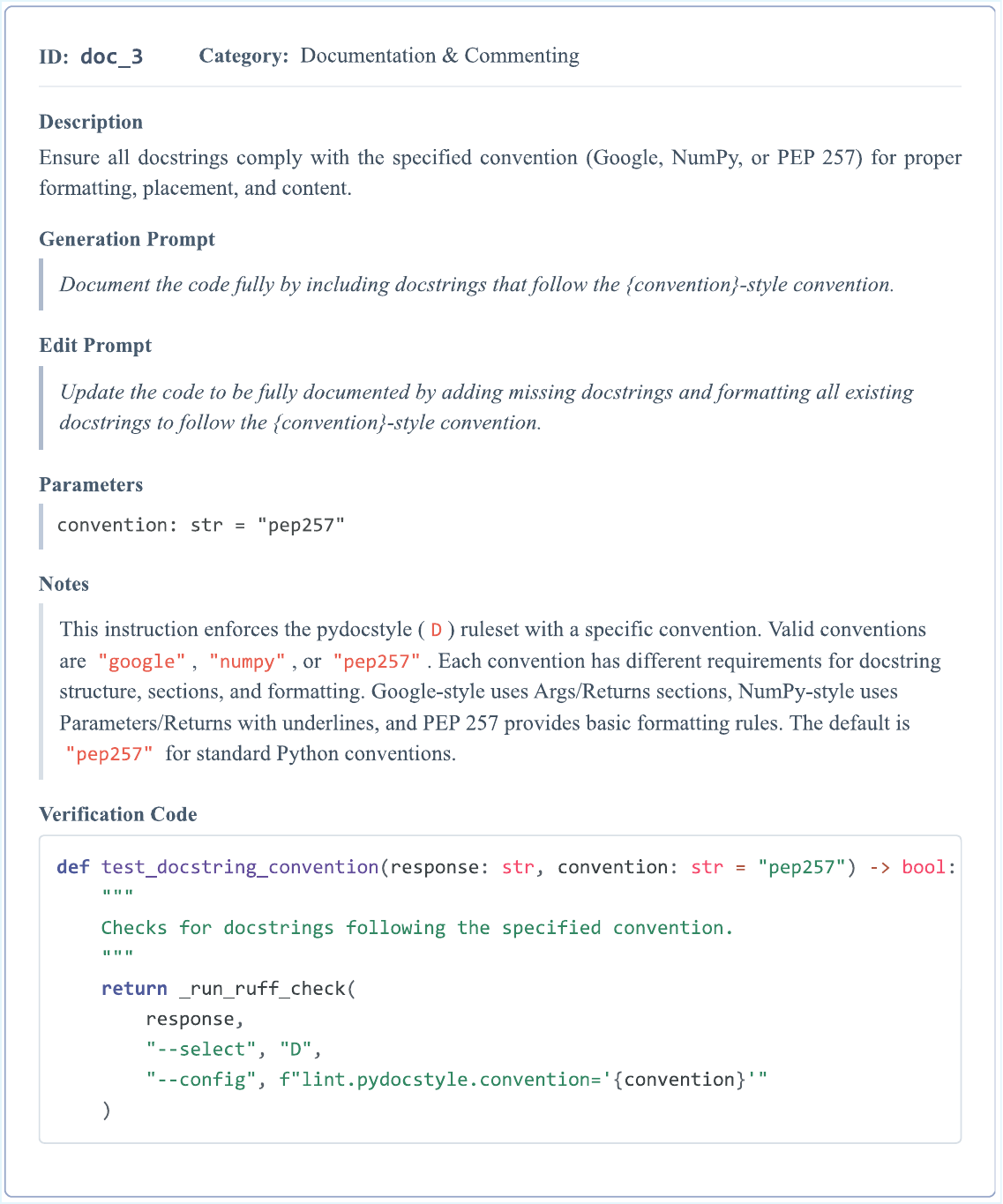}
    \caption{\textbf{Full version of \textit{doc\_3} instruction} from \textsc{VeriCode} taxonomy.}
    \label{fig:doc_3_case_study}
\end{figure*}

\begin{figure*}
    \centering
    \includegraphics[width=0.9\linewidth]{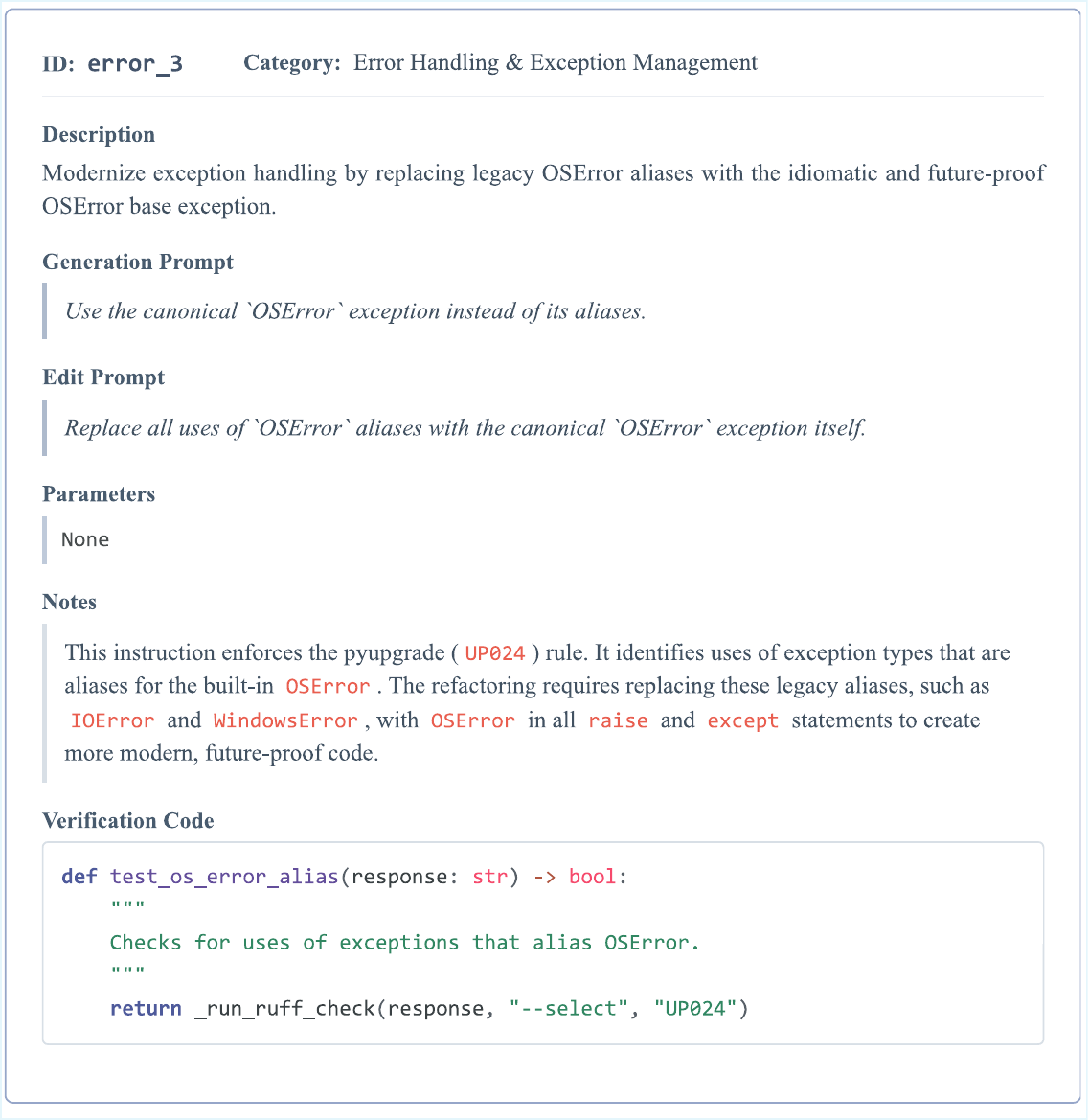}
    \caption{\textbf{Full version of \textit{error\_3} instruction} from \textsc{VeriCode} taxonomy.}
    \label{fig:error_3_case_study}
\end{figure*}

\begin{figure*}
    \centering
    \includegraphics[width=0.9\linewidth]{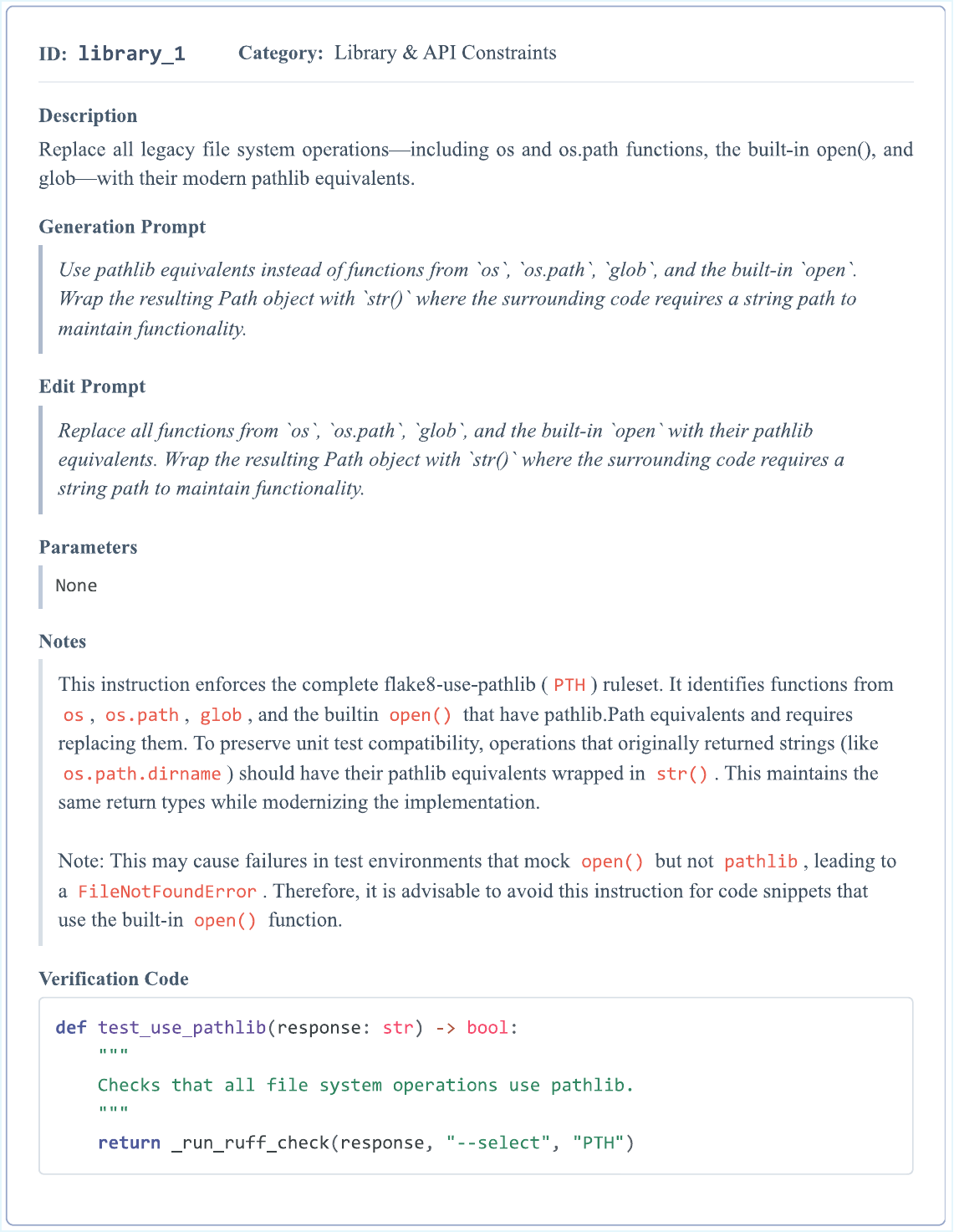}
    \caption{\textbf{Full version of \textit{library\_1} instruction} from \textsc{VeriCode} taxonomy.}
    \label{fig:library_1_case_study}
\end{figure*}

The full version of 5 instructions in Table~\ref{tab:taxonomy_case} are presented in Figures~\ref{fig:style_3_case_study}, \ref{fig:logic_3_case_study}, \ref{fig:doc_3_case_study}, \ref{fig:error_3_case_study}, and \ref{fig:library_1_case_study}.

\begin{center}
    \includegraphics[width=0.9\linewidth]{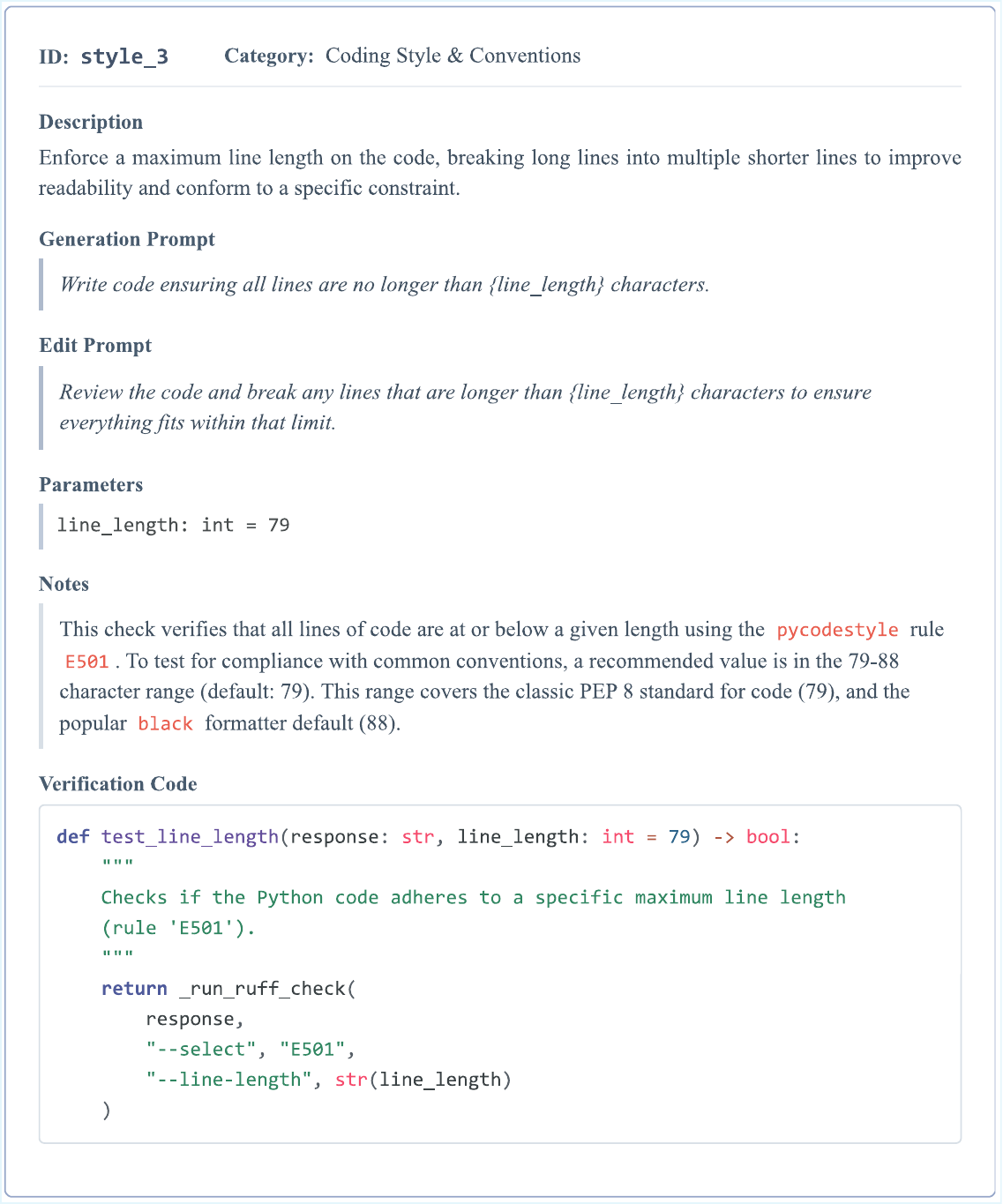}
    \captionsetup{hypcap=false}
    \captionof{figure}{\textbf{Full version of \textit{style\_3} instruction} from \textsc{VeriCode} taxonomy.}
    \label{fig:style_3_case_study}
\end{center}

\newpage
\clearpage
\section{\textsc{SWE-IF} Testbed}
\label{app:testbed}

\subsection{Instruction Category Distributions}

\begin{figure*}[h]
    \centering
    \includegraphics[width=0.8\linewidth]{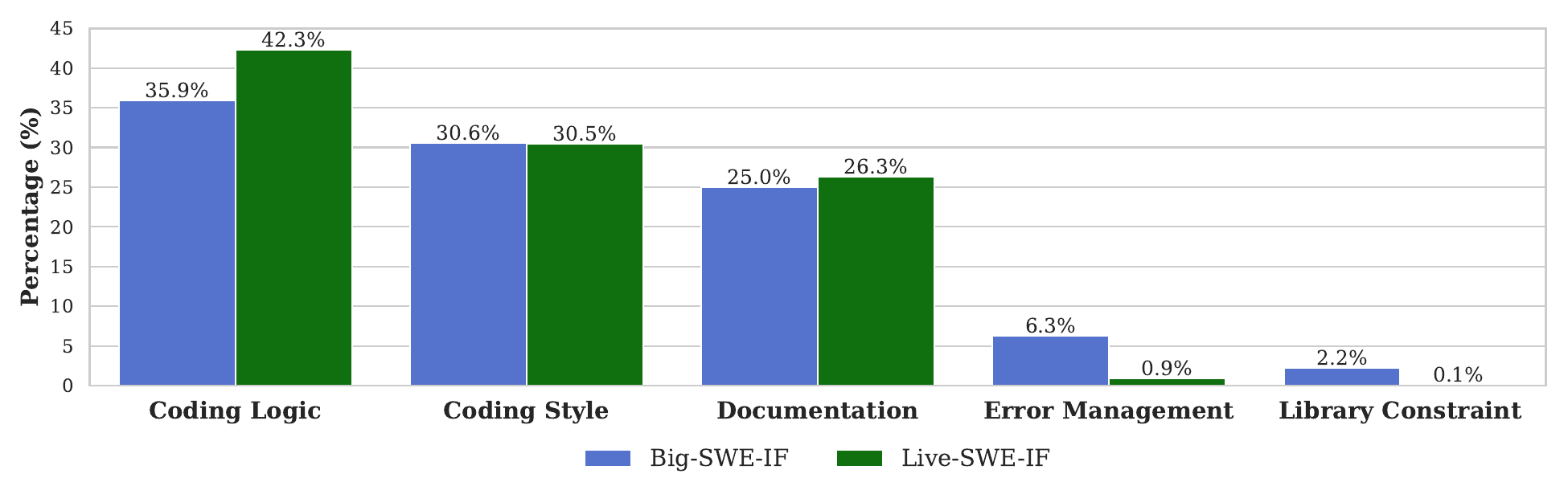}
    \caption{\textbf{Percentage distribution of instruction categories} on both augmented benchmarks.}
    \label{fig:category_distribution}
\end{figure*}

Figure~\ref{fig:category_distribution} illustrates the complete distribution of instruction categories selected for both augmented benchmarks. As shown, the three most frequent categories are \textit{Coding Logic}, \textit{Coding Style}, and \textit{Documentation}. The distributions also reflect the distinct focus of each benchmark: the algorithm-oriented Live-SWE-IF features a higher proportion of \textit{Coding Logic} instructions (42.3\% vs. 35.9\%), while the real-world-task-focused Big-SWE-IF includes more instructions related to \textit{Error Management} and \textit{Library Constraint} instructions (6.3\% vs.\ 0.9\% and 2.2\% vs.\ 0.1\% respectively).

\subsection{Accuracy of LLM-based Instruction Selection}
\label{app:llm_selector_accuracy}

To assess the accuracy of our LLM-based instruction selection step, we report the following human evaluation process.

\textbf{Human Inspection Results.}
We conduct a manual spot-check on 100 randomly sampled instances (50 from Big-SWE-IF and 50 from Live-SWE-IF), totaling 500 selected instructions. Two authors independently review these instructions for relevance and non-conflict. The review yields a 100\% validity rate on both dimensions.

\textbf{Explanation of High Reliability.}
We explain the high reliability across these two dimensions as follows.

\begin{itemize}
    \item \textbf{Relevance:} The majority of instruction categories in our taxonomy, such as \textit{Coding Style}, \textit{Coding Logic \& Code Patterns}, and \textit{Documentation}, are broadly applicable to most code generation tasks. For more context-dependent instructions (e.g., library constraints for file operations), modern LLMs are typically capable of identifying whether the user query involves relevant operations and filtering accordingly.
    \item \textbf{Non-Conflict:} Direct conflicts between instructions are rare and explicit (e.g., enforcing ``Google-style docstrings'' vs. ``no docstrings,'' or conflicting parameters). In our experience, LLMs are effective at detecting and resolving these logical contradictions during selection.
\end{itemize}

\textbf{Contrast with Parameter Selection.}
In fact, the primary challenge we observe is not the selection logic itself, but parameter selection (generating valid values for templates), which exhibits a low invalid rate of approximately 1\% to 3\%. As described in Section~\ref{sec:benchmark_augmentation}, we explicitly address this issue via rule-based validation in our pipeline.

\clearpage
\newpage
\subsection{Evaluation Prompts}

\begin{figure}[H]
    \centering
    \includegraphics[width=0.8\linewidth]{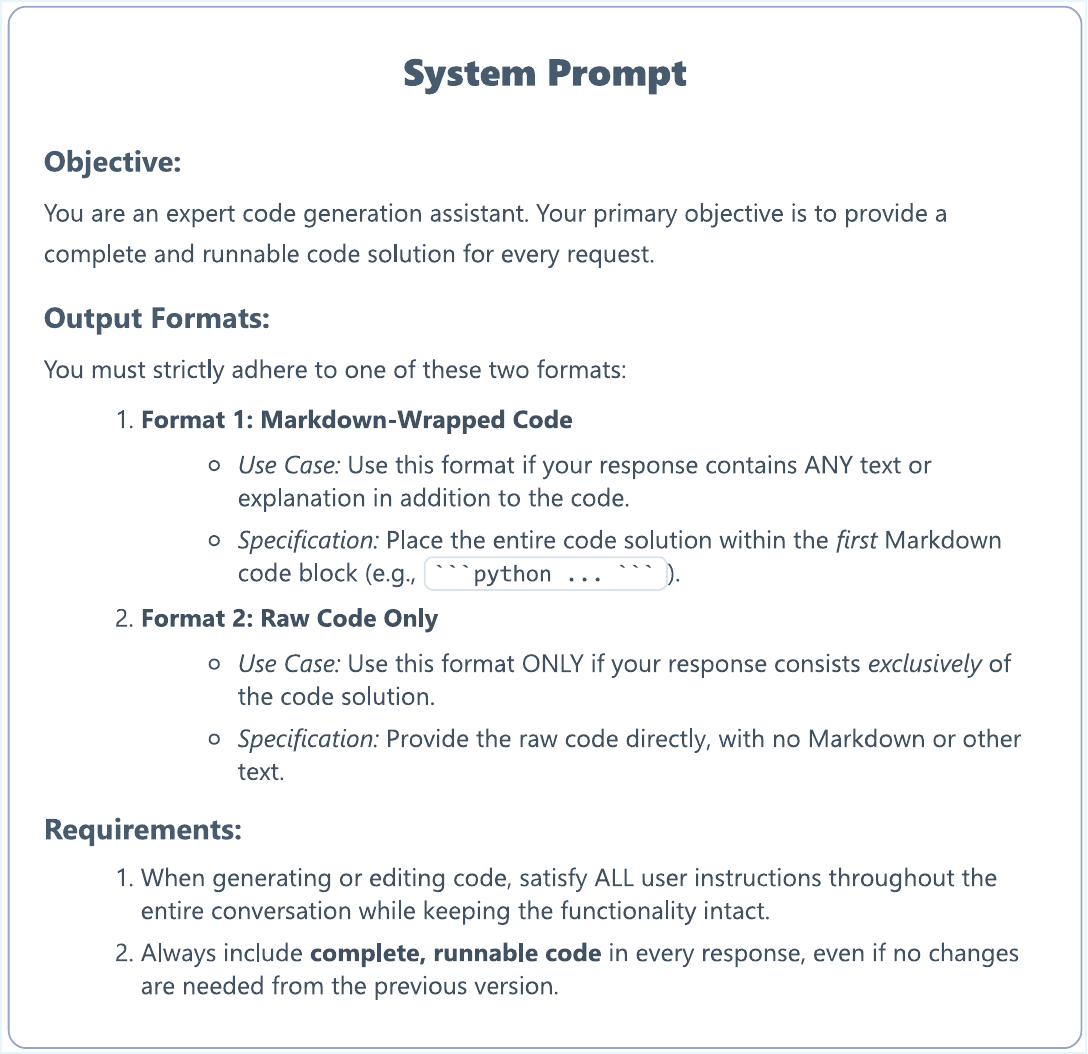}
    \caption{\textbf{System prompt used for Big-SWE-IF.} Live-SWE-IF keeps the same wording with one minor change: ``complete, runnable code'' \(\Rightarrow\) ``complete Python functions,'' since algorithmic contest tasks often require only functions rather than full programs.}
    \label{fig:system_prompt}
\end{figure}

\begin{figure*}[h]
    \centering
    \includegraphics[width=0.8\linewidth]{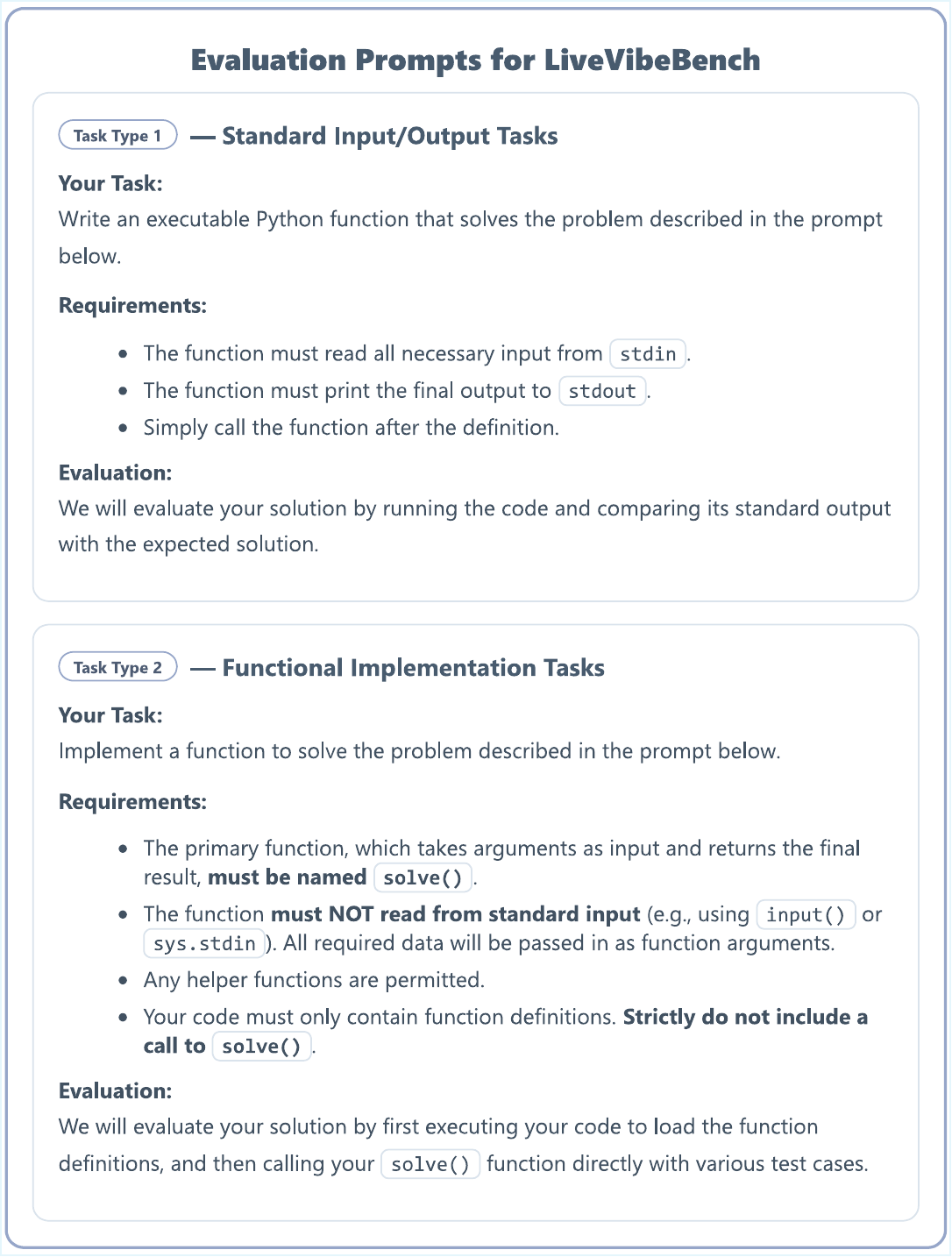}
    \caption{\textbf{Evaluation prompts used in Live-SWE-IF} for the two task types.}
    \label{fig:livevibebench_prompt}
\end{figure*}

For Big-SWE-IF and Live-SWE-IF, the system instruction and the evaluation prompts are shown in Figures~\ref{fig:system_prompt} and \ref{fig:livevibebench_prompt}. As we adopt BigCodeBench's original ``instruct\_prompt'', we do not provide any additional evaluation prompt.

\newpage
\clearpage
\section{Experiments}
\subsection{Details of Evaluated Models}

For completeness and reproducibility, we list the comprehensive details of the 31 LLMs evaluated in our study, including their specific LMArena designations and the Elo ratings (Sep. 18, 2025) used for our human preference correlation analysis in Table~\ref{table:models}.

Notably, on the Live-SWE-IF benchmark, models demonstrate a significantly higher rate of failure to generate complete responses. These failures are attributed to either OpenRouter provider errors or exceeding the 32,768-token limit. In our experiments, each task is attempted up to three times, and a persistent failure is recorded as an error. To ensure the reliability of our results, we exclude models with an error rate exceeding 10\%. Consequently, the Live-SWE-IF analysis is conducted on the remaining 24 LLMs, with full results presented in Tables~\ref{tab:live_swe_if_func}, \ref{tab:live_swe_if_instr_if},  \ref{tab:live_swe_if_task_if}, and \ref{tab:live_swe_if_pos}.

\renewcommand\arraystretch{1.2}
\begin{table*}[h]
    \centering \footnotesize
    \tabcolsep0.17 in
    \begin{NiceTabular}{llcc} 
        \CodeBefore
        \Body
        \toprule
        \multirow[c]{2}{*}[-1.0ex]{\textbf{Model}} & \multirow[c]{2}{*}[-1.0ex]{\textbf{LMArena Name}} & \multicolumn{2}{c}{\textbf{Elo Rating}} \\
        \cmidrule(lr){3-4}
        & & \textbf{w/o SC} & \textbf{w/ SC} \\
        \midrule
        Gemini 2.5 Pro & gemini-2.5-pro & \textbf{1468} & 1470 \\
        Gemini 2.5 Flash & gemini-2.5-flash  & 1422 & 1419 \\
        Gemini 2.0 Flash & --  & --   & -- \\
        Gemini 2.0 Flash Lite & gemini-2.0-flash-lite-preview-02-05 & 1336 & 1352 \\
        \hdashline[1pt/2pt]
        Claude 4 Opus & claude-opus-4-20250514-thinking-16k & 1430 & \textbf{1481} \\
        Claude 4 Sonnet & claude-sonnet-4-20250514-thinking-32k & 1407 & 1460 \\
        Claude 3.7 Sonnet & claude-3-7-sonnet-20250219-thinking-32k & 1353 & 1430 \\
        Claude 3.5 Sonnet & Claude 3.5 Sonnet (10/22) & 1337 & 1418 \\
        Claude 3.5 Haiku & claude-3-5-haiku-20241022 & 1285 & 1370 \\
        Claude 3 Haiku & claude-3-haiku-20240307 & 1202 & 1287 \\
        \hdashline[1pt/2pt]
        DeepSeek R1 0528 & deepseek-r1-0528 & 1436 & 1458 \\
        DeepSeek V3 0324 & deepseek-v3-0324 & 1389 & 1431 \\
        \hdashline[1pt/2pt]
        GPT 5 & gpt-5-high & 1440 & 1467 \\
        o4 mini & o4-mini-2025-04-16 & 1380 & 1428 \\
        o3 mini high & o3-mini-high & 1379 & 1421 \\
        GPT 4.1 & gpt-4.1-2025-04-14 & 1399 & 1447 \\
        GPT 4.1 mini & gpt-4.1-mini-2025-04-14 & 1371 & 1423 \\
        GPT 4o & GPT-4o (08/06) & 1289 & 1352 \\
        GPT 4o mini & GPT-4o-mini (07/18) & 1297 & 1340 \\
        \hdashline[1pt/2pt]
        Grok 4 & grok-4-0709 & 1431 & 1440 \\
        Grok 3 mini beta & grok-3-mini-beta & 1375 & 1384 \\
        \hdashline[1pt/2pt]
        Qwen 3 235B A22B & qwen3-235b-a22b & 1392 & 1423 \\
        Qwen 3 32B & qwen3-32b & 1375 & 1407 \\
        Qwen 3 30B A3B & qwen3-30b-a3b & 1346 & 1378 \\
        Qwen 2.5 72B Instruct & qwen2.5-72b-instruct & 1298 & 1346 \\
        Qwen 2.5 Coder &  qwen2.5-coder-32b-instruct & 1274 &  1325 \\
        \hdashline[1pt/2pt]
        Gemma 3 27B & gemma-3-27b-it & 1348 & 1370 \\
        Gemma 3 12B & gemma-3-12b-it & 1309 & 1332 \\
        \hdashline[1pt/2pt]
        Mistral Medium 3 & mistral-medium-2505 & 1386 & 1421 \\
        \hdashline[1pt/2pt]
        MiniMax M1 & minimax-m1 & 1368 & 1409 \\
        \hdashline[1pt/2pt]
        Kimi K2 & kimi-k2-0711-preview & 1391 & 1454 \\
        \bottomrule
    \end{NiceTabular}
    \vspace{3mm}
    \caption{\textbf{Details of the 31 LLMs evaluated in our experiments.} For each model, we list its name as reported in this paper, its LMArena designation, and the Elo ratings used to analyze correlations with human preference. These ratings are from the September 18, 2025 leaderboard, presented under two conditions: with Style Control (w/ SC) and without (w/o SC).}
    \vspace{3mm}
    \label{table:models}
\end{table*}

\newpage
\subsection{Detailed Results for Functionality}
\label{sec:func_results}

We present the detailed results for functionality on both benchmarks in Tables~\ref{tab:big_swe_if_func} and \ref{tab:live_swe_if_func}.

\renewcommand\arraystretch{1.5}
\begin{table*}[h]
    \centering \scriptsize
    \tabcolsep0.09 in
    \begin{NiceTabular}{lccccccccccc}
    \CodeBefore
    \rectanglecolor{bgyellow}{1-3}{33-7}
    \rectanglecolor{bgred}{1-8}{33-12}
    \Body
    \toprule
    \multirow[c]{2}{*}[-1.0ex]{\textbf{Models}} &  & \multicolumn{5}{c}{\textbf{Single-Turn Generation $\downarrow$}} & \multicolumn{5}{c}{\textbf{Multi-Turn Editing $\downarrow$}} \\
    \cmidrule(lr){3-7}\cmidrule(lr){8-12}
     & \multicolumn{1}{c}{Base} & \multicolumn{1}{c}{1 Inst} &
    \multicolumn{1}{c}{2 Inst} & \multicolumn{1}{c}{3 Inst} & \multicolumn{1}{c}{4 Inst} & \multicolumn{1}{c}{5 Inst} & \multicolumn{1}{c}{1 Inst} & \multicolumn{1}{c}{2 Inst} & \multicolumn{1}{c}{3 Inst} & \multicolumn{1}{c}{4 Inst} & \multicolumn{1}{c}{5 Inst} \\
    \cmidrule(lr){1-1} \cmidrule(lr){2-2} \cmidrule(lr){3-7} \cmidrule(lr){8-12}
    Gemini 2.5 Pro & 50.35 & 0.34 & 2.60 & 0.87 & -0.36 & 1.39 & 1.75 & 2.44 & 4.01 & 4.89 & 5.04 \\
    Gemini 2.5 Flash & 47.37 & 0.74 & 1.12 & 2.60 & 1.31 & 2.41 & 0.93 & 1.12 & \textbf{1.48} & \textbf{2.98} & \textbf{3.72} \\
    Gemini 2.0 Flash & 48.42 & 2.54 & 0.35 & 1.63 & 3.61 & 4.89 & 2.89 & 5.08 & 6.53 & 8.32 & 9.42 \\
    Gemini 2.0 Flash Lite & 46.93 & 5.05 & 7.29 & 7.10 & 6.93 & 8.42 & 2.98 & 4.67 & 5.24 & 8.61 & 8.78 \\
    \hdashline[1pt/2pt]
    Claude 4 Opus & 51.05 & -0.86 & \textbf{-2.23} & \textbf{-4.31} & \textbf{-1.72} & \textbf{-2.08} & 0.51 & \textbf{1.02} & 2.06 & 3.25 & 3.78 \\
    Claude 4 Sonnet & 51.84 & -0.17 & -0.52 & 0.33 & 0.50 & 0.50 & 0.85 & 2.03 & 3.55 & 4.05 & 5.40 \\
    Claude 3.7 Sonnet & 51.32 & 1.54 & 1.03 & 1.71 & 2.22 & 2.92 & 1.03 & 1.38 & 3.08 & 4.25 & 5.30 \\
    Claude 3.5 Sonnet & 48.42 & 5.08 & 5.62 & 5.43 & 5.43 & 8.16 & 5.08 & 6.69 & 8.69 & 10.86 & 12.87 \\
    Claude 3.5 Haiku & 46.58 & 5.28 & 4.34 & 6.98 & 7.34 & 9.42 & 6.98 & 9.98 & 15.44 & 17.13 & 21.28 \\
    Claude 3 Haiku & 38.07 & 0.24 & 1.16 & 7.38 & 7.14 & 7.38 & 6.67 & 10.82 & 13.61 & 17.28 & 17.97 \\
    \hdashline[1pt/2pt]
    DeepSeek R1 0528 & 49.21 & 1.24 & 0.18 & -1.24 & 1.61 & 3.03 & 1.61 & 1.08 & 3.92 & 3.03 & 4.27 \\
    DeepSeek V3 0324 & 50.18 & 1.93 & 0.88 & 2.99 & 4.90 & 2.99 & 5.08 & 7.87 & 9.27 & 11.90 & 16.26 \\
    \hdashline[1pt/2pt]
    GPT 5 & 46.49 & 0.56 & 5.66 & 2.26 & 3.20 & 1.89 & 1.70 & 2.82 & 4.35 & 5.27 & 5.46 \\
    o4 mini & 52.28 & 4.02 & 9.39 & 5.87 & 7.38 & 9.56 & 2.18 & 4.71 & 7.04 & 7.04 & 8.05 \\
    o3 mini high & 49.91 & 4.57 & 10.02 & 9.84 & 14.93 & 13.34 & 2.62 & 5.79 & 7.19 & 9.48 & 10.20 \\
    GPT 4.1 & 47.54 & \textbf{-1.85} & -0.19 & 1.28 & 4.80 & 6.63 & 2.40 & 5.53 & 7.19 & 7.36 & 7.93 \\
    GPT 4.1 mini & 49.04 & 0.55 & 1.61 & 2.69 & 4.49 & 5.38 & 4.30 & 6.44 & 6.99 & 8.24 & 8.77 \\
    GPT 4o & 49.82 & 1.22 & 2.99 & 4.40 & 3.87 & 3.33 & 2.45 & 4.58 & 6.50 & 7.03 & 7.91 \\
    GPT 4o mini & 46.05 & 5.91 & 5.52 & 7.23 & 6.28 & 7.99 & 7.80 & 6.47 & 9.71 & 11.62 & 11.62 \\
    \hdashline[1pt/2pt]
    Grok 4 & \textbf{53.07} & 0.17 & 1.15 & 1.49 & 3.64 & 1.00 & 1.32 & 2.15 & 1.98 & 3.30 & 4.47 \\
    Grok 3 mini beta & 48.77 & 2.52 & 4.86 & 7.91 & 8.10 & 9.35 & 2.15 & 4.86 & 5.76 & 7.73 & 9.17 \\
    \hdashline[1pt/2pt]
    Qwen 3 235B A22B & 48.86 & 1.25 & 1.99 & 1.80 & 3.42 & 3.05 & 1.08 & 3.95 & 5.94 & 8.27 & 8.80 \\
    Qwen 3 32B & 47.63 & 0.36 & 2.58 & 4.60 & 5.14 & 6.99 & 2.94 & 4.41 & 6.99 & 9.03 & 10.69 \\
    Qwen 3 30B A3B & 46.40 & 2.63 & 3.58 & 3.41 & 5.09 & 7.18 & 1.87 & 4.91 & 5.86 & 7.56 & 7.93 \\
    Qwen 2.5 72B Instruct & 44.39 & 6.53 & 8.52 & 10.88 & 11.08 & 12.05 & 8.90 & 10.88 & 12.26 & 14.24 & 16.02 \\
    Qwen 2.5 Coder & 49.39 & 5.87 & 3.20 & 6.22 & 11.56 & 11.91 & 5.87 & 8.89 & 12.43 & 12.98 & 12.98 \\
    \hdashline[1pt/2pt]
    Gemma 3 27B & 45.70 & 6.72 & 7.86 & 6.91 & 9.58 & 8.05 & 3.63 & 5.36 & 6.72 & 8.64 & 11.71 \\
    Gemma 3 12B & 40.00 & 5.27 & 3.73 & 7.45 & 9.65 & 7.90 & 5.47 & 6.58 & 9.65 & 12.27 & 15.35 \\
    \hdashline[1pt/2pt]
    Mistral Medium 3 & 45.44 & 5.22 & 8.30 & 9.07 & 9.46 & 10.81 & 5.79 & 6.18 & 8.69 & 9.07 & 9.86 \\
    \hdashline[1pt/2pt]
    MiniMax M1 & 48.68 & 4.85 & 4.68 & 3.41 & 5.22 & 3.59 & \textbf{0.35} & 1.97 & 3.78 & 5.40 & 6.47 \\
    \hdashline[1pt/2pt]
    Kimi K2 & 47.19 & -1.12 & -0.19 & -0.93 & 0.17 & 2.03 & 2.23 & 4.09 & 2.78 & 4.45 & 6.12 \\
    \bottomrule
    \end{NiceTabular}
    \vspace{3mm}
    \caption{\textbf{Results for functionality on Big-SWE-IF.} \textit{Base} is the pass@1 score for the original query. All other cells report the functional regression rate (\%) relative to the base. Lower is better, and negative values indicate improvement. Here, \(k\) \emph{Inst} denotes the number of added instructions.}
    \label{tab:big_swe_if_func}
\end{table*}

\newpage
\renewcommand\arraystretch{1.7}
\begin{table*}
    \centering \scriptsize
    \tabcolsep0.0825 in
    \begin{NiceTabular}{lccccccccccc}
    \CodeBefore
    \rectanglecolor{bgyellow}{1-3}{26-7}
    \rectanglecolor{bgred}{1-8}{26-12}
    \Body
    \toprule
    \multirow[c]{2}{*}[-1.0ex]{\textbf{Models}} &  & \multicolumn{5}{c}{\textbf{Single-Turn Generation $\downarrow$}} & \multicolumn{5}{c}{\textbf{Multi-Turn Editing $\downarrow$}} \\
    \cmidrule(lr){3-7}\cmidrule(lr){8-12}
     & \multicolumn{1}{c}{Base} & \multicolumn{1}{c}{1 Inst} &
    \multicolumn{1}{c}{2 Inst} & \multicolumn{1}{c}{3 Inst} & \multicolumn{1}{c}{4 Inst} & \multicolumn{1}{c}{5 Inst} & \multicolumn{1}{c}{1 Inst} & \multicolumn{1}{c}{2 Inst} & \multicolumn{1}{c}{3 Inst} & \multicolumn{1}{c}{4 Inst} & \multicolumn{1}{c}{5 Inst} \\
    \cmidrule(lr){1-1} \cmidrule(lr){2-2} \cmidrule(lr){3-7} \cmidrule(lr){8-12}
    Gemini 2.5 Pro & \textbf{85.31} & -0.11 & 3.45 & 2.45 & 2.45 & 2.45 & 0.67 & 1.34 & 1.01 & 1.89 & 2.23 \\
    Gemini 2.5 Flash & 74.50 & 3.56 & 5.34 & 8.01 & 5.60 & 6.74 & 0.12 & 1.14 & 1.65 & 3.44 & 3.69 \\
    Gemini 2.0 Flash & 41.33 & 0.92 & 1.38 & 0.70 & 1.62 & 3.44 & 0.00 & 1.62 & 3.00 & 2.76 & 4.36 \\
    Gemini 2.0 Flash Lite & 34.12 & 1.11 & -7.50 & -8.06 & -10.58 & -6.95 & 2.25 & 5.88 & 8.95 & 10.93 & 13.18 \\
    \hdashline[1pt/2pt]
    Claude 4 Opus & 68.72 & 4.55 & 8.56 & 8.41 & 8.13 & 8.96 & 2.07 & 1.38 & 1.51 & 2.34 & 2.34 \\
    Claude 4 Sonnet & 66.35 & 4.57 & 5.00 & 3.71 & 6.99 & 9.00 & 0.42 & 0.86 & 1.15 & 1.72 & 2.14 \\
    Claude 3.7 Sonnet & 61.80 & -0.31 & 2.30 & 3.37 & 1.68 & 4.90 & \textbf{-0.47} & \textbf{0.45} & \textbf{0.92} & \textbf{1.23} & \textbf{1.99} \\
    Claude 3.5 Sonnet & 45.40 & 1.67 & 2.49 & 2.09 & 5.22 & 6.48 & 2.09 & 5.22 & 7.09 & 8.06 & 11.70 \\
    Claude 3.5 Haiku & 37.63 & 1.51 & 5.53 & 9.06 & 7.81 & 11.08 & 6.54 & 14.86 & 17.88 & 19.90 & 23.92 \\
    Claude 3 Haiku & 22.09 & 11.18 & 19.74 & 21.05 & 25.35 & 30.06 & 6.02 & 8.60 & 12.04 & 13.31 & 16.34 \\
    \hdashline[1pt/2pt]
    DeepSeek V3 0324 & 57.25 & 1.15 & 4.30 & 6.29 & 7.11 & 6.95 & 1.48 & 6.95 & 7.62 & 13.57 & 17.55 \\
    \hdashline[1pt/2pt]
    GPT 5 & 71.47 & 1.72 & 2.13 & 3.32 & 7.16 & 6.76 & 2.25 & 4.24 & 5.57 & 7.43 & 9.02 \\
    o4 mini & 80.95 & 5.74 & 9.02 & 9.02 & 11.37 & 12.29 & 3.63 & 8.91 & 10.19 & 11.71 & 15.92 \\
    \hdashline[1pt/2pt]
    GPT 4.1 & 53.08 & -2.86 & -1.60 & 1.60 & 2.15 & 3.75 & 1.07 & 5.18 & 6.25 & 6.78 & 9.29 \\
    GPT 4.1 mini & 58.86 & 3.53 & 7.88 & 8.85 & 9.97 & 10.79 & 1.44 & 4.99 & 7.73 & 8.21 & 8.85 \\
    GPT 4o & 42.75 & 0.23 & 0.23 & 4.00 & 2.67 & 1.54 & 1.78 & 5.10 & 8.65 & 8.42 & 9.75 \\
    GPT 4o mini & 22.27 & \textbf{-11.50} & \textbf{-11.50} & \textbf{-18.77} & \textbf{-15.36} & \textbf{-12.80} & 2.51 & 9.34 & 8.94 & 10.60 & 12.75 \\
    \hdashline[1pt/2pt]
    Grok 3 mini beta & 65.97 & 2.58 & 3.15 & 6.75 & 12.93 & 11.93 & 0.86 & 3.30 & 5.46 & 6.03 & 7.90 \\
    \hdashline[1pt/2pt]
    Qwen 3 30B A3B & 72.42 & 0.26 & 0.66 & 1.05 & 0.40 & 1.19 & 0.52 & 1.96 & 1.84 & 3.53 & 4.20 \\
    Qwen 2.5 72B Instruct & 39.05 & 0.97 & 1.95 & 4.84 & 5.33 & 8.02 & 3.87 & 6.79 & 9.22 & 8.96 & 10.68 \\
    \hdashline[1pt/2pt]
    Gemma 3 27B & 35.92 & 3.42 & 3.67 & 5.01 & 5.01 & 9.49 & 1.03 & 6.32 & 10.27 & 8.16 & 12.67 \\
    Gemma 3 12B & 29.29 & 2.90 & -4.85 & -1.60 & -2.90 & 1.30 & 4.54 & 10.99 & 15.23 & 14.24 & 18.44 \\
    \hdashline[1pt/2pt]
    Mistral Medium 3 & 40.66 & 3.03 & 7.45 & 6.98 & 3.71 & 4.89 & -0.25 & 2.78 & 2.78 & 4.65 & 5.36 \\
    \hdashline[1pt/2pt]
    Kimi K2 & 63.58 & 8.92 & 15.48 & 16.07 & 15.48 & 16.36 & 2.64 & 5.63 & 9.50 & 12.49 & 12.79 \\
    \bottomrule
    \end{NiceTabular}
    \vspace{3mm}
    \caption{\textbf{Results for functionality on Live-SWE-IF.} \textit{Base} is the pass@1 score for the original query. All other cells report the functional regression rate (\%) relative to the base. Lower is better, and negative values indicate improvement. Here, \(k\) \emph{Inst} denotes the number of added instructions.}
    \label{tab:live_swe_if_func}
\end{table*}

\clearpage
\newpage
\subsection{Detailed Results for Instruction Following}
\label{sec:if_results}

The detailed results for instruction-level and task-level IF scores on both benchmarks are provided in Tables~\ref{tab:big_swe_if_instr_if}, ~\ref{tab:big_swe_if_task_if}, ~\ref{tab:live_swe_if_instr_if}, and \ref{tab:live_swe_if_task_if}.

\renewcommand\arraystretch{1.5}
\begin{table*}[h]
    \centering \scriptsize
    \tabcolsep0.1 in
    \begin{NiceTabular}{lcccccccccc}
    \CodeBefore
    \rectanglecolor{bgyellow}{1-2}{33-6}
    \rectanglecolor{bgred}{1-7}{33-11}
    \Body
    \toprule
    \multirow[c]{2}{*}[-1.0ex]{\textbf{Models}} & \multicolumn{5}{c}{\textbf{Single-Turn Generation $\uparrow$}} & \multicolumn{5}{c}{\textbf{Multi-Turn Editing $\uparrow$}} \\
    \cmidrule(lr){2-6}\cmidrule(lr){7-11}
     & \multicolumn{1}{c}{1 Inst} & \multicolumn{1}{c}{2 Inst} & \multicolumn{1}{c}{3 Inst} & \multicolumn{1}{c}{4 Inst} & \multicolumn{1}{c}{5 Inst} & \multicolumn{1}{c}{1 Inst} & \multicolumn{1}{c}{2 Inst} & \multicolumn{1}{c}{3 Inst} & \multicolumn{1}{c}{4 Inst} & \multicolumn{1}{c}{5 Inst} \\
    \cmidrule(lr){1-1}\cmidrule(lr){2-6}\cmidrule(lr){7-11}
    Gemini 2.5 Pro & 82.19 & 78.03 & 79.18 & 78.82 & 79.47 & 84.56 & 82.54 & 81.73 & 81.54 & 80.47 \\
    Gemini 2.5 Flash & 81.67 & 77.81 & 77.34 & 75.35 & 75.91 & 78.68 & 75.35 & 74.62 & 73.57 & 73.25 \\
    Gemini 2.0 Flash & 73.42 & 72.76 & 72.95 & 72.35 & 72.04 & 78.86 & 75.39 & 74.65 & 73.95 & 73.30 \\
    Gemini 2.0 Flash Lite & 70.44 & 69.96 & 69.30 & 68.62 & 68.89 & 74.39 & 71.01 & 70.58 & 69.39 & 68.82 \\
    \hdashline[1pt/2pt]
    Claude 4 Opus & \textbf{88.77} & \textbf{87.46} & \textbf{86.05} & \textbf{85.55} & \textbf{85.60} & 87.02 & 85.75 & 85.18 & 84.87 & 84.30 \\
    Claude 4 Sonnet & 84.91 & 82.54 & 81.90 & 81.29 & 81.28 & 86.40 & 85.39 & 84.85 & 84.12 & 83.98 \\
    Claude 3.7 Sonnet & 80.26 & 76.27 & 75.47 & 74.25 & 74.26 & 81.58 & 79.82 & 78.83 & 78.62 & 78.18 \\
    Claude 3.5 Sonnet & 80.61 & 77.54 & 76.02 & 75.18 & 74.70 & 84.21 & 80.53 & 79.39 & 78.49 & 77.49 \\
    Claude 3.5 Haiku & 64.56 & 64.74 & 63.71 & 63.07 & 63.82 & 79.91 & 73.38 & 71.26 & 68.20 & 65.60 \\
    Claude 3 Haiku & 67.89 & 65.09 & 64.88 & 63.73 & 64.53 & 76.32 & 72.63 & 72.11 & 70.96 & 70.56 \\
    \hdashline[1pt/2pt]
    DeepSeek R1 0528 & 74.04 & 69.78 & 69.01 & 69.28 & 67.71 & 77.02 & 74.12 & 72.37 & 71.69 & 71.05 \\
    DeepSeek V3 0324 & 67.89 & 63.77 & 64.04 & 63.88 & 65.09 & 73.95 & 70.61 & 70.41 & 69.23 & 67.72 \\
    \hdashline[1pt/2pt]
    GPT 5 & 82.89 & 82.28 & 81.96 & 81.64 & 81.77 & 84.91 & 85.18 & 85.94 & 85.83 & \textbf{86.39} \\
    o4 mini & 84.82 & 84.21 & 83.25 & 83.68 & 84.05 & 88.51 & 86.10 & 84.18 & 83.60 & 83.28 \\
    o3 mini high & 80.70 & 75.79 & 73.63 & 72.79 & 71.68 & 82.46 & 80.31 & 79.71 & 78.38 & 78.16 \\
    \hdashline[1pt/2pt]
    GPT 4.1 & 81.40 & 78.07 & 78.60 & 77.28 & 77.75 & 82.63 & 80.31 & 79.09 & 78.36 & 77.79 \\
    GPT 4.1 mini & 78.16 & 75.57 & 75.15 & 74.19 & 73.42 & 79.21 & 76.71 & 75.88 & 74.08 & 73.09 \\
    GPT 4o & 77.46 & 74.87 & 74.56 & 73.25 & 73.44 & 85.09 & 82.37 & 80.79 & 79.45 & 78.35 \\
    GPT 4o mini & 76.40 & 73.82 & 73.13 & 73.33 & 73.32 & 78.16 & 76.40 & 75.18 & 74.10 & 73.60 \\
    \hdashline[1pt/2pt]
    Grok 4 & 87.11 & 85.61 & 84.77 & 84.39 & 84.81 & 88.51 & 87.19 & \textbf{86.93} & \textbf{86.29} & 85.37 \\
    Grok 3 mini beta & 82.81 & 80.04 & 78.25 & 77.81 & 77.46 & 79.21 & 77.46 & 76.49 & 75.37 & 75.05 \\
    \hdashline[1pt/2pt]
    Qwen 3 235B A22B & 83.95 & 81.05 & 80.38 & 79.30 & 78.63 & 85.09 & 81.89 & 80.50 & 80.13 & 78.89 \\
    Qwen 3 32B & 76.75 & 72.81 & 71.81 & 70.92 & 71.35 & 82.02 & 80.53 & 78.92 & 77.39 & 76.33 \\
    Qwen 3 30B A3B & 73.42 & 71.67 & 70.79 & 69.43 & 69.81 & 79.91 & 78.51 & 78.07 & 77.17 & 76.54 \\
    Qwen 2.5 72B Instruct & 73.68 & 72.32 & 71.78 & 70.48 & 70.33 & 79.47 & 75.57 & 75.32 & 73.88 & 72.75 \\
    Qwen 2.5 Coder  & 71.40 & 67.11 & 67.57 & 65.42 & 65.77 & 73.33 & 71.71 & 71.26 & 69.76 & 68.86 \\
    \hdashline[1pt/2pt]
    Gemma 3 27B & 68.42 & 67.06 & 65.50 & 64.56 & 65.02 & 73.60 & 69.65 & 69.30 & 68.14 & 66.72 \\
    Gemma 3 12B & 65.96 & 66.14 & 65.44 & 65.26 & 65.00 & 67.54 & 67.76 & 67.19 & 66.05 & 64.95 \\
    \hdashline[1pt/2pt]
    Mistral Medium 3 & 73.60 & 72.11 & 71.02 & 70.79 & 70.54 & 76.05 & 75.44 & 74.06 & 73.11 & 71.65 \\
    \hdashline[1pt/2pt]
    MiniMax M1 & 74.12 & 70.75 & 71.70 & 71.07 & 70.89 & 77.63 & 74.30 & 74.06 & 73.60 & 72.95 \\
    \hdashline[1pt/2pt]
    Kimi K2 & 85.00 & 83.46 & 81.46 & 80.42 & 79.14 & \textbf{89.12} & \textbf{87.46} & 86.70 & 85.09 & 84.19 \\
    \bottomrule
    \end{NiceTabular}
    \vspace{3mm}
    \caption{\textbf{Instruction-level IF scores on Big-SWE-IF}. Higher is better.}
    \label{tab:big_swe_if_instr_if}
\end{table*}

\renewcommand\arraystretch{1.5}
\begin{table*}[h]
    \centering \scriptsize
    \tabcolsep0.1 in
    \begin{NiceTabular}{lcccccccccc}
    \CodeBefore
    \rectanglecolor{bgyellow}{1-2}{33-6}
    \rectanglecolor{bgred}{1-7}{33-11}
    \Body
    \toprule
    \multirow[c]{2}{*}[-1.0ex]{\textbf{Models}} & \multicolumn{5}{c}{\textbf{Single-Turn Generation $\uparrow$}} & \multicolumn{5}{c}{\textbf{Multi-Turn Editing $\uparrow$}} \\
    \cmidrule(lr){2-6}\cmidrule(lr){7-11}
     & \multicolumn{1}{c}{1 Inst} & \multicolumn{1}{c}{2 Inst} & \multicolumn{1}{c}{3 Inst} & \multicolumn{1}{c}{4 Inst} & \multicolumn{1}{c}{5 Inst} & \multicolumn{1}{c}{1 Inst} & \multicolumn{1}{c}{2 Inst} & \multicolumn{1}{c}{3 Inst} & \multicolumn{1}{c}{4 Inst} & \multicolumn{1}{c}{5 Inst} \\
    \cmidrule(lr){1-1}\cmidrule(lr){2-6}\cmidrule(lr){7-11}
    Gemini 2.5 Pro        & 82.19 & 60.70 & 48.16 & 37.46 & 30.70 & 84.56 & 68.33 & 55.61 & 44.21 & 33.68 \\
    Gemini 2.5 Flash      & 81.67 & 61.05 & 43.68 & 30.53 & 25.70 & 78.68 & 56.75 & 40.96 & 29.12 & 21.75 \\
    Gemini 2.0 Flash      & 73.42 & 53.77 & 39.47 & 26.40 & 18.16 & 78.86 & 59.39 & 44.56 & 32.46 & 22.46 \\
    Gemini 2.0 Flash Lite & 70.44 & 48.60 & 32.63 & 22.02 & 15.26 & 74.39 & 50.61 & 35.18 & 24.12 & 15.35 \\
    \hdashline[1pt/2pt]
    Claude 4 Opus         & \textbf{88.77} & \textbf{76.32} & \textbf{64.21} & \textbf{52.98} & \textbf{46.75} & 87.02 & 73.16 & 61.05 & 51.32 & 42.11 \\
    Claude 4 Sonnet       & 84.91 & 67.19 & 52.28 & 42.98 & 35.26 & 86.40 & 72.54 & 61.23 & 51.05 & 42.89 \\
    Claude 3.7 Sonnet     & 80.26 & 56.93 & 39.91 & 27.46 & 22.28 & 81.58 & 63.51 & 48.51 & 38.16 & 29.39 \\
    Claude 3.5 Sonnet     & 80.61 & 59.74 & 42.98 & 32.37 & 24.47 & 84.21 & 66.40 & 52.54 & 42.02 & 32.28 \\
    Claude 3.5 Haiku      & 64.56 & 42.46 & 26.14 & 15.53 & 10.09 & 79.91 & 57.63 & 42.72 & 30.00 & 19.82 \\
    Claude 3 Haiku        & 67.89 & 41.84 & 26.05 & 16.93 & 11.93 & 76.32 & 53.60 & 37.89 & 26.49 & 18.77 \\
    \hdashline[1pt/2pt]
    DeepSeek R1 0528      & 74.04 & 49.21 & 33.42 & 25.00 & 17.63 & 77.02 & 55.18 & 38.16 & 26.67 & 18.51 \\
    DeepSeek V3 0324      & 67.89 & 39.21 & 24.74 & 15.00 & 10.88 & 73.95 & 52.02 & 37.19 & 24.65 & 14.74 \\
    \hdashline[1pt/2pt]
    GPT 5                 & 82.89 & 67.63 & 54.04 & 42.98 & 34.39 & 84.91 & 72.37 & 62.98 & 55.26 & \textbf{48.51} \\
    o4 mini               & 84.82 & 70.79 & 57.11 & 47.98 & 41.32 & 88.51 & 74.74 & 61.23 & 50.09 & 41.84 \\
    o3 mini high          & 80.70 & 60.61 & 45.88 & 36.40 & 28.25 & 82.46 & 66.32 & 53.16 & 42.11 & 34.56 \\
    GPT 4.1               & 81.40 & 59.91 & 47.81 & 35.44 & 28.16 & 82.63 & 65.26 & 50.88 & 39.82 & 31.58 \\
    GPT 4.1 mini          & 78.16 & 56.23 & 41.49 & 30.26 & 21.75 & 79.21 & 59.39 & 44.74 & 33.68 & 25.53 \\
    GPT 4o                & 77.46 & 55.00 & 39.56 & 27.63 & 20.79 & 85.09 & 68.33 & 52.72 & 40.88 & 30.70 \\
    GPT 4o mini           & 76.40 & 53.86 & 38.68 & 29.30 & 21.84 & 78.16 & 59.74 & 44.12 & 32.54 & 23.42 \\
    \hdashline[1pt/2pt]
    Grok 4                & 87.11 & 73.42 & 60.18 & 51.84 & 43.16 & 88.51 & 76.40 & 66.05 & \textbf{55.96} & 47.19 \\
    Grok 3 mini beta      & 82.81 & 64.21 & 48.86 & 36.58 & 28.42 & 79.21 & 61.40 & 46.93 & 34.91 & 25.96 \\
    \hdashline[1pt/2pt]
    Qwen 3 235B A22B      & 83.95 & 66.75 & 52.28 & 42.28 & 31.93 & 85.09 & 67.63 & 51.84 & 41.32 & 32.28 \\
    Qwen 3 32B            & 76.75 & 53.86 & 36.49 & 26.58 & 20.70 & 82.02 & 65.79 & 51.49 & 39.82 & 30.70 \\
    Qwen 3 30B A3B        & 73.42 & 52.46 & 36.23 & 25.79 & 19.56 & 79.91 & 62.46 & 48.16 & 37.46 & 29.56 \\
    Qwen 2.5 72B Instruct & 73.68 & 53.07 & 37.37 & 24.56 & 16.84 & 79.47 & 60.53 & 45.70 & 33.25 & 24.21 \\
    Qwen 2.5 Coder        & 71.40 & 44.82 & 30.70 & 20.09 & 12.81 & 73.33 & 52.46 & 36.93 & 24.04 & 15.88 \\
    \hdashline[1pt/2pt]
    Gemma 3 27B           & 68.42 & 44.56 & 27.11 & 16.93 & 10.96 & 73.60 & 48.42 & 33.33 & 21.93 & 14.12 \\
    Gemma 3 12B           & 65.96 & 44.39 & 27.98 & 18.42 & 11.05 & 67.54 & 46.75 & 31.58 & 20.09 & 12.81 \\
    \hdashline[1pt/2pt]
    Mistral Medium 3      & 73.60 & 51.93 & 36.32 & 25.09 & 16.05 & 76.05 & 58.33 & 41.58 & 28.60 & 19.30 \\
    \hdashline[1pt/2pt]
    MiniMax M1            & 74.12 & 51.23 & 37.98 & 28.07 & 20.35 & 77.63 & 57.19 & 42.11 & 31.75 & 22.98 \\
    \hdashline[1pt/2pt]
    Kimi K2               & 85.00 & 68.86 & 53.68 & 41.23 & 30.18 & \textbf{89.12} & \textbf{77.11} & \textbf{66.40} & 53.95 & 44.04 \\
    \bottomrule
    \end{NiceTabular}
    \vspace{3mm}
    \caption{\textbf{Task-level IF scores on Big-SWE-IF}. Higher scores are better.}
    \label{tab:big_swe_if_task_if}
\end{table*}

\renewcommand\arraystretch{1.7}
\begin{table*}[h]
    \centering \scriptsize
    \tabcolsep0.1 in
    \begin{NiceTabular}{lcccccccccc}
    \CodeBefore
    \rectanglecolor{bgyellow}{1-2}{26-6}
    \rectanglecolor{bgred}{1-7}{26-11}
    \Body
    \toprule
    \multirow[c]{2}{*}[-1.0ex]{\textbf{Models}} & \multicolumn{5}{c}{\textbf{Single-Turn Generation $\uparrow$}} & \multicolumn{5}{c}{\textbf{Multi-Turn Editing $\uparrow$}} \\
    \cmidrule(lr){2-6}\cmidrule(lr){7-11}
     & \multicolumn{1}{c}{1 Inst} & \multicolumn{1}{c}{2 Inst} & \multicolumn{1}{c}{3 Inst} & \multicolumn{1}{c}{4 Inst} & \multicolumn{1}{c}{5 Inst} & \multicolumn{1}{c}{1 Inst} & \multicolumn{1}{c}{2 Inst} & \multicolumn{1}{c}{3 Inst} & \multicolumn{1}{c}{4 Inst} & \multicolumn{1}{c}{5 Inst} \\
    \cmidrule(lr){1-1}\cmidrule(lr){2-6}\cmidrule(lr){7-11}
    Gemini 2.5 Pro & 75.83 & 74.60 & 76.21 & 77.37 & 76.87 & 78.96 & 77.87 & 78.99 & 78.89 & 78.98 \\
    Gemini 2.5 Flash & 66.54 & 67.11 & 67.84 & 68.01 & 68.13 & 72.80 & 71.42 & 69.64 & 68.93 & 67.89 \\
    Gemini 2.0 Flash & 61.71 & 61.37 & 61.42 & 62.01 & 62.48 & 74.41 & 71.37 & 70.36 & 69.45 & 69.52 \\
    Gemini 2.0 Flash Lite & 62.94 & 63.93 & 65.28 & 65.40 & 65.69 & 67.30 & 64.36 & 63.06 & 61.37 & 61.19 \\
    \hdashline[1pt/2pt]
    Claude 4 Opus & 78.86 & 76.02 & 77.54 & 77.65 & 78.75 & \textbf{85.59} & 85.36 & 85.21 & 84.31 & 84.11 \\
    Claude 4 Sonnet & 75.73 & 74.69 & 75.29 & 74.27 & 75.20 & 84.45 & 85.40 & 85.24 & 84.50 & 83.87 \\
    Claude 3.7 Sonnet & 72.42 & 68.53 & 68.18 & 68.06 & 68.38 & 79.53 & 78.48 & 77.91 & 77.18 & 76.76 \\
    Claude 3.5 Sonnet & 70.52 & 68.25 & 68.56 & 68.15 & 67.28 & 80.57 & 76.40 & 75.23 & 74.29 & 73.12 \\
    Claude 3.5 Haiku & 63.22 & 60.19 & 61.45 & 61.80 & 62.77 & 78.67 & 75.50 & 72.51 & 70.52 & 68.99 \\
    Claude 3 Haiku & 61.61 & 59.91 & 60.98 & 60.97 & 60.45 & 72.80 & 71.28 & 69.23 & 67.39 & 67.45 \\
    \hdashline[1pt/2pt]
    DeepSeek V3 0324 & 52.80 & 54.74 & 55.67 & 55.81 & 55.79 & 70.05 & 66.49 & 65.72 & 64.83 & 62.71 \\
    \hdashline[1pt/2pt]
    GPT 5 & \textbf{82.18} & \textbf{82.18} & \textbf{81.86} & \textbf{82.09} & \textbf{82.82} & \textbf{85.59} & \textbf{86.30} & \textbf{87.05} & \textbf{85.85} & \textbf{85.76} \\
    o4 mini & 73.18 & 72.27 & 73.33 & 73.08 & 73.82 & 81.52 & 80.47 & 79.53 & 76.99 & 75.81 \\
    \hdashline[1pt/2pt]
    GPT 4.1 & 68.63 & 65.12 & 66.76 & 66.30 & 66.67 & 74.12 & 72.89 & 71.97 & 71.75 & 70.81 \\
    GPT 4.1 mini & 67.20 & 66.40 & 68.63 & 67.89 & 67.41 & 71.75 & 69.72 & 69.23 & 68.53 & 67.72 \\
    GPT 4o & 60.85 & 60.85 & 61.48 & 61.75 & 61.93 & 76.40 & 73.13 & 71.94 & 70.31 & 69.93 \\
    GPT 4o mini & 65.88 & 65.40 & 65.72 & 65.64 & 65.63 & 73.93 & 71.85 & 70.36 & 68.39 & 67.22 \\
    \hdashline[1pt/2pt]
    Grok 3 mini beta & 70.05 & 70.05 & 69.61 & 69.12 & 68.99 & 78.67 & 75.45 & 73.46 & 72.23 & 71.09 \\
    \hdashline[1pt/2pt]
    Qwen 3 30B A3B & 67.77 & 62.89 & 63.76 & 64.10 & 63.00 & 73.27 & 71.04 & 70.74 & 68.98 & 68.91 \\
    Qwen 2.5 72B Instruct & 64.83 & 63.65 & 65.97 & 64.88 & 66.14 & 74.50 & 69.43 & 69.38 & 67.70 & 67.51 \\
    \hdashline[1pt/2pt]
    Gemma 3 27B & 61.99 & 62.09 & 62.53 & 63.51 & 63.56 & 66.92 & 64.83 & 64.01 & 63.44 & 63.41 \\
    Gemma 3 12B & 61.33 & 62.09 & 62.46 & 63.25 & 62.29 & 66.92 & 63.65 & 63.44 & 61.73 & 60.76 \\
    \hdashline[1pt/2pt]
    Mistral Medium 3 & 62.37 & 61.28 & 61.90 & 62.44 & 62.45 & 69.67 & 64.31 & 63.95 & 63.67 & 63.37 \\
    \hdashline[1pt/2pt]
    Kimi K2 & 62.75 & 63.65 & 64.80 & 64.38 & 64.76 & 76.97 & 74.64 & 74.31 & 72.94 & 73.65 \\
    \bottomrule
    \end{NiceTabular}
    \vspace{3mm}
    \caption{\textbf{Instruction-level IF scores on Live-SWE-IF}. Higher is better.}
    \label{tab:live_swe_if_instr_if}
\end{table*}

\renewcommand\arraystretch{1.7}
\begin{table*}[h]
    \centering \scriptsize
    \tabcolsep0.1 in
    \begin{NiceTabular}{lcccccccccc}
    \CodeBefore
    \rectanglecolor{bgyellow}{1-2}{26-6}
    \rectanglecolor{bgred}{1-7}{26-11}
    \Body
    \toprule
    \multirow[c]{2}{*}[-1.0ex]{\textbf{Models}} & \multicolumn{5}{c}{\textbf{Single-Turn Generation $\uparrow$}} & \multicolumn{5}{c}{\textbf{Multi-Turn Editing $\uparrow$}} \\
    \cmidrule(lr){2-6}\cmidrule(lr){7-11}
     & \multicolumn{1}{c}{1 Inst} & \multicolumn{1}{c}{2 Inst} & \multicolumn{1}{c}{3 Inst} & \multicolumn{1}{c}{4 Inst} & \multicolumn{1}{c}{5 Inst} & \multicolumn{1}{c}{1 Inst} & \multicolumn{1}{c}{2 Inst} & \multicolumn{1}{c}{3 Inst} & \multicolumn{1}{c}{4 Inst} & \multicolumn{1}{c}{5 Inst} \\
    \cmidrule(lr){1-1}\cmidrule(lr){2-6}\cmidrule(lr){7-11}
    Gemini 2.5 Pro & 75.83 & 56.78 & 45.50 & 37.63 & 29.57 & 78.96 & 61.61 & 51.18 & 41.04 & 32.80 \\
    Gemini 2.5 Flash & 66.54 & 45.97 & 32.89 & 23.03 & 17.06 & 72.80 & 51.09 & 34.98 & 25.31 & 17.82 \\
    Gemini 2.0 Flash & 61.71 & 37.25 & 22.18 & 13.46 & 8.44 & 74.41 & 51.28 & 36.02 & 24.64 & 17.73 \\
    Gemini 2.0 Flash Lite & 62.94 & 41.33 & 27.68 & 18.39 & 12.89 & 67.30 & 42.75 & 27.11 & 17.35 & 10.62 \\
    \hdashline[1pt/2pt]
    Claude 4 Opus & 78.86 & 57.91 & 47.96 & 38.96 & 35.17 & \textbf{85.59} & 72.89 & 61.71 & 52.04 & 43.70 \\
    Claude 4 Sonnet & 75.73 & 56.40 & 44.17 & 35.36 & 28.53 & 84.45 & 73.46 & 62.37 & 52.70 & 44.64 \\
    Claude 3.7 Sonnet & 72.42 & 47.01 & 31.85 & 23.51 & 18.96 & 79.53 & 62.46 & 48.53 & 38.58 & 30.33 \\
    Claude 3.5 Sonnet & 70.52 & 47.01 & 31.94 & 22.37 & 14.88 & 80.57 & 60.28 & 44.45 & 35.73 & 27.20 \\
    Claude 3.5 Haiku & 63.22 & 35.92 & 22.84 & 16.40 & 11.66 & 78.67 & 58.58 & 41.23 & 30.52 & 22.46 \\
    Claude 3 Haiku & 61.61 & 36.68 & 23.13 & 15.83 & 9.95 & 72.80 & 52.32 & 36.59 & 26.82 & 18.58 \\
    \hdashline[1pt/2pt]
    DeepSeek V3 0324 & 52.80 & 29.76 & 19.24 & 11.28 & 7.77 & 70.05 & 45.31 & 31.56 & 20.38 & 11.37 \\
    \hdashline[1pt/2pt]
    GPT 5 & \textbf{82.18} & \textbf{68.53} & \textbf{55.17} & \textbf{47.01} & \textbf{40.95} & \textbf{85.59} & \textbf{74.50} & \textbf{66.64} & \textbf{57.35} & \textbf{50.14} \\
    o4 mini & 73.18 & 53.93 & 43.22 & 33.36 & 27.20 & 81.52 & 66.64 & 54.60 & 42.84 & 32.61 \\
    GPT 4.1 & 68.63 & 42.27 & 29.48 & 20.57 & 13.65 & 74.12 & 54.31 & 41.04 & 32.70 & 24.08 \\
    GPT 4.1 mini & 67.20 & 43.60 & 30.71 & 21.99 & 14.88 & 71.75 & 49.67 & 34.60 & 26.26 & 18.20 \\
    GPT 4o & 60.85 & 36.40 & 22.75 & 14.98 & 9.95 & 76.40 & 53.93 & 37.91 & 28.44 & 20.38 \\
    GPT 4o mini & 65.88 & 42.65 & 28.25 & 20.19 & 14.41 & 73.93 & 52.32 & 36.21 & 24.93 & 17.06 \\
    \hdashline[1pt/2pt]
    Grok 3 mini beta & 70.05 & 51.09 & 38.10 & 28.15 & 20.66 & 78.67 & 58.39 & 43.89 & 32.99 & 24.64 \\
    \hdashline[1pt/2pt]
    Qwen 3 30B A3B & 67.77 & 42.09 & 29.67 & 22.65 & 14.50 & 73.27 & 52.89 & 39.72 & 29.10 & 21.90 \\
    Qwen 2.5 72B Instruct & 64.83 & 40.76 & 28.63 & 18.86 & 15.92 & 74.50 & 50.24 & 37.06 & 26.73 & 19.24 \\
    \hdashline[1pt/2pt]
    Gemma 3 27B & 61.99 & 37.73 & 24.17 & 16.30 & 11.00 & 66.92 & 41.71 & 27.30 & 17.91 & 12.32 \\
    Gemma 3 12B & 61.33 & 38.20 & 24.17 & 16.30 & 9.19 & 66.92 & 41.71 & 26.54 & 16.40 & 10.05 \\
    \hdashline[1pt/2pt]
    Mistral Medium 3 & 62.37 & 37.25 & 23.13 & 15.83 & 9.86 & 69.67 & 42.94 & 27.96 & 18.48 & 11.66 \\
    \hdashline[1pt/2pt]
    Kimi K2 & 62.75 & 41.61 & 27.77 & 19.05 & 11.94 & 76.97 & 57.35 & 44.17 & 35.73 & 27.87 \\
    \bottomrule
    \end{NiceTabular}
    \vspace{3mm}
    \caption{\textbf{Task-level IF scores on Live-SWE-IF}. Higher is better.}
    \label{tab:live_swe_if_task_if}
\end{table*}

\clearpage
\newpage
\section{Analysis}
\subsection{Instruction Position Analysis}

As listed in Tables~\ref{tab:big_swe_if_pos} and~\ref{tab:live_swe_if_pos}, we also provide detailed results for per-position instruction-level IF scores on both benchmarks.

\renewcommand\arraystretch{1.5}
\begin{table*}[h]
    \centering \scriptsize
    \tabcolsep0.1 in
    \begin{NiceTabular}{lcccccccccc}
    \CodeBefore
    \rectanglecolor{bgyellow}{1-2}{33-6}
    \rectanglecolor{bgred}{1-7}{33-11}
    \Body
    \toprule
    \multirow[c]{2}{*}[-1.0ex]{\textbf{Models}} & \multicolumn{5}{c}{\textbf{Single-Turn Generation $\uparrow$}} & \multicolumn{5}{c}{\textbf{Multi-Turn Editing $\uparrow$}} \\
    \cmidrule(lr){2-6}\cmidrule(lr){7-11}
     & \multicolumn{1}{c}{Pos 1} & \multicolumn{1}{c}{Pos 2} & \multicolumn{1}{c}{Pos 3} & \multicolumn{1}{c}{Pos 4} & \multicolumn{1}{c}{Pos 5} & \multicolumn{1}{c}{Pos 1} & \multicolumn{1}{c}{Pos 2} & \multicolumn{1}{c}{Pos 3} & \multicolumn{1}{c}{Pos 4} & \multicolumn{1}{c}{Pos 5} \\
    \cmidrule(lr){1-1}\cmidrule(lr){2-6}\cmidrule(lr){7-11}
    Gemini 2.5 Pro & 81.40 & 79.91 & 78.86 & 78.60 & 78.60 & 79.91 & 78.16 & 79.56 & 81.58 & 83.16 \\
    Gemini 2.5 Flash & 79.82 & 74.04 & 75.44 & 75.00 & 75.26 & 72.19 & 69.74 & 72.81 & 73.25 & 78.25 \\
    Gemini 2.0 Flash & 74.56 & 71.23 & 72.02 & 70.96 & 71.40 & 72.81 & 71.75 & 74.47 & 73.77 & 73.68 \\
    Gemini 2.0 Flash Lite & 72.11 & 68.77 & 67.89 & 67.89 & 67.81 & 68.16 & 67.81 & 69.12 & 68.25 & 70.79 \\
    \hdashline[1pt/2pt]
    Claude 4 Opus & \textbf{87.46} & \textbf{85.18} & \textbf{85.18} & \textbf{83.95} & \textbf{86.23} & 84.56 & 82.81 & 84.82 & 83.60 & 85.70 \\
    Claude 4 Sonnet & 83.77 & 79.65 & 80.26 & 81.67 & 81.49 & 84.21 & 80.18 & 83.95 & 84.39 & 87.19 \\
    Claude 3.7 Sonnet & 77.81 & 73.16 & 72.19 & 73.07 & 75.09 & 77.63 & 76.75 & 76.93 & 78.60 & 80.88 \\
    Claude 3.5 Sonnet & 77.46 & 72.81 & 72.72 & 75.18 & 75.35 & 76.40 & 75.53 & 75.09 & 78.33 & 82.11 \\
    Claude 3.5 Haiku & 65.00 & 62.46 & 61.49 & 62.89 & 67.28 & 60.70 & 63.68 & 64.65 & 65.79 & 73.16 \\
    Claude 3 Haiku & 67.89 & 62.72 & 62.28 & 63.86 & 65.88 & 72.11 & 66.40 & 69.47 & 68.95 & 75.88 \\
    \hdashline[1pt/2pt]
    DeepSeek R1 0528 & 69.30 & 66.58 & 66.58 & 67.63 & 67.98 & 68.86 & 68.33 & 70.53 & 71.32 & 76.23 \\
    DeepSeek V3 0324 & 67.63 & 62.81 & 64.47 & 64.74 & 65.79 & 61.67 & 63.95 & 68.25 & 69.12 & 75.61 \\
    \hdashline[1pt/2pt]
    GPT 5 & 82.72 & 81.40 & 81.05 & 81.58 & 82.11 & \textbf{86.05} & \textbf{86.05} & \textbf{87.02} & \textbf{85.96} & 86.84 \\
    o4 mini & 85.53 & 82.81 & 83.07 & 83.77 & 86.05 & 81.40 & 80.44 & 82.11 & 84.04 & \textbf{88.42} \\
    o3 mini high & 73.25 & 72.19 & 69.04 & 71.84 & 72.11 & 77.46 & 74.65 & 77.19 & 78.95 & 82.54 \\
    GPT 4.1 & 78.95 & 76.75 & 76.40 & 78.33 & 78.33 & 78.68 & 75.88 & 77.63 & 77.19 & 79.56 \\
    GPT 4.1 mini & 75.00 & 72.63 & 72.63 & 71.49 & 75.35 & 75.26 & 70.26 & 71.84 & 72.02 & 76.05 \\
    GPT 4o & 78.77 & 72.72 & 71.40 & 71.75 & 72.54 & 78.42 & 75.79 & 77.19 & 78.42 & 81.93 \\
    GPT 4o mini & 76.14 & 71.32 & 71.84 & 72.81 & 74.47 & 71.58 & 71.84 & 72.89 & 73.51 & 78.16 \\
    \hdashline[1pt/2pt]
    Grok 4 & 86.32 & 83.77 & 84.39 & 83.86 & 85.70 & 84.91 & 84.56 & 85.70 & 84.74 & 86.93 \\
    Grok 3 mini beta & 79.56 & 76.40 & 76.40 & 75.44 & 79.47 & 73.07 & 72.89 & 72.89 & 75.26 & 81.14 \\
    \hdashline[1pt/2pt]
    Qwen 3 235B A22B & 82.54 & 78.16 & 77.28 & 77.28 & 77.89 & 77.98 & 76.84 & 77.54 & 80.61 & 81.49 \\
    Qwen 3 32B & 73.68 & 70.61 & 71.75 & 69.82 & 70.88 & 76.49 & 73.42 & 76.23 & 77.02 & 78.51 \\
    Qwen 3 30B A3B & 71.67 & 67.81 & 68.60 & 68.95 & 72.02 & 77.19 & 74.39 & 77.02 & 75.53 & 78.60 \\
    Qwen 2.5 72B Instruct & 73.07 & 71.05 & 69.47 & 68.33 & 69.74 & 73.07 & 70.88 & 72.89 & 72.72 & 74.21 \\
    Qwen 2.5 Coder  & 68.07 & 64.74 & 64.74 & 64.39 & 66.93 & 67.98 & 65.96 & 69.12 & 69.56 & 71.67 \\
    \hdashline[1pt/2pt]
    Gemma 3 27B & 68.07 & 64.30 & 63.86 & 64.74 & 64.12 & 65.96 & 65.35 & 67.63 & 66.75 & 67.89 \\
    Gemma 3 12B & 67.98 & 66.14 & 64.30 & 63.60 & 62.98 & 65.00 & 62.98 & 65.35 & 63.77 & 67.63 \\
    \hdashline[1pt/2pt]
    Mistral Medium 3 & 74.65 & 69.82 & 68.42 & 70.79 & 69.04 & 70.88 & 68.86 & 69.74 & 71.75 & 77.02 \\
    \hdashline[1pt/2pt]
    MiniMax M1 & 72.37 & 70.70 & 70.09 & 70.00 & 71.32 & 70.61 & 70.61 & 72.89 & 73.86 & 76.75 \\
    \hdashline[1pt/2pt]
    Kimi K2 & 83.42 & 79.12 & 77.46 & 78.25 & 77.46 & 84.47 & 83.86 & 82.19 & 83.51 & 86.93 \\
    \bottomrule
    \end{NiceTabular}
    \vspace{3mm}
    \caption{\textbf{Instruction-position analysis on Big-SWE-IF}. We report instruction-level IF scores under the setting of five instructions, comparing positions 1--5 in each setting. In Single-Turn Generation, position $i$ denotes the $i$-th item in the numbered instruction list given to the model. In Multi-Turn Editing, position $i$ indicates the $i$-th instruction introduced as a separate turn.}
    \label{tab:big_swe_if_pos}
\end{table*}

\renewcommand\arraystretch{1.7}
\begin{table*}[h]
    \centering \scriptsize
    \tabcolsep0.1 in
    \begin{NiceTabular}{lcccccccccc}
    \CodeBefore
    \rectanglecolor{bgyellow}{1-2}{26-6}
    \rectanglecolor{bgred}{1-7}{26-11}
    \Body
    \toprule
    \multirow[c]{2}{*}[-1.0ex]{\textbf{Models}} & \multicolumn{5}{c}{\textbf{Single-Turn Generation $\uparrow$}} & \multicolumn{5}{c}{\textbf{Multi-Turn Editing $\uparrow$}} \\
    \cmidrule(lr){2-6}\cmidrule(lr){7-11}
     & \multicolumn{1}{c}{Pos 1} & \multicolumn{1}{c}{Pos 2} & \multicolumn{1}{c}{Pos 3} & \multicolumn{1}{c}{Pos 4} & \multicolumn{1}{c}{Pos 5} & \multicolumn{1}{c}{Pos 1} & \multicolumn{1}{c}{Pos 2} & \multicolumn{1}{c}{Pos 3} & \multicolumn{1}{c}{Pos 4} & \multicolumn{1}{c}{Pos 5} \\
    \cmidrule(lr){1-1}\cmidrule(lr){2-6}\cmidrule(lr){7-11}
    Gemini 2.5 Pro & 76.78 & 75.83 & 77.54 & 76.97 & 77.25 & 76.49 & 76.59 & 78.77 & 80.57 & 82.46 \\
    Gemini 2.5 Flash & 70.52 & 67.39 & 69.57 & 67.58 & 65.59 & 66.16 & 64.93 & 67.39 & 67.87 & 73.08 \\
    Gemini 2.0 Flash & 63.41 & 60.95 & 63.79 & 61.33 & 62.94 & 69.10 & 67.87 & 68.53 & 69.76 & 72.32 \\
    Gemini 2.0 Flash Lite & 67.39 & 65.21 & 65.97 & 63.60 & 66.26 & 60.38 & 60.09 & 61.42 & 60.19 & 63.89 \\
    \hdashline[1pt/2pt]
    Claude 4 Opus & 79.91 & 77.91 & 78.29 & 77.06 & 80.57 & 84.83 & 84.64 & 83.41 & 82.75 & 84.93 \\
    Claude 4 Sonnet & 75.64 & 75.45 & 75.26 & 74.03 & 75.64 & 81.90 & 85.31 & 84.17 & 82.75 & \textbf{85.21} \\
    Claude 3.7 Sonnet & 70.81 & 65.97 & 69.38 & 67.01 & 68.72 & 76.30 & 76.40 & 76.02 & 75.73 & 79.34 \\
    Claude 3.5 Sonnet & 67.87 & 67.01 & 67.77 & 65.78 & 67.96 & 72.32 & 71.94 & 71.94 & 72.51 & 76.87 \\
    Claude 3.5 Haiku & 63.03 & 62.46 & 61.52 & 62.65 & 64.17 & 67.58 & 66.26 & 68.06 & 68.44 & 74.60 \\
    Claude 3 Haiku & 62.65 & 57.82 & 60.66 & 60.38 & 60.76 & 68.63 & 66.35 & 68.25 & 64.74 & 69.29 \\
    \hdashline[1pt/2pt]
    DeepSeek V3 0324 & 55.64 & 56.11 & 56.11 & 55.55 & 55.55 & 58.29 & 59.34 & 62.75 & 64.74 & 68.44 \\
    \hdashline[1pt/2pt]
    GPT 5 & \textbf{83.51} & \textbf{82.37} & \textbf{83.70} & \textbf{81.33} & \textbf{83.22} & \textbf{86.26} & \textbf{86.92} & \textbf{86.07} & \textbf{84.55} & 85.02 \\
    o4 mini & 75.17 & 74.31 & 74.41 & 71.56 & 73.65 & 73.46 & 74.31 & 76.11 & 75.55 & 79.62 \\
    GPT 4.1 & 67.20 & 65.97 & 68.44 & 64.55 & 67.20 & 71.28 & 69.76 & 72.42 & 69.29 & 71.28 \\
    GPT 4.1 mini & 68.82 & 67.49 & 69.95 & 63.79 & 67.01 & 67.30 & 66.54 & 69.38 & 65.97 & 69.38 \\
    GPT 4o & 62.84 & 61.61 & 63.32 & 60.28 & 61.61 & 68.44 & 68.15 & 71.09 & 68.06 & 73.93 \\
    GPT 4o mini & 67.01 & 64.83 & 66.26 & 64.45 & 65.59 & 67.30 & 65.97 & 67.49 & 64.83 & 70.52 \\
    \hdashline[1pt/2pt]
    Grok 3 mini beta & 69.10 & 67.39 & 69.48 & 68.44 & 70.52 & 67.11 & 69.86 & 69.76 & 72.23 & 76.49 \\
    \hdashline[1pt/2pt]
    Qwen 3 30B A3B & 62.94 & 63.03 & 64.36 & 62.65 & 61.99 & 66.73 & 68.06 & 69.86 & 69.19 & 70.71 \\
    Qwen 2.5 72B Instruct & 66.54 & 64.08 & 68.25 & 65.88 & 65.97 & 66.35 & 65.88 & 68.82 & 67.20 & 69.29 \\
    \hdashline[1pt/2pt]
    Gemma 3 27B & 65.97 & 62.56 & 64.17 & 63.41 & 61.71 & 62.27 & 61.80 & 62.65 & 63.13 & 67.20 \\
    Gemma 3 12B & 64.36 & 62.37 & 63.13 & 60.76 & 60.85 & 60.76 & 58.67 & 60.57 & 58.96 & 64.83 \\
    \hdashline[1pt/2pt]
    Mistral Medium 3 & 63.32 & 61.23 & 63.41 & 61.71 & 62.56 & 60.38 & 58.67 & 63.79 & 64.83 & 69.19 \\
    \hdashline[1pt/2pt]
    Kimi K2 & 66.64 & 64.17 & 65.97 & 62.94 & 64.08 & 72.61 & 72.51 & 74.69 & 73.27 & 75.17 \\
    \bottomrule
    \end{NiceTabular}
    \vspace{3mm}
    \caption{\textbf{Instruction-position analysis on Live-SWE-IF}. We report instruction-level IF scores under the setting of five instructions, comparing positions 1--5 in each setting. In Single-Turn Generation, position $i$ denotes the $i$-th item in the numbered instruction list given to the model. In Multi-Turn Editing, position $i$ indicates the $i$-th instruction introduced as a separate turn.}
    \label{tab:live_swe_if_pos}
\end{table*}

\clearpage
\newpage
\subsection{Correlation Analysis}
\label{sec:full_correlation}

\begin{figure}[H]
    \centering
    \begin{subfigure}[b]{0.9\textwidth}
        \centering
        \includegraphics[width=\textwidth]{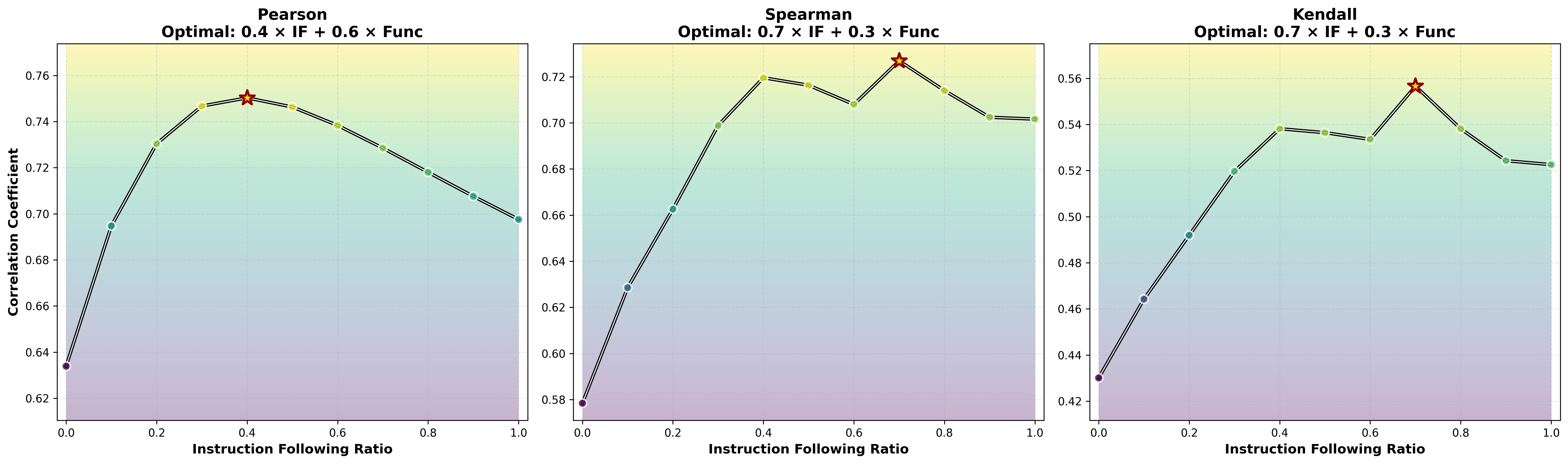}
        \caption{Big-SWE-IF w/ Style Control}
        \label{fig:bigvibebench_w_sc}
    \end{subfigure}
    \hfill
    \begin{subfigure}[b]{0.9\textwidth}
        \centering
        \includegraphics[width=\textwidth]{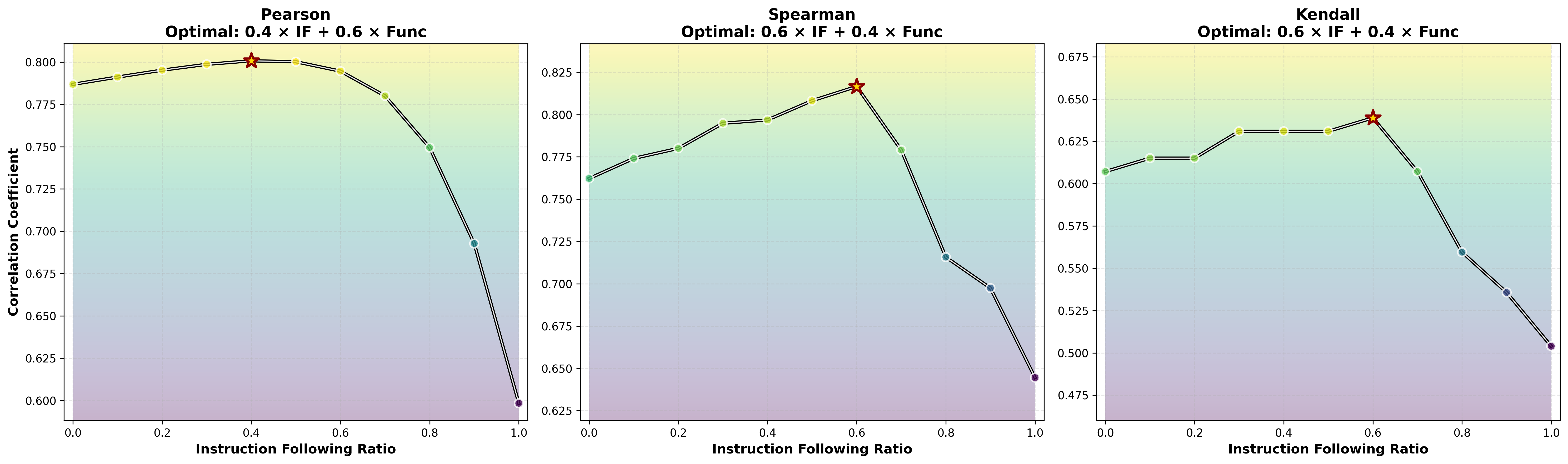}
        \caption{Live-SWE-IF w/ Style Control}
        \label{fig:livevibebench_w_sc}
    \end{subfigure}
    \hfill
    \begin{subfigure}[b]{0.9\textwidth}
        \centering
        \includegraphics[width=\textwidth]{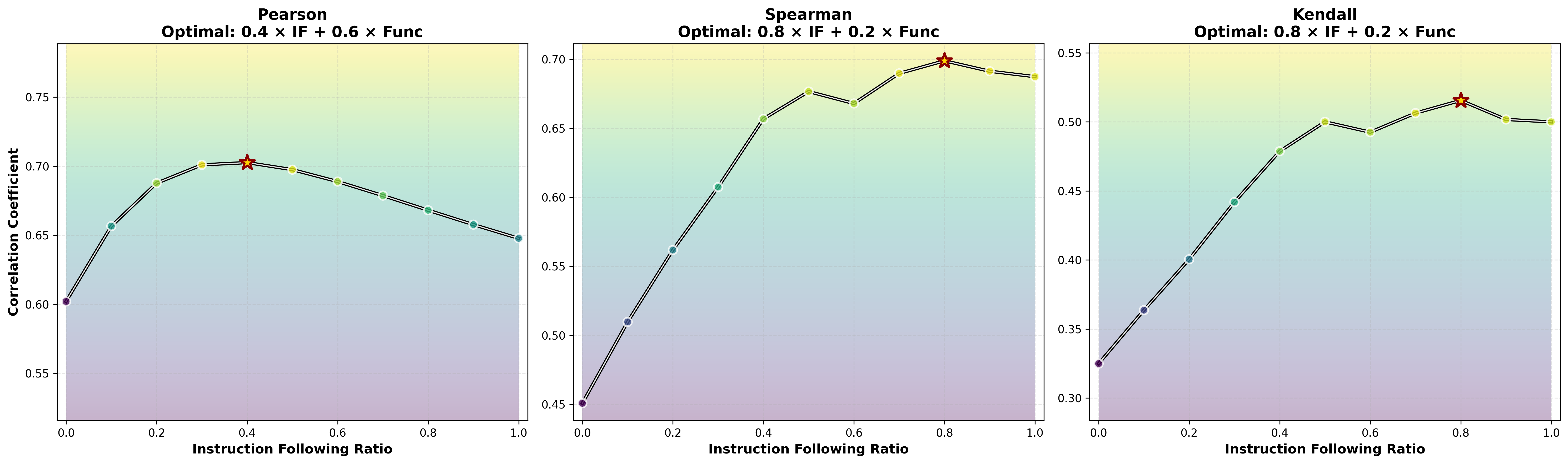}
        \caption{Big-SWE-IF w/o Style Control}
        \label{fig:bigvibebench_wo_sc}
    \end{subfigure}
    \hfill
    \begin{subfigure}[b]{0.9\textwidth}
        \centering
        \includegraphics[width=\textwidth]{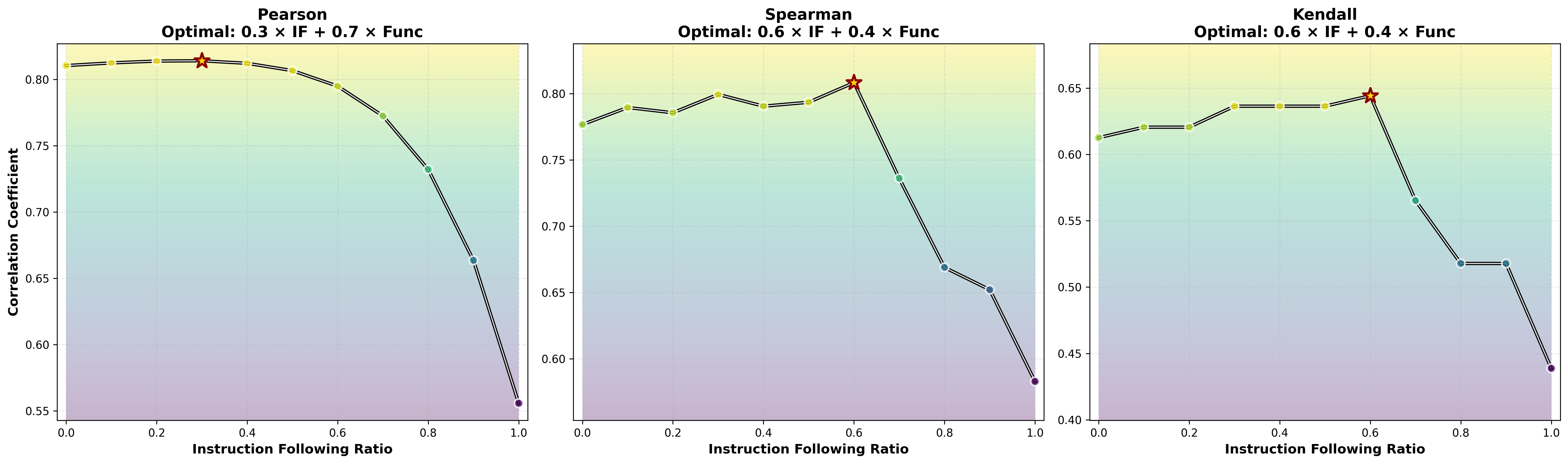}
        \caption{Live-SWE-IF w/o Style Control}
        \label{fig:livevibebench_wo_sc}
    \end{subfigure}
    \caption{\textbf{Human preference aligns best with a mixture of IF and functionality.} We correlate LMArena coding Elo with a composite score \(\alpha\,\text{IF} + (1-\alpha)\,\text{Func}\), where \(\alpha\in[0,1]\) is the weight on IF (x-axis). We also vary the correlation type and toggle the style-control function on/off. Nevertheless, the peak correlation (starred) consistently occurs at a mixture of the two metrics across all settings.}
    \label{fig:full_correlation_analysis}
\end{figure}

\clearpage
\newpage
\subsection{Error Analysis via Category-wise IF Profiles}
\label{app:category_profiles}

To better understand why some models perform better than others and what their errors look like, we analyze instruction-level IF by instruction category for top-performing models in the single-turn setting with five instructions, summarized in Tables~\ref{tab:big_swe_if_category_if} and~\ref{tab:live_swe_if_category_if} (the \textit{Library} category is omitted for Live-SWE-IF).

\renewcommand\arraystretch{1.3}
\begin{table}[H]
    \centering \footnotesize
    \tabcolsep0.13 in
    \begin{NiceTabular}{lcccccc}
    \toprule
    \textbf{Models} & \textbf{Doc.} & \textbf{Error} & \textbf{Library} & \textbf{Logic} & \textbf{Style} & \textbf{Overall} \\
    \midrule
    Gemini 2.5 Pro    & 52.32 & 99.44 & \textbf{100.00} & 84.38 & \textbf{90.31} & 79.47 \\
    Gemini 2.5 Flash  & 60.25 & 96.38 & 97.58 & 81.06 & 76.89 & 75.91 \\
    Claude 4 Opus     & 79.35 & 98.89 & \textbf{100.00} & 88.73 & 83.26 & \textbf{85.60} \\
    Claude 4 Sonnet   & 74.16 & 98.61 & 99.19 & 87.07 & 75.46 & 81.28 \\
    GPT 5             & 68.61 & \textbf{99.72} & \textbf{100.00} & \textbf{93.61} & 73.62 & 81.77 \\
    o4 mini           & \textbf{80.55} & 99.44 & 98.39 & 90.29 & 75.40 & 84.05 \\
    Kimi K2           & 78.44 & 98.33 & \textbf{100.00} & 70.91 & 83.94 & 79.14 \\
    \bottomrule
    \end{NiceTabular}
    \vspace{2mm}
    \caption{\textbf{Category-wise instruction-level IF rates on Big-SWE-IF} in the single-turn setting with five instructions.}
    \label{tab:big_swe_if_category_if}
\end{table}

\renewcommand\arraystretch{1.4}
\begin{table}[H]
    \centering \footnotesize
    \tabcolsep0.135 in
    \begin{NiceTabular}{lccccc}
    \toprule
    \textbf{Models} & \textbf{Doc.} & \textbf{Error} & \textbf{Logic} & \textbf{Style} & \textbf{Overall} \\
    \midrule
    Gemini 2.5 Pro    & 53.90 & \textbf{100.00} & 76.49 & \textbf{96.52} & 76.87 \\
    Gemini 2.5 Flash  & 54.18 & \textbf{100.00} & 68.06 & 79.24 & 68.13 \\
    Claude 4 Opus     & 67.46 & \textbf{100.00} & 81.25 & 84.34 & 78.75 \\
    Claude 4 Sonnet   & 69.34 & 98.00 & 74.79 & 80.11 & 75.20 \\
    GPT 5             & 76.41 & \textbf{100.00} & \textbf{84.69} & 85.21 & \textbf{82.82} \\
    o4 mini           & \textbf{78.28} & 92.00 & 68.28 & 77.07 & 73.82 \\
    Kimi K2           & 73.23 & 96.00 & 49.62 & 77.44 & 64.76 \\
    \bottomrule
    \end{NiceTabular}
    \vspace{2mm}
    \caption{\textbf{Category-wise instruction-level IF rates on Live-SWE-IF} for top-performing models in the single-turn setting with five instructions. The \textit{Library} category is omitted because it is negligible in Live-SWE-IF.}
    \label{tab:live_swe_if_category_if}
\end{table}

\textbf{Near-universal compliance for error handling and library usage.}
Across models, the \textit{Error Handling} category achieves near-perfect IF, and the \textit{Library} category on Big-SWE-IF is also close to 100\%. This suggests that modern frontier models reliably follow constraints that have explicit, localized code signatures, such as adding guard clauses, handling exceptions, or using a mandated library.

\textbf{Documentation remains the main bottleneck.}
In contrast, \textit{Documentation} is consistently the lowest-performing category across both benchmarks. Even among strong models, documentation IF is far below other categories, indicating that following the requested convention is still error-prone.

\textbf{Model specialization by category.}
The category breakdown also reveals distinct performance profiles. On Big-SWE-IF, Gemini 2.5 Pro performs best on \textit{Coding Style}, while GPT 5 is strongest on \textit{Logic \& Code Patterns}. o4 mini achieves the highest documentation IF on both benchmarks. Claude 4 Opus exhibits the most balanced profile, without an obvious weak category, which aligns with its strong overall IF.

\textbf{Consistency across benchmarks.}
These patterns are largely consistent between the real-world-task setting (Big-SWE-IF) and the algorithmic setting (Live-SWE-IF), suggesting that the observed strengths and weaknesses reflect model tendencies rather than benchmark-specific artifacts.

These insights provide actionable guidance for model selection based on specific user needs (e.g., choosing Gemini for style-strict projects or GPT-5 for logic-heavy tasks) and highlight targeted areas for future model development.

\clearpage
\newpage
\subsection{Trending Analysis across Model Families}
\label{app:trending_analysis}

\begin{figure}[H]
    \centering
    \makebox[\textwidth][c]{%
        \begin{subfigure}[b]{0.36\textwidth}
            \centering
            \includegraphics[width=\textwidth]{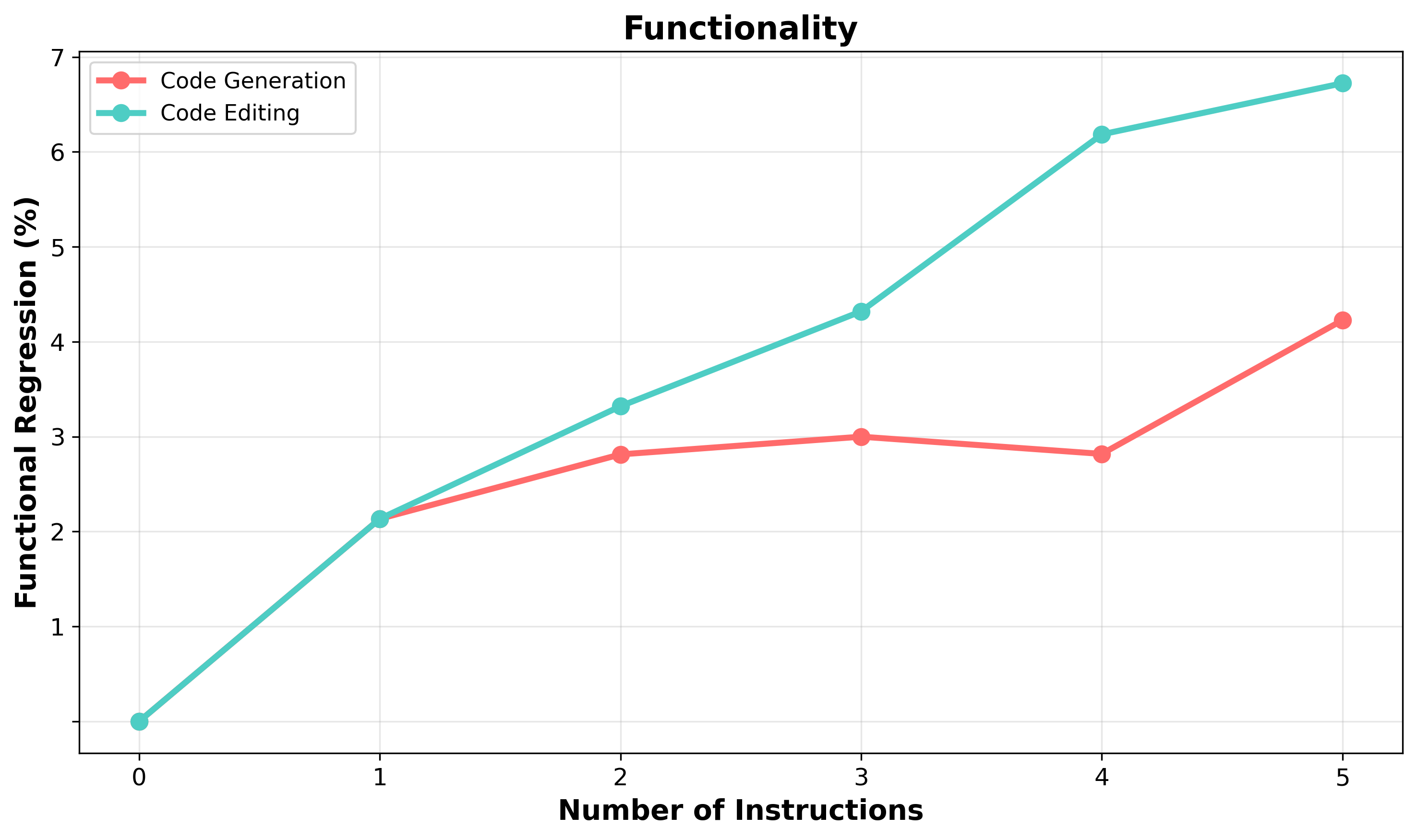}
            \caption{Gemini}
            \label{fig:trend_func_gemini}
        \end{subfigure}
        \hspace{0.04\textwidth}%
        \begin{subfigure}[b]{0.36\textwidth}
            \centering
            \includegraphics[width=\textwidth]{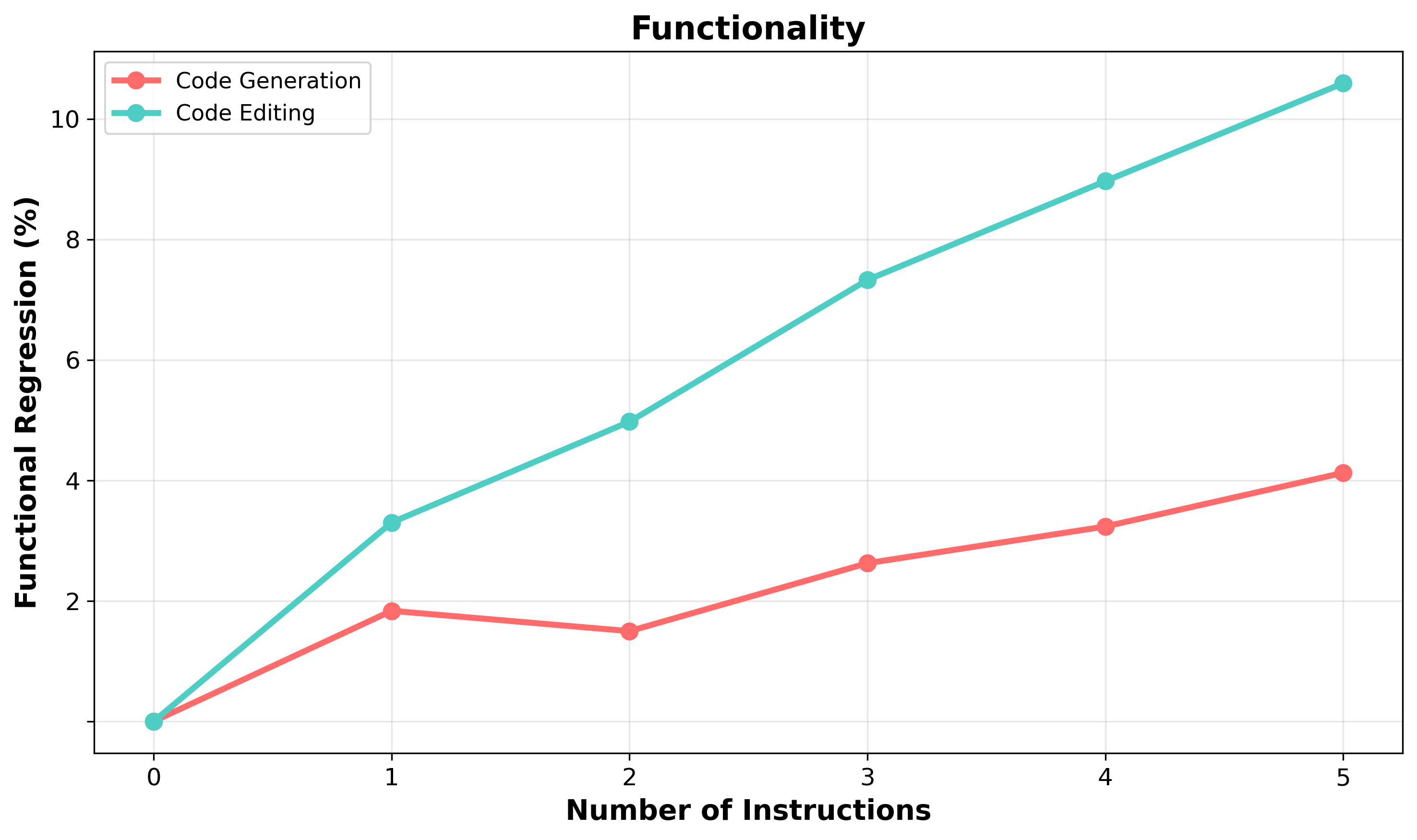}
            \caption{Claude}
            \label{fig:trend_func_claude}
        \end{subfigure}
    }

    \vspace{-1mm}

    \makebox[\textwidth][c]{%
        \begin{subfigure}[b]{0.36\textwidth}
            \centering
            \includegraphics[width=\textwidth]{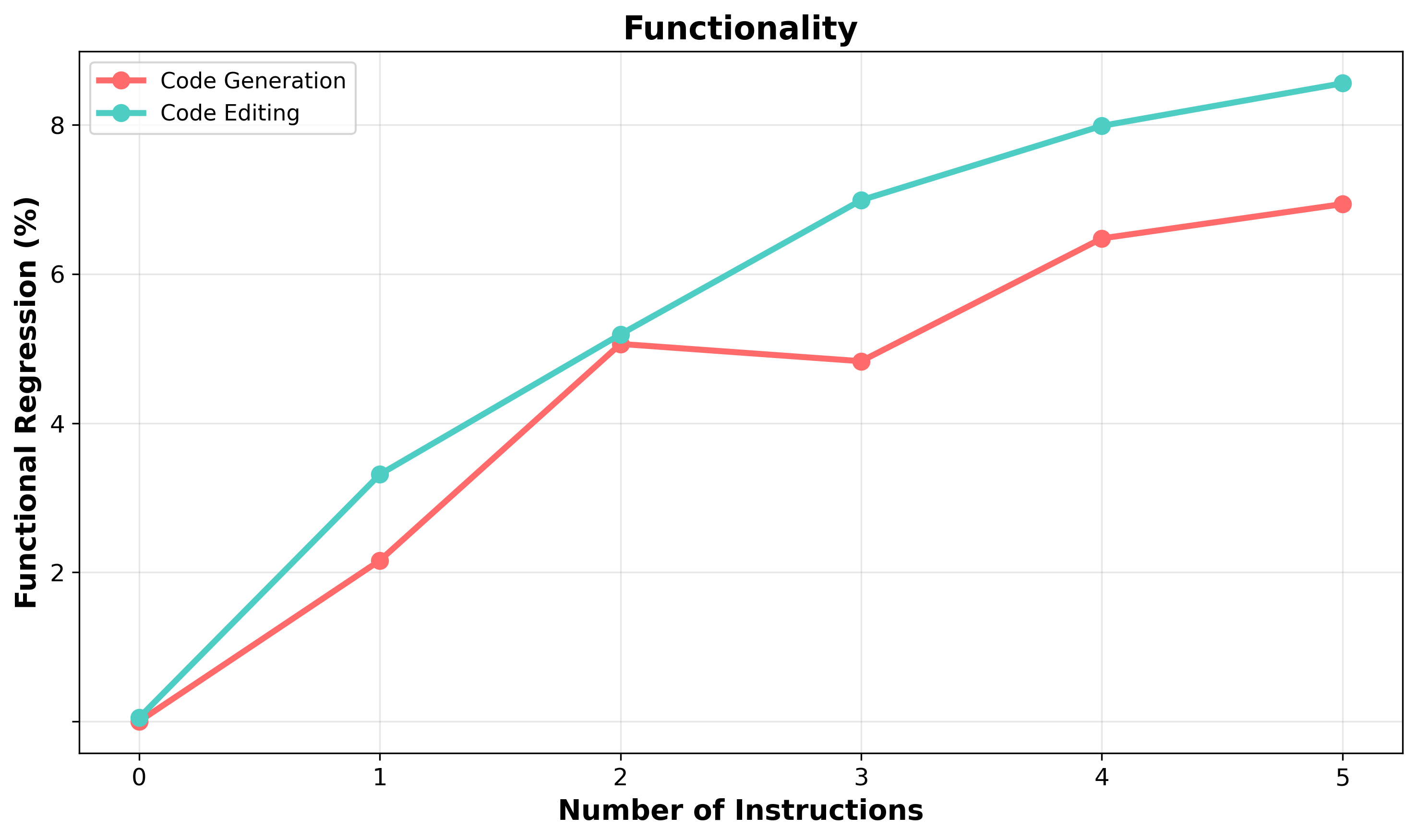}
            \caption{OpenAI}
            \label{fig:trend_func_openai}
        \end{subfigure}
        \hspace{0.04\textwidth}%
        \begin{subfigure}[b]{0.36\textwidth}
            \centering
            \includegraphics[width=\textwidth]{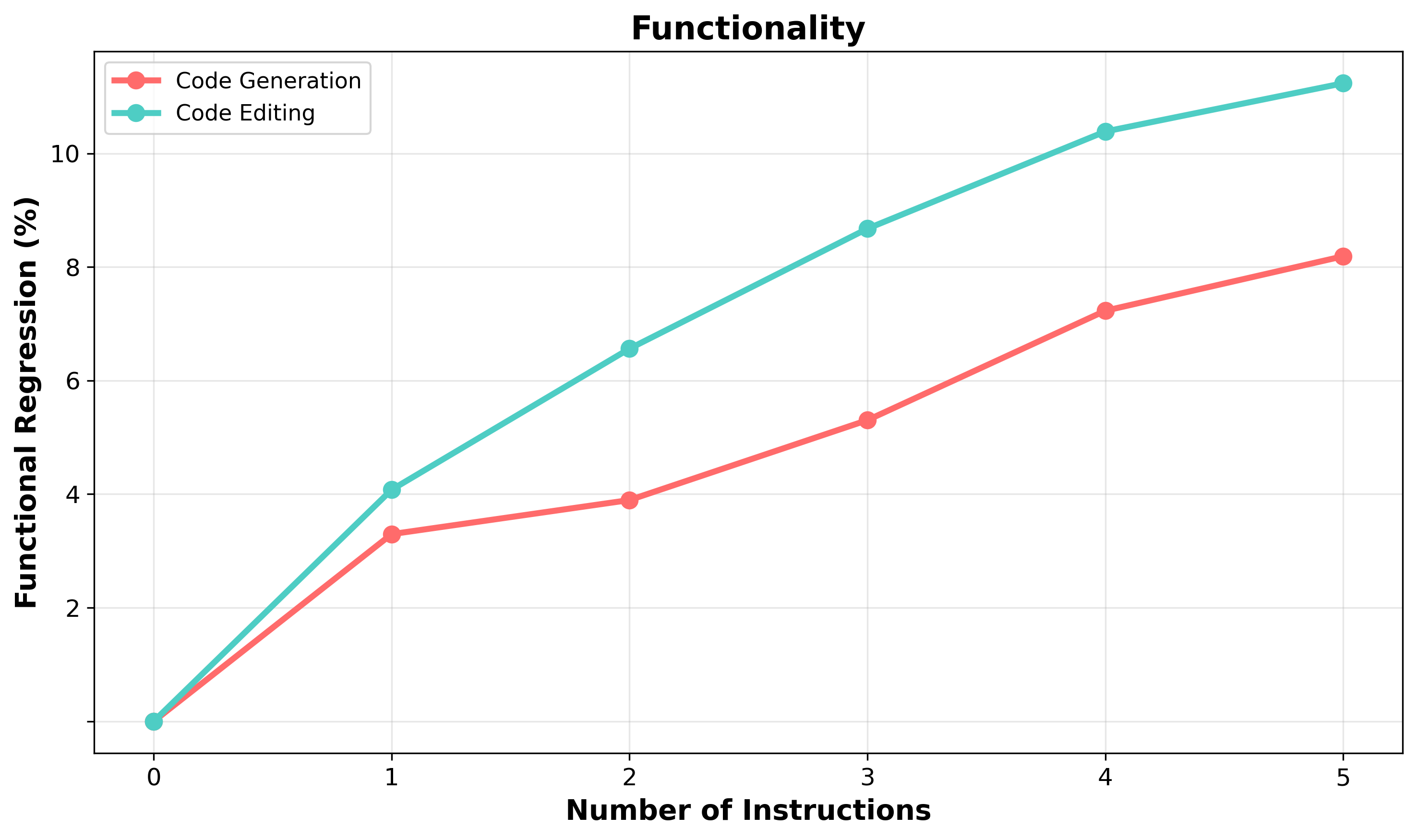}
            \caption{Qwen}
            \label{fig:trend_func_qwen}
        \end{subfigure}
    }

    \vspace{-2mm}
    \caption{\textbf{Results for functionality across model families.} Each subplot shows functional regression versus the number of injected instructions, for single-turn generation and multi-turn editing.}
    \label{fig:trending_func_families}
\end{figure}

\begin{figure}[H]
    \centering
    \begin{subfigure}[b]{0.48\textwidth}
        \centering
        \includegraphics[width=\textwidth]{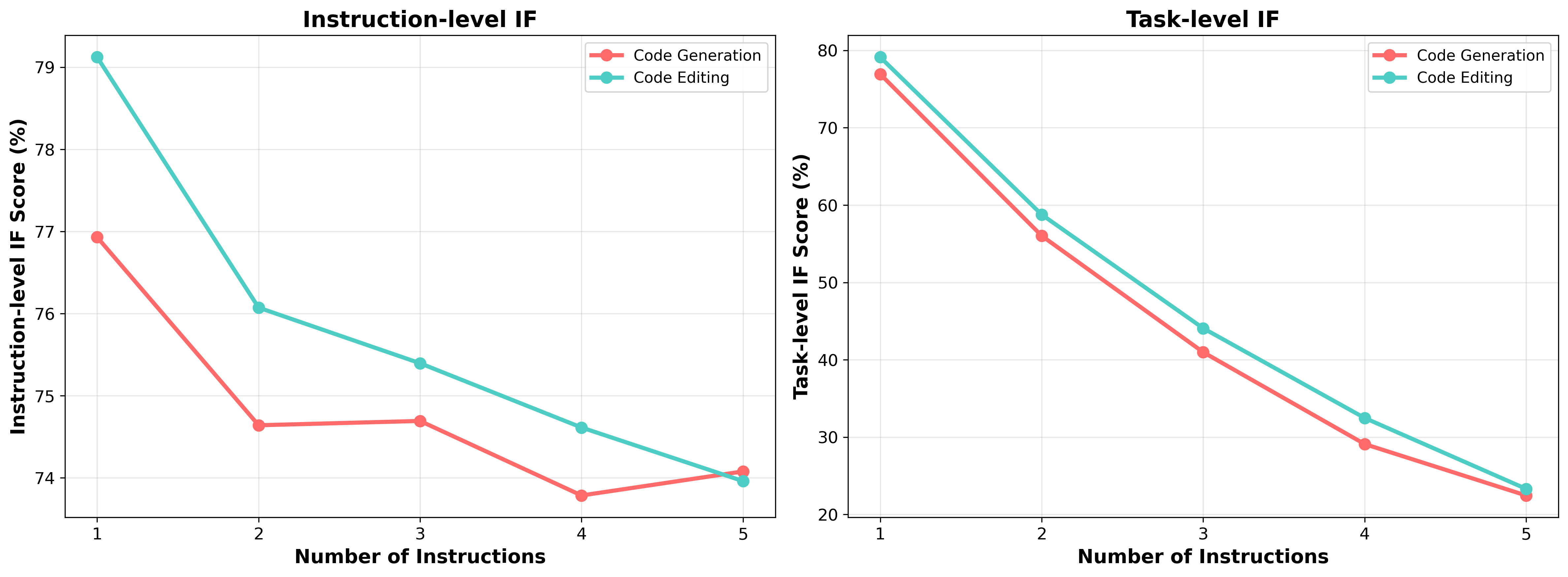}
        \caption{Gemini}
        \label{fig:trend_if_gemini}
    \end{subfigure}
    \hfill
    \begin{subfigure}[b]{0.48\textwidth}
        \centering
        \includegraphics[width=\textwidth]{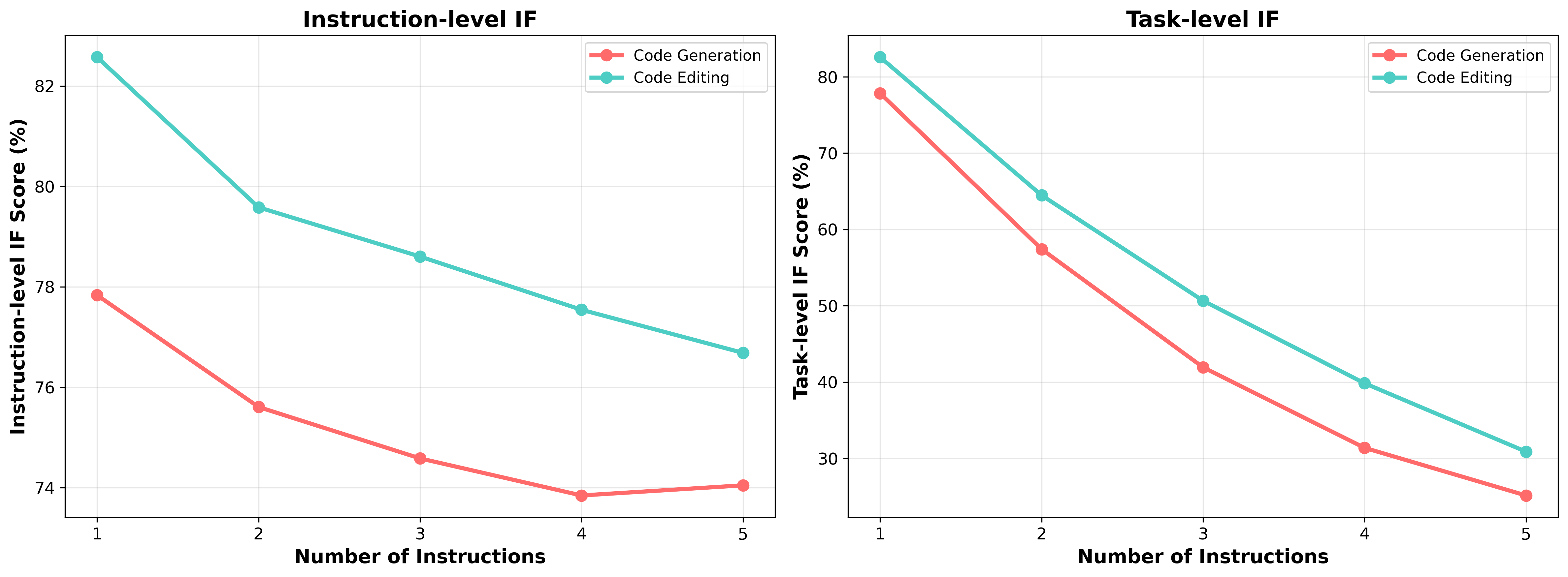}
        \caption{Claude}
        \label{fig:trend_if_claude}
    \end{subfigure}

    \vspace{1mm}

    \begin{subfigure}[b]{0.48\textwidth}
        \centering
        \includegraphics[width=\textwidth]{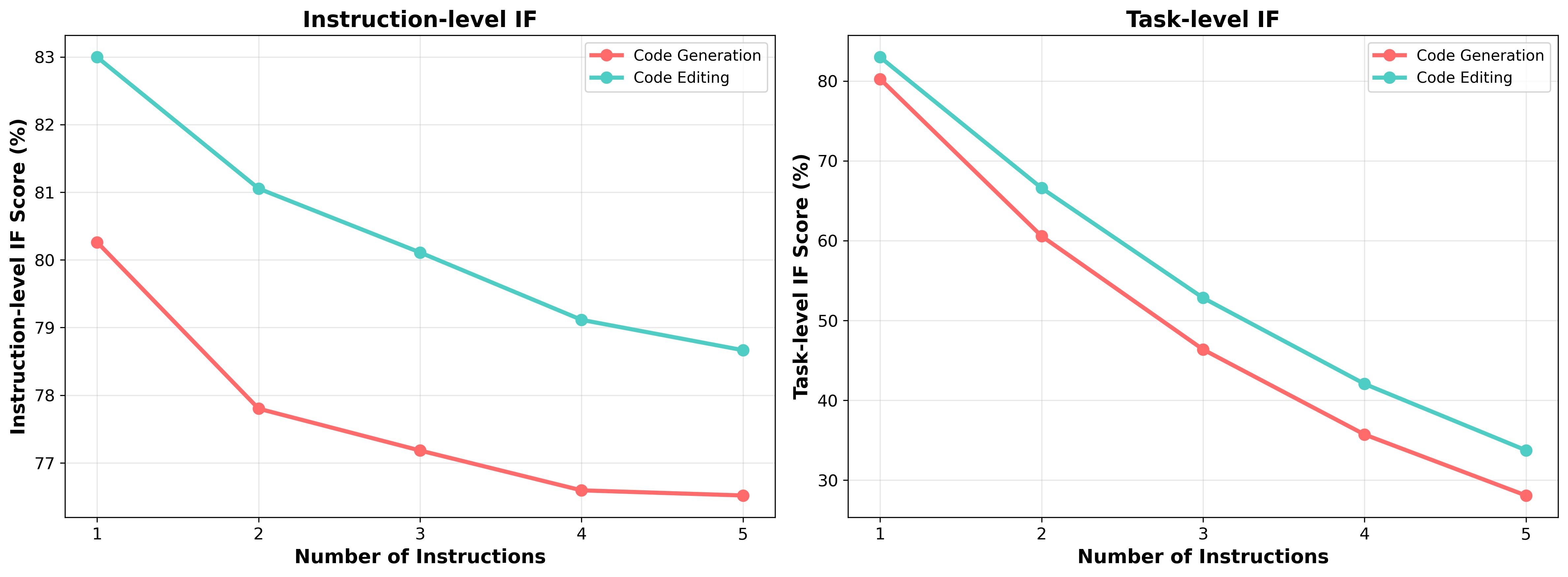}
        \caption{OpenAI}
        \label{fig:trend_if_openai}
    \end{subfigure}
    \hfill
    \begin{subfigure}[b]{0.48\textwidth}
        \centering
        \includegraphics[width=\textwidth]{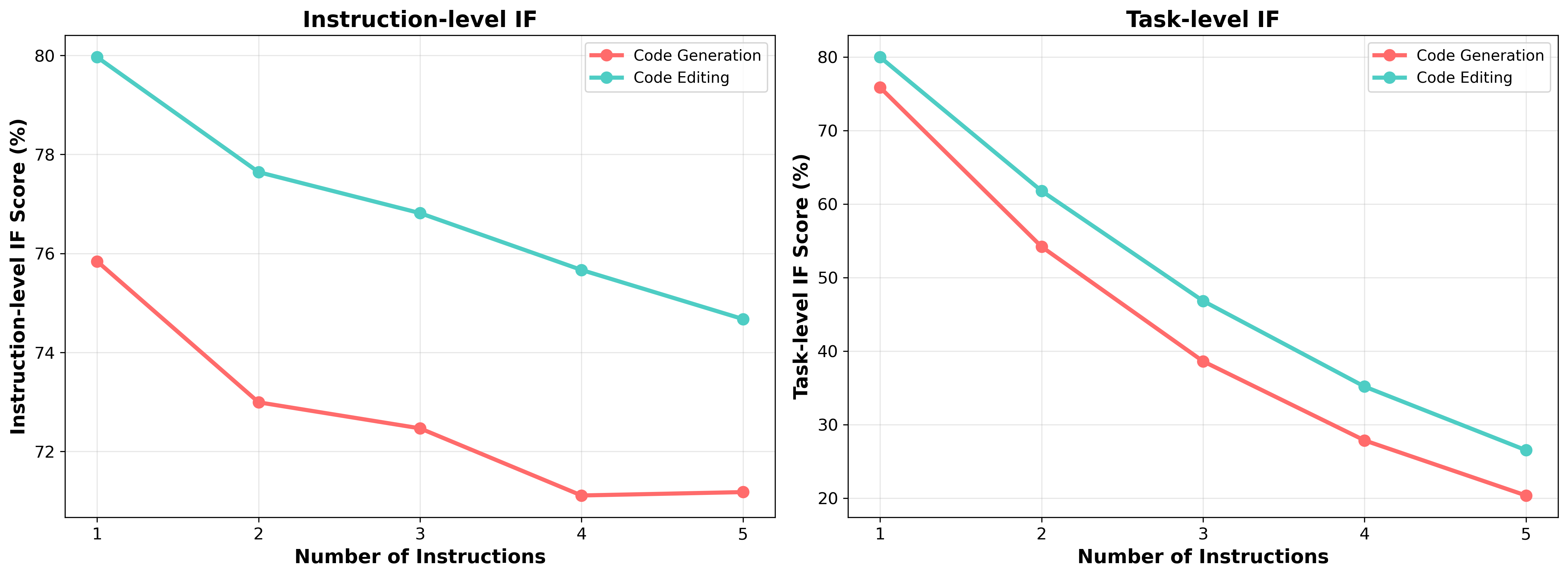}
        \caption{Qwen}
        \label{fig:trend_if_qwen}
    \end{subfigure}

    \caption{\textbf{Results for instruction following across model families.} Each subplot reports instruction-level and task-level IF scores versus the number of injected instructions, for single-turn generation and multi-turn editing.}
    \label{fig:trending_if_families}
\end{figure}

To highlight the overall trends at the family level, we report trajectories of functionality and instruction following as the number of injected instructions increases (Figures~\ref{fig:trending_func_families} and~\ref{fig:trending_if_families}).

\vspace{-4mm}

\vspace{-2mm}

Across all families, adding more instructions consistently increases functional regression (Figure~\ref{fig:trending_func_families}) and decreases IF (Figure~\ref{fig:trending_if_families}). Moreover, multi-turn editing typically yields higher IF than single-turn generation, while single-turn generation tends to preserve functionality better.

\clearpage
\newpage
\subsection{Statistical Significance and Result Stability}
\label{app:stat_sig}

Our experimental design incorporates several measures to ensure robustness: we utilize a large-scale dataset comprising over 2,000 instances and 10,000 instructions for evaluation. Moreover, we conduct experiments using deterministic or low temperatures (0.0 and 0.2) across a diverse set of 31 models to ensure our conclusions are comprehensive and applicable. To further assess run-to-run stability, we additionally conduct repeated experiments (five runs) on the Gemini model family on Big-SWE-IF. Table~\ref{tab:stat_sig_gemini} reports mean pass@1 and standard deviation.

\renewcommand\arraystretch{1.25}
\begin{table}[H]
    \centering \footnotesize
    \tabcolsep0.12 in

    \begin{NiceTabular}{lcccccc}
        \toprule
        \multicolumn{7}{l}{\textbf{Single-Turn Generation (Big-SWE-IF, 5 runs)}} \\
        \midrule
        \textbf{Models} & \textbf{0 Inst} & \textbf{1 Inst} & \textbf{2 Inst} & \textbf{3 Inst} & \textbf{4 Inst} & \textbf{5 Inst} \\
        \midrule
        Gemini 2.5 Pro        & 50.48 $\pm$ 0.22 & 50.11 $\pm$ 0.25 & 49.06 $\pm$ 0.16 & 49.95 $\pm$ 0.18 & 50.41 $\pm$ 0.29 & 49.62 $\pm$ 0.16 \\
        Gemini 2.5 Flash      & 47.38 $\pm$ 0.19 & 46.90 $\pm$ 0.25 & 46.74 $\pm$ 0.31 & 46.24 $\pm$ 0.18 & 46.62 $\pm$ 0.28 & 46.22 $\pm$ 0.08 \\
        Gemini 2.0 Flash      & 48.37 $\pm$ 0.18 & 47.30 $\pm$ 0.22 & 48.14 $\pm$ 0.37 & 47.72 $\pm$ 0.19 & 46.79 $\pm$ 0.26 & 46.08 $\pm$ 0.35 \\
        Gemini 2.0 Flash Lite & 46.95 $\pm$ 0.24 & 44.65 $\pm$ 0.41 & 43.49 $\pm$ 0.07 & 43.58 $\pm$ 0.25 & 43.62 $\pm$ 0.11 & 42.89 $\pm$ 0.19 \\
        \midrule
        \multicolumn{7}{l}{\textbf{Multi-Turn Editing (Big-SWE-IF, 5 runs)}} \\
        \midrule
        \textbf{Models} & \textbf{0 Inst} & \textbf{1 Inst} & \textbf{2 Inst} & \textbf{3 Inst} & \textbf{4 Inst} & \textbf{5 Inst} \\
        \midrule
        Gemini 2.5 Pro        & 50.48 $\pm$ 0.22 & 49.50 $\pm$ 0.08 & 49.09 $\pm$ 0.12 & 48.35 $\pm$ 0.15 & 47.95 $\pm$ 0.18 & 47.79 $\pm$ 0.06 \\
        Gemini 2.5 Flash      & 47.38 $\pm$ 0.19 & 47.28 $\pm$ 0.29 & 46.94 $\pm$ 0.11 & 46.81 $\pm$ 0.09 & 46.69 $\pm$ 0.07 & 45.92 $\pm$ 0.14 \\
        Gemini 2.0 Flash      & 48.37 $\pm$ 0.18 & 47.00 $\pm$ 0.13 & 45.99 $\pm$ 0.10 & 45.24 $\pm$ 0.06 & 44.41 $\pm$ 0.19 & 43.89 $\pm$ 0.16 \\
        Gemini 2.0 Flash Lite & 46.85 $\pm$ 0.24 & 45.41 $\pm$ 0.35 & 44.75 $\pm$ 0.15 & 44.40 $\pm$ 0.11 & 42.92 $\pm$ 0.08 & 42.77 $\pm$ 0.12 \\
        \bottomrule
    \end{NiceTabular}

    \vspace{2mm}
    \caption{\textbf{Statistical significance analysis via repeated runs.} We report mean pass@1 (\%) and standard deviation across five runs for the Gemini model family on Big-SWE-IF.}
    \label{tab:stat_sig_gemini}
\end{table}

\textbf{High stability.}
Across all instruction counts and both interaction modes, the standard deviations are consistently small (mostly below 0.35), indicating that the regression trends are unlikely to be artifacts of run-to-run randomness.

\textbf{Interaction mode stability.}
Multi-turn editing often exhibits even lower variance than single-turn generation, consistent with the richer context provided by prior rounds.

\clearpage
\newpage
\section{Limitations}

Our evaluation follows the prevailing setting in code-generation benchmarks by focusing on Python function-level tasks with deterministic verifiers. We discuss the main limitations of our current instantiation along three axes: (1) the focus on Python, (2) the focus on function-level tasks, and (3) the scope of deterministic verifiers; in each case, the underlying framework is designed to generalize beyond the current choice.

\subsection{Experimental Language Choice}
\label{app:language_choice}

To clarify the scope of our study, we explain why our experiments focus on Python while the overall framework remains language-agnostic.

\textbf{Python-centric experimental choice.}
Our decision to focus on Python is driven by the current code evaluation landscape, which is highly Python-centric. Many widely used and frequently reported benchmarks in recent LLM literature, including BigCodeBench, LiveCodeBench, and SWE-Bench, are Python-only. Using these benchmarks keeps our task distribution directly comparable to prior work.

\textbf{Language-agnostic framework design.}
At the same time, the framework design is language-agnostic. Major programming languages have established industrial standards and mature linter ecosystems. Our pipeline of sourcing rules from these standards, applying filtering, and implementing deterministic verifiers can be applied to other languages, for example by using ESLint for JavaScript or Checkstyle for Java.

\subsection{Why We Focus on Function-level Benchmarks}
\label{app:function_level_focus}

To clarify the evaluation setting, we summarize why function-level tasks are a practical starting point.

\begin{itemize}[leftmargin=*]
    \item \textbf{Difficulty and foundational capability.} Even at the function level, current models still struggle to robustly follow multiple instructions (see our task-level IF under five instructions). Mastering these atomic constraints is a prerequisite for scaling to multi-file contexts with added dependency and consistency challenges.
    \item \textbf{Alignment with available human preference data.} Human preference data for coding, such as LMArena, is dominated by snippet-level or single-file interactions rather than repository-level tasks, making function-level benchmarks a better match to current preference distributions.
    \item \textbf{Methodological extensibility.} Our plug-in methodology is not limited to functions. Since the taxonomy derives from industrial standards (e.g., linters), verifiable instructions can be injected into broader settings, including repository-level benchmarks such as SWE-Bench.
\end{itemize}

\subsection{Scope of Deterministic Verifiers}
\label{app:verifier_scope}

Our verifiers are deterministic, implemented via industrial linters, AST analysis, and regular expressions, providing reliable, reproducible, and scalable pass/fail signals. This is a deliberate and practical starting point that enables objective evaluation without the cost or variance of human or LLM judgment. Nonetheless, real-world and agentic coding increasingly involve open-ended, subjective requirements such as API design, maintainability, and algorithmic elegance that lie beyond what deterministic checks can capture. While our verifiable instructions already correlate well with human preference, extending the framework to such open-ended constraints, for instance through LLM-as-a-judge verification, is an important direction as vibe coding matures. We further note that static analysis can yield rare false positives or negatives in complex cases; since our framework is orthogonal to the specific checkers used, more advanced tools can be integrated as drop-in replacements.

\end{document}